\documentclass{article}

\usepackage[nohyperref,accepted]{icml2023}

\usepackage{amsmath,amsfonts,bm}

\def\eqref#1{equation~\ref{#1}}

\def\floor#1{\left\lfloor #1 \right\rfloor}

\def\eps{{\epsilon}}

\DeclareMathAlphabet{\mathsfit}{\encodingdefault}{\sfdefault}{m}{sl}
\SetMathAlphabet{\mathsfit}{bold}{\encodingdefault}{\sfdefault}{bx}{n}

\newcommand{\R}{\mathbb{R}}

\def\LG{\mathcal L_{\mathcal G}}

\usepackage{shortcuts}

\usepackage{hyperref}
\definecolor{linkcolor}{RGB}{83,83,182}
\hypersetup{
    colorlinks=true,
    citecolor=linkcolor,
    linkcolor=linkcolor
}
\usepackage{url}

\theoremstyle{plain}
\newtheorem{theorem}{Theorem}[section]

\newtheorem{remark}[theorem]{Remark}

\icmltitlerunning{FaDIn: Fast Discretized Inference for Hawkes Processes with General Parametric Kernels}

\begin{document}

\twocolumn[
\icmltitle{FaDIn: Fast Discretized Inference for Hawkes Processes with General Parametric Kernels}



\icmlsetsymbol{equal}{*}

\begin{icmlauthorlist}
\icmlauthor{Guillaume Staerman}{equal,yyy}
\icmlauthor{Cédric Allain}{equal,yyy}
\icmlauthor{Alexandre Gramfort}{yyy}
\icmlauthor{Thomas Moreau}{yyy}

\end{icmlauthorlist}

\icmlaffiliation{yyy}{Université Paris-Saclay, Inria, CEA, Palaiseau, 91120, France}

\icmlcorrespondingauthor{Guillaume Staerman}{guillaume.staerman@inria.fr}

\icmlkeywords{Machine Learning, ICML}

\vskip 0.3in
]



\printAffiliationsAndNotice{\icmlEqualContribution} 

\begin{abstract}
Temporal point processes (TPP) are a natural tool for modeling event-based data. Among all TPP models, Hawkes processes have proven to be the most widely used, mainly due to their adequate modeling for various applications, particularly when considering exponential or non-parametric kernels. Although non-parametric kernels are an option, such models require large datasets. While exponential kernels are more data efficient and relevant for specific applications where events immediately trigger more events, they are ill-suited for applications where latencies need to be estimated, such as in neuroscience. This work aims to offer an efficient solution to TPP inference using general parametric kernels with finite support. The developed solution consists of a fast  $\ell_2$ gradient-based solver leveraging a discretized version of the events. After theoretically supporting the use of discretization, the statistical and computational efficiency of the novel approach is demonstrated through various numerical experiments. Finally, the method's effectiveness is evaluated by modeling the occurrence of stimuli-induced patterns from brain signals recorded with magnetoencephalography (MEG).
Given the use of general parametric kernels, results show that the proposed approach leads to an improved estimation of pattern latency than the state-of-the-art.
\end{abstract}

\section{Introduction}

The statistical framework of Temporal Point Processes (TPPs; see \eg ~\citealt{daley2003introduction}) is well adapted for modeling event-based data.
It offers a principled way to predict the rate of events as a function of time and the previous events' history.
TPPs are historically used to model intervals between events, such as in renewal theory, which studies the sequence of intervals between successive replacements of a component susceptible to failure.
TPPs find many applications in neuroscience, in particular, to model single-cell recordings and neural spike trains~\citep{truccolo2005point, okatan2005analyzing, kim2011granger,rad2011information}, occasionally associated with spatial statistics~\citep{pillow2008spatio} or network models~\citep{galves2015modeling}.
%
Multivariate Hawkes processes (MHP;~\citealt{hawkes1971point}) are likely the most popular, as they can model interactions between each univariate process.
They also have the peculiarity that a process can be self-exciting, meaning that a past event will increase the probability of having another event in the future on the same process.
The conditional intensity function is the key quantity for TPPs. With MHP, it is composed of a baseline parameter and kernels.
It describes the probability of occurrence of an event depending on time.
The kernel function represents how processes influence each other or themselves.
The most commonly used inference method to obtain the baseline and the kernel parameters of MHP is the maximum likelihood (MLE; see \eg \citealp{daley2007introduction} or \citealp{lewis2011nonparametric}). One alternative and often overlooked estimation criterion is the least squares $\ell_2$ error, inspired by the theory of empirical risk minimization (ERM; \citealt{reynaud2010near,hansen2015lasso,bacry2020sparse}).

%
A key feature of MHP modeling is the choice of kernels that can be either
non-parametric or parametric.
In the non-parametric setting, kernel functions are approximated by histograms \citep{lewis2011nonparametric,lemonnier2014nonparametric}, by a linear combination of pre-defined functions \citep{zhou2013learning,xu2016learning} or, alternatively,  by functions lying in a RKHS \citep{yang2017online}.
In addition to the frequentist approach, many Bayesian approaches, such as Gibbs sampling \citep{ishwaran2001gibbs} or (stochastic) variational inference \citep{hoffman2013stochastic}, have been adapted to MHP in particular to fit non-parametric kernels.
Bayesian methods also rely on the modeling of the kernel by histograms (\eg \citealp{rousseau2018nonparametric}) or by a linear combination of pre-defined functions (\eg \citealp{linderman2015scalable}).
These approaches are designed whether in continuous-time \citep{rasmussen2013bayesian,zhang2018efficient,rousseau2018nonparametric,sulem2021bayesian} or in discrete-time \citep{mohler2013modeling,linderman2015scalable,zhang2018efficient,browning2022flexible}.
These functions allow great flexibility for the shape of the kernel, yet this comes at the risk of poor estimation of it when only a small amount of data is available \citep{xu2017learning}. 
Another approach to estimating the intensity function is to consider parametrized kernels.
Although it can introduce a potential bias by assuming a particular kernel shape, this approach has several benefits.
First, it reduces inference burden, as the parameter, say $\eta$, is typically lower dimensional than non-parametric kernels.
Moreover, for kernels satisfying the Markov property \citep{bacry2015hawkes}, computing the conditional intensity function is linear in the total number of timestamps/events.
The most popular kernel belonging to this family is the exponential kernel \citep{ogata1981lewis}. It is defined by $\eta=(\alpha,\gamma)\mapsto\alpha \gamma \exp(-\gamma t)$, where $\alpha$ and $\gamma$ are the scaling and the decay parameters, respectively \citep{veen2008estimation,zhou2013learninglasso}.
However, as pointed out by \citet{lemonnier2014nonparametric}, the maximum likelihood estimator for MHP with exponential kernels is efficient only if the decay $\gamma$ is fixed.
Thus, only the scaling parameter $\alpha$ is usually inferred.
This implies that the hyperparameter $\gamma$ must be chosen in advance, usually using a grid search, a random search, or Bayesian optimization. This leads to a computational burden when the dimension of the MHP is high.
The second option is to define a $\gamma$ decay parameter common to all kernels, which results in a loss of expressiveness of the model.
In both cases, the relevance of the exponential kernel relies on the choice of the decay parameter, which may not be adapted to the data \citep{hall2016tracking}.
For more general parametric kernels which do not verify the Markov property, the inference procedure with both MLE or $\ell_2$ loss scales poorly as they have quadratic computational scaling with the number of events, making their use limited in practice (see \eg \citealp[Chapter~1]{bompaire:tel-02316143}).
Recently, neural network-based MHP estimation has been introduced, offering, with sufficient data, relevant models at the cost of high computational cost \citep{mei2017neural,shchur2019intensity,pan2021self}.
These limitations for parametric and non-parametric kernels prevent their usage in some applications, as pointed out by \citet{Carreira2021} in finance or \citet{allain2021dripp} in neuroscience.
A strong motivation for this work is also neuroscience applications.

The quantitative analysis of electrophysiological signals such as electroencephalography (EEG) or magnetoencephalography (MEG) is a challenging modern neuroscience research topic~\citep{cohen-book:14}. 
By giving a non-invasive way to record human neural activity with a high temporal resolution, EEG and MEG offer a unique opportunity to study cognitive processes as triggered by controlled stimulation \citep{baillet2017}.
Convolutional dictionary learning (CDL) is an unsupervised algorithm recently proposed to study M/EEG signals~\citep{Jas2017, dupre2018multivariate}. It consists in extracting patterns of interest in M/EEG signals. It learns a  combination of time-invariant patterns -- called \textit{atoms} -- and their activation function to reconstruct the signal sparsely. 
However, while CDL recovers the local structure of signals, it does not provide any global information, such as interactions between patterns or how their activations are affected by stimuli. Atoms typically correspond to transient bursts of neural activity~\citep{ShermanE4885} or artifacts such as eye blinks or heartbeats.
By offering an event-based perspective on non-invasive electromagnetic brain signals, CDL makes Hawkes processes amenable to M/EEG-based studies. Given the estimated events, one important goal is to uncover potential temporal dependencies between external stimuli presented to the subject and the appearance of the atoms in the data.
More precisely, one is interested in statistically quantifying such dependencies, \eg by estimating the mean and variance of the neural response latency following a stimulus. In~\citet{allain2021dripp}, the authors address this precise problem.
Their approach is based on an EM algorithm and a Truncated Gaussian kernel, which can cope with only a few brain data, as opposed to non-parametric kernels, which are more data-hungry.
Beyond neuroscience, \citet{Carreira2021} uses a likelihood-based approach using exponential kernels to model order book events. Their approach uses high-frequency trading data, considering the latency at hand in the proposed loss.

This paper proposes a new inference method -- named FaDIn -- to estimate any parametric kernels for Hawkes processes.
Our approach is based on two key features.
First, we use finite-support kernels and a discretization applied to the ERM-inspired least-squares loss.
Second, we propose to employ some precomputations that significantly reduce the computational cost.
We then show, empirically and theoretically, that the implicit bias induced by the discretization procedure is negligible compared to the statistical error.
Further, we highlight the efficiency of FaDIn in computation and statistical estimation over the non-parametric approach. Finally, we demonstrate the benefit of using a general kernel with MEG data.
The flexibility of FaDIn allows us to model neural response to external stimuli with a much better-adapted kernel than the existing method derived in \citet{allain2021dripp}.

\section{Fast Discretized Inference for Hawkes processes (FaDIn)}
After recalling key notions of Hawkes processes, we introduce our proposed framework FaDIn.
\subsection{Hawkes processes}

Given a stopping time $T\in \mathbb{R}_+$ and an observation period $[0,T]$, a temporal point process (TPP) is a stochastic process whose realization consists of a set of distinct timestamps ${\mathscr{F}_T=\{t_n, \; t_n \in [0,T] \}}$  occurring in continuous time.
The behavior of a TPP is fully characterized by its \textit{intensity function} that corresponds to the expected infinitesimal rate at which events are occurring at time $t\in [0,T]$.
The values of this function may depend on time (\eg inhomogeneous Poisson processes) or rely on past events such as self-exciting processes (see \citealt{daley2003introduction} for an excellent account of TPP).
For the latter, the occurrence of one event will modify the probability of having a new event in the near future.
The conditional intensity function $\lambda: [0, T] \rightarrow \bbR_+$ has the following form: 
\begin{equation*}
    \lambda\pars{t \middle| \mathscr{F}_t} \coloneqq \lim_{\dint t \to 0} \frac{\proba{N_{t + \dint t} - N_t = 1}[\mathscr{F}_t]}{\dint t}
      ,
\end{equation*}
where $N_t \coloneqq \sum_{n \geq 1} \1[t_n \leq t]$ is the counting process associated to the PP.
Among this family, Multivariate Hawkes processes (MHP; \citealp{hawkes1971point}) model the interactions of $p\in \bbN_*$ self-exciting TPPs.
Given $p$ sets of timestamps $\mathscr{F}_T^i = \{t_n^i, \; t_n^i\in [0, T] \}_{n=1}^{N_T^i}, i=1, \dots, p$
, each process is described by the following intensity function:
\begin{equation}\label{eq:intensity_mhp}
    \lambda_i (t) = \mu_i + \sum_{j=1}^{p} \int_{0}^{t} \phi_{ij} (t-s)~\mathrm{d}N_s^{j}   ,
\end{equation}
where $\mu_i$ is the baseline parameter, $N_t=[N_t^1,\ldots, N_t^p]$  the associated multivariate counting process and $\phi_{ij}:[0, T] \rightarrow \bbR_+$ the excitation function -- called \textit{kernel} -- representing the influence of $j$-th process' past events onto $i$-th process' future events.
From an inference perspective, the goal is to estimate the baseline and kernels associated with the MHP from the data.
In this paper, we focus on the ERM-inspired least squares loss. Assuming a class of parametric kernel parametrized by $\eta$, the objective is to find parameters that minimize (see \eg Eq. (I.2) in \citealp[Chapter~1]{bompaire:tel-02316143}):
%
\begin{equation}\label{eq:l2continuous}
    \mathcal{L}\pars{\theta, \mathscr{F}_T} =   \frac{1}{N_T}\sum_{i=1}^{p}  \pars{\int_{0}^{T}\lambda_{i}(s)^2~\mathrm{d}s - 2 \hspace{-0.2cm}\sum_{t_n^{i} \in \mathscr{F}_T^i} \hspace{-0.1cm}\lambda_{i}\pars{t_n^{i}}} ,
\end{equation}
where 
$N_T=\sum_{i=1}^p N_T^i$ is the total number of timestamps, and where $\theta \coloneqq (\mu, \eta)$.
Interestingly, when used with an exponential kernel, this loss benefits from some precomputations of complexity $O(N_T)$, making the subsequent iterative optimization procedure independent of $N_T$. This computational ease is the main advantage of the loss $\mathcal{L}$ over the log-likelihood function.
However, when using a general parametric kernel, these precomputations require $O((N_T)^2)$ operations killing the computational benefit of the $\ell_2$ loss $\mathcal{L}$ over the log-likelihood. It is worth noting that this loss differs from the quadratic error minimized between the counting processes and the integral of the intensity function, as used in \citet{wang2016isotonic, eichler2017graphical} and \citet{xu2018benefits}.

\subsection{FaDIn}

The approach we propose in this paper fills the need for general parametric kernels in many applications.
We provide a computationally and statistically efficient solver -- coined FaDIn -- that works with many parametric kernels using gradient-based algorithms.
Precisely, it relies on three key ideas:
($i$) the use of parametric finite-support kernels,
($ii$) a discretization of the time interval $[0, T]$, and
($iii$) precomputations allowing an efficient optimization procedure detailed below.

\noindent \textbf{Finite support kernels} A core bottleneck for MLE or $\ell_2$ estimation of parametric kernels is the need to compute the intensity function for all events.
For general kernels, the intensity function usually requires $O((N_T)^2)$ operations, which makes it intractable for long-time-length processes.
To make this computation more efficient, we consider finite support kernels. 
Using a finite support kernel amounts to setting a limit in time for the influence of a past event on the intensity, \ie $\forall t \notin \intervalleFF{0}{W}, \phi_{ij}(t) = 0$, where $W$ denotes the length of the kernel's support.
This assumption matches applications where an event cannot have influence far in the future, such as in neuroscience \citep{krumin2010correlation, eichler2017graphical, allain2021dripp}, genetics \cite{reynaud2010} or high-frequency trading \citep{bacry2015hawkes, Carreira2021}.
The intensity function~\autoref{eq:intensity_mhp} can then be reformulated as a convolution between the kernel $\phi_{ij}$ and the sum of Dirac functions $z_i (t)=\sum_{t^i_n \in \mathscr F^i_t} \delta_{t^i_n} (t)$ located at the event occurrences $t^i_n$:
$$
    \lambda_i(t) = \mu_i + \sum_{j=1}^p \phi_{ij} * z_j(t), \quad t\in \intervalleFF{0}{T}.
$$
As $\phi_{ij}$ has finite support, the intensity can be computed efficiently with this formula.
Indeed, only events in the interval $\intervalleFF{t-W}{t}$ need to be considered. See \autoref{sec:app:w} for more details.



\noindent \textbf{Discretization} To make these computations even more efficient, we propose to rely on discretized processes.
Most Hawkes processes estimation procedures involve a continuous paradigm to minimize (\ref{eq:l2continuous}) or its log-likelihood counterpart.
Discretization has been investigated so far for non-parametric kernels~\citep{Kirchner2016,kirchner2018nonparametric,kurisu2018discretization}.
The discretization of a TPP consists in projecting each event $t^i_n$ on a regular grid $\mathcal{G}=\{0, \Delta, 2\Delta, \ldots, G\Delta \}$, where $G \Delta =T$. We refer to $\Delta$ as the stepsize of the discretization. 
Let $\widetilde{\mathscr{F}}_T^i$ be the set of projected timestamps of $\mathscr{F}_T^i$ on the grid $\mathcal{G}$. The intensity function of the $i$-th process of our discretized MHP is defined as:
\begin{align}\label{eq:intens_sum}
    \tilde{\lambda}_i[s] &= \mu_i + \sum_{j=1}^{p} \sum_{\tilde{t}_m^{j} \in \widetilde{\mathscr{F}}_{s\Delta}^j} \phi_{ij} (s\Delta - \tilde{t}_m^j)
    \nonumber \\& = \mu_i + \sum_{j=1}^{p} \underbrace{\sum_{\tau=1}^L \phi_{ij}^\Delta[\tau] z_j[s-\tau]}_{(\phi_{ij}^\Delta * z_j)[s]},
    \quad s\in \intervalleEntier{0}{G}  ,
\end{align}
where $L = \floor{\frac W \Delta}$ denotes the number of points on the discretized support, $\phi_{ij}^\Delta[s] = \phi_{ij}(s\Delta)$ is the kernel value on the grid and $z_i[s] = \#\braces{|t_n^i - s\Delta|\leq \frac{\Delta}{2}}$ denotes the number of events projected on the grid timestamp $s$. Here $\floor{\cdot}$ denotes the floor function. From now and throughout the rest of the paper, we denote $\phi_{ij}(\cdot):  \mathbb{R}_+ \rightarrow \mathbb{R}_+$ as a  function  while $\phi_{ij}^{\Delta}[\cdot]$ represents the discrete vector $\phi_{ij}^{\Delta}\in \mathbb{R}_+^L$. Compared to the continuous formulation, the intensity function can be computed more efficiently as one can rely on discrete convolutions, whose worst-case complexity scales as $O(N_T L)$.
It can also be further accelerated using Fast Fourier Transform when $N_T$ is large.
Another benefit of the discretization is that for kernels whose values are costly to compute, at most $L$ values need to be calculated.
This can have a strong computational impact when $N_T \gg L$ as all values can be precomputed and stored.

\looseness=-1
While discretization improves the computational efficiency, it also introduces a bias in the computation of the intensity function and, thus potentially, in estimating the kernel parameters. The impact of the discretization on the estimation is considered in \autoref{sub:bias} and \autoref{subsec:consistency}.
Note that this bias is similar to the one incurred by quantizing the kernel as histograms for non-parametric estimators.

\noindent \textbf{Loss and precomputations}
FaDIn aims at minimizing the discretized $\ell_2$ loss, which approximates the integral on the left part of (\ref{eq:l2continuous}) by a sum on the grid $\mathcal G$ after projecting timestamps of $\mathscr{F}_T$ on it.
It boils down to optimizing the following loss $\mathcal{L}_{\mathcal G}\pars{\theta, \widetilde{\mathscr{F}}_T}$ defined as:

\begin{align}
\label{eq:l2discret}
  \frac{1}{N_T}\sum_{i=1}^{p}  \pars{\Delta\sum_{s\in \intervalleEntier{0}{G}}\pars{\tilde{\lambda}_{i}[s]}^2
            - 2\sum_{\tilde{t}_n^i \in \widetilde{\mathscr{F}}_T^i}\tilde{\lambda}_{i}\bracks{\frac{\tilde{t}_n^{i}}{\Delta}}} .
\end{align}

To find the parameters of the intensity function $\theta$, FaDIn minimizes $\LG{}$ using a first-order gradient-based algorithm.
The computational bottleneck of the proposed algorithm is thus the computation of the gradient $\nabla\LG{}$ regarding parameters $\theta$.
Using the discretized finite-support kernel, this gradient can be computed using convolution, giving the same computational complexity as the computation of the intensity function $ O(N_T L)$.
However, gradient computation can still be too expensive for long processes with many events to get reasonable inference times.
Using the least squares error of the process\autoref{eq:l2discret}, one can further reduce the complexity of computing the gradient by precomputing some constants $\Phi_j(\tau; G)$, $\Psi_{j,k}(\tau, \tau'; G)$ and $\Phi_j(\tau; \widetilde{\mathscr{F}}_T^i)$ that do not depend on the parameter $\theta$.
Indeed, by developing and rearranging the terms in\autoref{eq:l2discret}, one obtains:
{\small
\begin{align*}
   &N_T~\mathcal{L}_{\mathcal G}\pars{\theta, \widetilde{\mathscr{F}}_T}=  \\& 
   (T+\Delta)\sum_{i=1}^{p} \mu_i^2+2\Delta \sum_{i=1}^{p}\mu_i \sum_{j=1}^{p} \sum_{\tau=1}^{L} \phi_{ij}^\Delta[\tau] \underbrace{\left(\sum_{s=1}^{G} z_{j}[s-\tau] \right)}_{\Phi_j(\tau; G)} \\& 
    + \Delta \sum_{i,j,k } \sum_{\tau=1}^{L}\sum_{\tau'=1}^{L} \phi_{ij}^\Delta[\tau] \phi_{ik}^\Delta[\tau'] \underbrace{\left( \sum_{s=1}^{G} z_{j}[s-\tau]~z_{k}[s-\tau '] \right)}_{\Psi_{j,k}(\tau, \tau'; G)} \\&
    - 2\Biggl(\sum_{i=1}^{p} N_T^i \mu_i + \sum_{ i,j}\sum_{\tau=1}^{L} \phi_{ij}^\Delta[\tau] \underbrace{\Biggl(\sum_{\tilde{t}_n^i \in \widetilde{\mathscr{F}}_T^i} z_{j}\bracks{\frac{\tilde{t}_n^i}{\Delta}-\tau}\Biggl)}_{\Phi_j\pars{\tau; \widetilde{\mathscr{F}}_T^i}}\Biggl) .
\end{align*}}

\noindent The term $\Psi_{j,k}(\tau, \tau'; G)$ dominates the computational cost of our precomputations.
It requires $O(G)$ operations for each tuples $(\tau, \tau')$ and $(j,k)$. 
Thus, it has a  total complexity of $O(p^2L^2G)$ and is the bottleneck of the precomputation phase.
For any $m\in \{1,\ldots, p \}$, the gradient of the loss w.r.t. the baseline parameter is given by:
{\small
\begin{equation*}
    N_T\frac{\partial\mathcal{L}_{\mathcal{G}}}{\partial \mu_{m}} 
     =2 (T+\Delta)\mu_m -  2N_T^m + 2\Delta\sum_{j=1}^{p} \sum_{\tau=1}^{L} \phi_{mj}^\Delta[\tau]\Phi_{j}(\tau; G) .  
\end{equation*}}
For any tuple $(m,l)\in \{1,\ldots, p\}^2$, the gradient of  $\eta_{ml}$ is:
{\small
\begin{align*}
        N_T\frac{\partial\mathcal{L}_{\mathcal{G}}}{\partial \eta_{ml}} &= 2\Delta \mu_m  \sum_{\tau=1}^{L} \frac{\partial \phi_{m,l}^\Delta[\tau]}{\partial \eta_{m,l}}~ \Phi_l(\tau; G)
        \\& \; \; \; + 2\Delta \sum_{k=1}^{p} \sum_{\tau=1}^{L} \sum_{\tau'=1}^{L} \phi_{mk}^\Delta[\tau'] \frac{\partial \phi_{m,l}^\Delta[\tau]}{\partial \eta_{m,l}}~\Psi_{l,k}(\tau, \tau'; G) \\&
        \; \; \;  -2 \sum_{\tau=1}^{L} \frac{\partial \phi_{m,l}^\Delta[\tau]}{\partial\eta_{m,l}}~ \Phi_l(\tau; \widetilde{\mathscr{F}}_T^m)  .
\end{align*}}

\noindent Gradients of kernel parameters dominate the computational cost of gradients.
The complexity is of $O(pL^2)$ for each kernel parameter, leading to a total complexity of $O(p^3L^2)$ and is independent of the number of events $N_T$.
Thus, a trade-off can be made between the precision of the method and its computational efficiency when varying the size of the kernel's support or the discretization.

\begin{remark}
The primary motivation for the  $\ell_2$ loss is the presence of terms that can be precomputed in contrast to the log-likelihood \citep{reynaud2010near,Reynaud-Bouret2014,bacry2020sparse}. A comparison is performed in~\autoref{sec:app:ll}.

\end{remark}


\paragraph{Optimization}
The inference is then conducted using gradient descent for the $\ell_2$ loss $\LG{}$. FaDIn thus allows for very general parametric kernels, as exact gradients for each parameter involved in the kernels can be derived efficiently as long as the kernel is differentiable and has finite support.
Gradient-based optimization algorithms can, therefore, be used without limitation, in contrast with the EM algorithm, which requires a close-form solution to zero the gradient, which is difficult for many kernels.
A critical remark is that the problem is generally non-convex and may converge to a local minimum.

\subsection{Impact of the discretization}
\label{sub:bias}

While discretization allows for efficient computations, it also introduces a perturbation in the loss value. In this section, we quantify the impact of this perturbation on the parameter estimation when $\Delta$ goes to 0.
Through this section, we observe a process $\mathscr F_T$ whose intensity function is given by the parametric form $\lambda(\cdot; \theta^*)$. Note that if the process $\mathscr F_T$'s intensity is not in the parametric family $\lambda(\cdot; \theta)$, $\theta^*$ is defined as the best approximation of its intensity function in the $\ell_2$ sense.
The goal of the inference process is thus to recover the parameters $\theta^*$.
%
%

When working with the discrete process $\widetilde{\mathscr F}_T$, the events $t_n^i$ of the original process are replaced with a projection on a grid $\tilde t_n^i = t_n^i + \delta_n^i$.
Here, $\delta_n^i$ is uniformly distributed on $[-\Delta/2, \Delta/2]$.
We consider the discrete FaDIn estimator $\widehat \theta_\Delta$ defined as $\widehat \theta_\Delta = \argmin_\theta \mathcal L_{\mathcal G}(\theta)$.
We can upper-bound the error incurred by $\widehat \theta_\Delta$ by the decomposition:
\begin{equation}
    \norme{\widehat \theta_{\Delta} - \theta ^*}_2
        \le \underbrace{\norme{\widehat \theta_c - \theta ^*}_2}_{(*)}
                + \underbrace{\norme{ \widehat \theta_{\Delta} - \widehat \theta_c }_2}_{(**)},
\end{equation}
where $\widehat\theta_c = \arg\min_\theta \mathcal L\pars{\theta}$ is the reference estimator for $\theta^*$ based on the standard $\ell_2$ estimator for continuous point processes.
This decomposition involves the statistical error $(*)$ and the bias error $(**)$ induced by the discretization.
The statistical term measures how far the parameters obtained by minimizing the $\ell_2$ continuous loss having access to a finite amount of data are from the true ones.
In contrast, the term $(**)$ represents the discretization bias induced by minimizing the discrete loss\autoref{eq:l2discret} instead of the continuous one\autoref{eq:l2continuous}.
In the following proposition, we focus on the discretization error $(**)$, which is related to the computational trade-off offered by our method and not on the statistical error of the continuous $\ell_2$ estimator $(**)$.
Our work showcases that this disregarded estimator can be efficiently computed, and we hope it will promote research to describe its asymptotic behavior. We now study the perturbation of the loss due to discretization.

\begin{restatable}{proposition}{perturbLoss}
\label{prop:bias}
Let $\mathscr F_T$ and $\widetilde{\mathscr F}_T$ be respectively a MHP process and its discretized version on a grid $\mathcal{G}$ with stepsize $\Delta$. Assume that the intensity function of $\mathscr F_T$  possesses  continuously differentiable finite support kernels  on $[0, W]$. Thus, assuming  $\Delta < \min_{t_n^i, t_m^j \in \mathscr F_T} |t_n^i -t_m^j|$, for any $i \in \intervalleEntier{1}{p}$, it holds:
\begin{equation*}
        \widetilde \lambda_i[s] = \lambda_i(s\Delta) - \sum_{j=1}^p \sum_{t_m^{j} \in \mathscr{F}_{s\Delta}^j} \hspace{-0.2cm}\delta_m^j \frac{\partial \phi_{ij}}{\partial t}(s\Delta - t_m^j; \theta)
    + O(\Delta^2),
\end{equation*}
and 
\begin{align*}
        \LG{}(\theta)& =  \frac{2}{N_T}
    \sum_{i,j} \sum_{ \substack{t_n^{i} \in \mathscr{F}_{T}^i \\ t_m^{j} \in \mathscr{F}_{s\Delta}^j}}
    (\delta_m^j - \delta_n^i) \frac{\partial \phi_{ij}}{\partial t}(t_n^i - t_m^j; \theta) \\& \quad + \mathcal L(\theta) + \Delta . h(\theta) + O(\Delta^2)  .
\end{align*}
\end{restatable}

The technical proof is deferred to \autoref{app:proof_bias} in the Appendix. The first result is a direct application of the Taylor expansion of the intensity for the kernels.
For the loss, the first perturbation term $\Delta .h(\theta)$ comes from approximating the integral with a finite Euler sum~\citep{TASAKI2009477} while the second one derives from the perturbation of the intensity.
This proposition shows that, as the discretization step $\Delta$ goes to 0, the perturbed intensity and $\ell_2$ loss are good estimates of their continuous counterpart. We now quantify the discretization error $(**)$ as $\Delta$ goes to 0.

\begin{restatable}{proposition}{biasSpeed}
\label{prop2}
We consider the same assumption as in \autoref{prop:bias}.
Then, if the estimators $\widehat\theta_c = \arg\min_\theta \mathcal L(\theta)$ and $\widehat\theta_\Delta = \arg\min_\theta \LG(\theta)$
are uniquely defined, $\widehat\theta_\Delta$ converges to $\widehat\theta_c$ as $\Delta \to 0$. Moreover, if $\mathcal L$ is $C^2$ and its hessian $\nabla^2\mathcal L\pars{\widehat\theta_c}$ is positive definite with $\varepsilon > 0$ its smallest eigenvalue, then
$
    \norme{\widehat \theta_{\Delta} - \widehat \theta_c }_2 \le \frac\Delta\varepsilon g\pars{\widehat\theta_\Delta}
$
, with $g\pars{\widehat\theta_\Delta} = O(1)$.
\end{restatable}

This proposition shows that asymptotically on $\Delta$, the estimator $\widehat\theta_\Delta$ is equivalent to $\widehat\theta_c$.
It also shows that the discrete estimator converges to the continuous one at the same speed as $\Delta$ decreases.
This is confirmed experimentally by results shown in \autoref{fig:app:emp_bias} in the Appendix.
Thus, one would need to select $\Delta$ so that the discretization error is small compared to the statistical one.
Notice that assumptions from \autoref{prop2} are not too restrictive. Indeed, they require the existence of a unique minimizer of $\mathcal L$, $\LG$ and $\mathcal L$.
Moreover, if $\mathcal L$ is $C^2$ in $\widehat \theta_c$, the previous hypothesis also implies the strong local convexity at this point.





\section{Numerical experiments}\label{sec:num}

We present various synthetic data experiments to support the advantages of the proposed approach. To begin, we investigate the bias induced by the discretization in  \autoref{subsec:consistency}. Afterwards, the statistical and computational efficiency of FaDIn is highlighted through a benchmark with popular non-parametric approaches \autoref{subsec:bench}. Due to the space limitation, sensitivity analysis regarding the parameter $W$ and additional non-parametric comparisons are provided in Appendices~\ref{sec:app:w} and \ref{subsec:nonparam}, respectively.
\begin{figure*}[!h]
    \centering
    \includegraphics[width=0.42\linewidth]{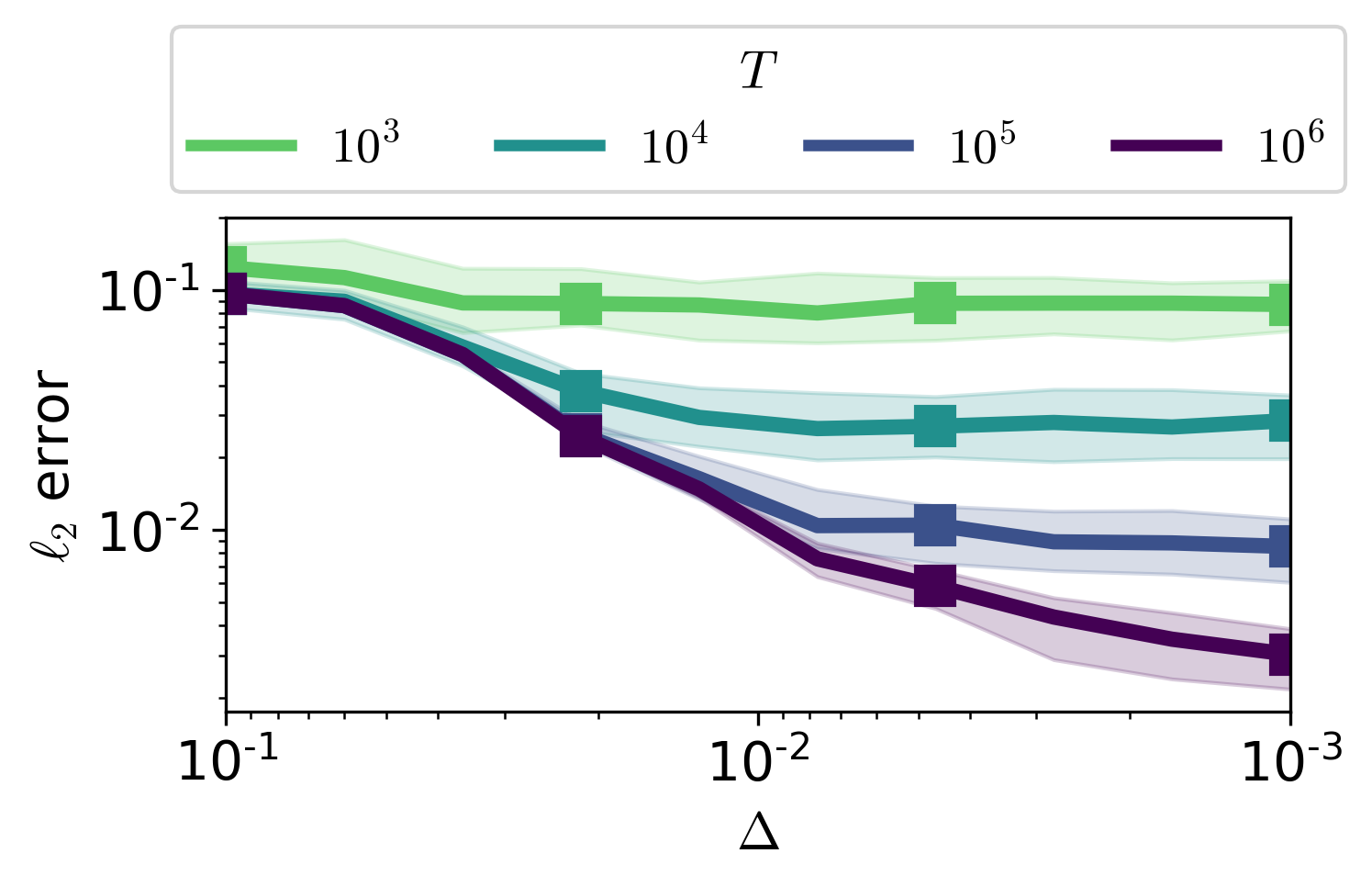}
    \includegraphics[width=0.42\linewidth]{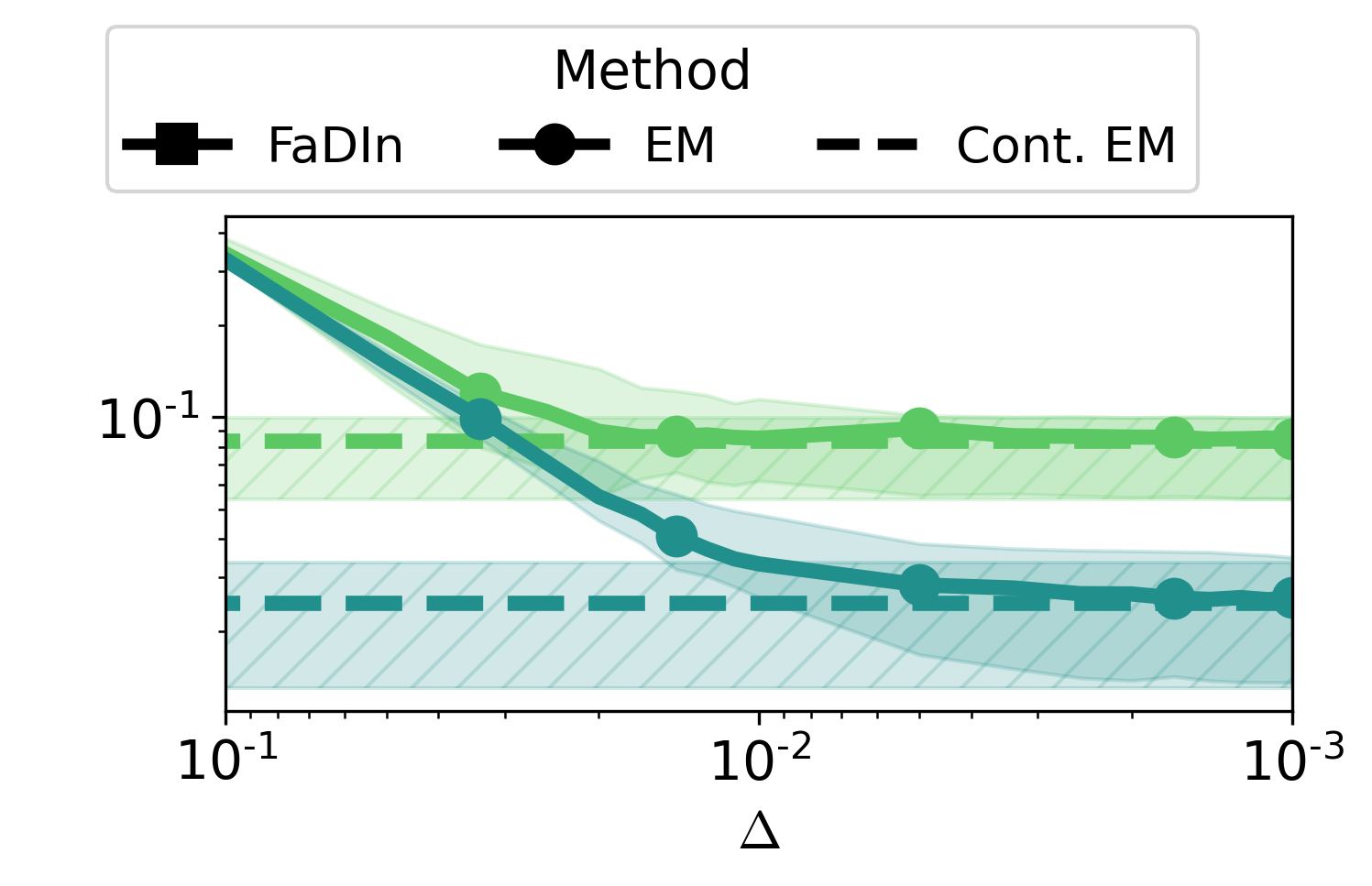}
    \vskip-1em
    \caption{Median and interquartile error bar of the $\ell_2$ norm between true parameters and parameter estimates computed with FaDIn (left) and with EM algorithm (right), continuously and discretely, w.r.t. the stepsize of the grid $\Delta$.
    }
    \label{fig:consistency}
\end{figure*}
\subsection{Consistency of Discretization}\label{subsec:consistency}

In order to study the estimation bias due to discretization, we run two experiments and report the results in \autoref{fig:consistency} (details and further experiments are presented in \autoref{subsec:app:em} and \autoref{subsec:app:fadin} in the Appendix).

The general paradigm is a one-dimensional TPP with intensity parametrized as in\autoref{eq:intensity_mhp} with a Truncated Gaussian kernel of mean $m \in \bbR$ and standard deviation $\sigma > 0$, with fixed support $\intervalleFF{0}{W} \subset \bbR^+$, $W>0$.
It corresponds to $\phi(\cdot) = \alpha \kappa(\cdot), \alpha \geq 0$ with
\begin{equation*}
    \kappa(\cdot) \coloneqq \kappa\pars{\cdot ; m, \sigma, W} = \frac{1}{\sigma} \frac{f\left(\frac{\cdot-m}{\sigma}\right)}{F\left(\frac{W-m}{\sigma}\right)-F\left(\frac{-m}{\sigma}\right)} \1[0\leq \cdot \leq W]
     ,
\end{equation*}
where $f$ (resp. $F$) is the probability density function (resp. cumulative distribution function) of the standard normal distribution.
Hence, the parameters to estimate are $\mu$ and $\eta= (\alpha, m, \sigma)$.

In both experiments, for multiple process length $T$, the discrete estimates are computed for varying grid stepsize $\Delta$, from $10^{-1}$ to $10^{-3}$. The parameter $W$ is set to 1. The $\ell_2$ norm of the difference between estimates and the true parameter values  -- the ones used for data simulation -- is computed and reported.
We first computed the parameter estimates with our FaDIn method for $T\in \{10^3, 10^5, 10^4, 10^6\}$, for 100 simulations each time.
Second, since we wish to separate discretization bias from statistical bias, we compute the estimates with an EM algorithm, both continuously and discretely, and that for 50 random data simulations.


One can observe that the $\ell_2$ errors between discrete estimates and true parameters tend towards zero as $T$ increases.
For $T$ fixed, one can see plateaus starting for stepsize values that are not particularly small, indicating that the discretization bias is limited.
The second experiment with the EM algorithm shows that when  plateau is reached, it corresponds to some statistical error.
In other words, even for a reasonably coarse stepsize, the bias induced by the discretization is slight compared to the statistical error.

\subsection{Statistical and computational efficiency of FaDIn}
\label{subsec:bench}

We compare FaDIn with non-parametric  and parametric methods by assessing  approaches' statistical and computational efficiency. To learn the non-parametric kernel, we select various existing methods. The first benchmarked method uses histogram kernels and relies on the EM algorithm, provided in \citet{zhou2013learning} and implemented in the \texttt{tick} library \citep{ticklibrary}. The kernel is set with one basis function. The three other approaches involve a linear combination of pre-defined raised cosine functions as non-parametric kernels. The inference is made either by stochastic gradient descent algorithm (Non-param SGD; \citealp{pmlr-v32-linderman14}) or by Bayesian approaches such as Gibbs sampling (Gibbs) or Variational Inference (VB) from~\citet{linderman2015scalable}. These algorithms are implemented in the \texttt{pyhawkes} library\footnote{\url{https://github.com/slinderman/pyhawkes}}.  In the following experiments, we set the number of basis to five for each method. The parametric approach we compare with is the  Neural Hawkes Process (NeuralHawkes; \citealp{mei2017neural}) where authors represents the intensity function by a LSTM module. The latter is calculated on a GPU.
\begin{figure*}[!h]
    \centering
    \begin{tabular}{cc}
        \multicolumn{2}{c}{\hspace{0.4cm}\includegraphics[scale=0.45]{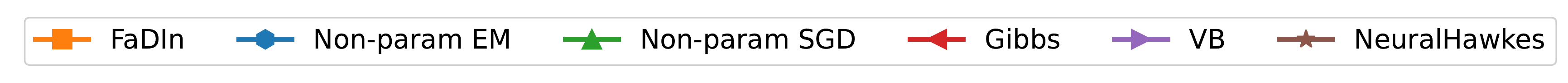}} \\
        \hspace{1.7cm} Estimation error & \hspace{0.9cm}Computation time \\
         \includegraphics[scale=0.52, trim=-2cm 0 0 0]{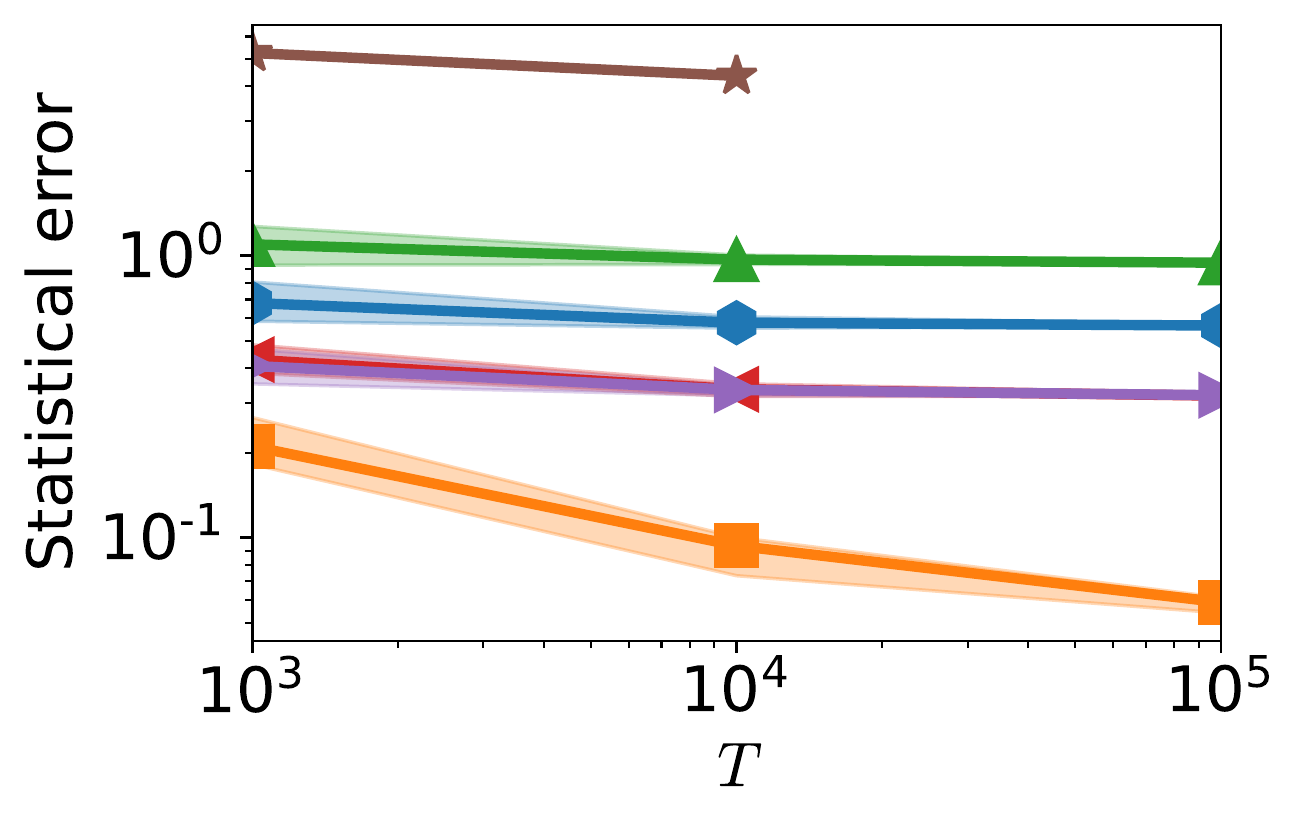} & \includegraphics[scale=0.52, trim=0cm 0 0 0]{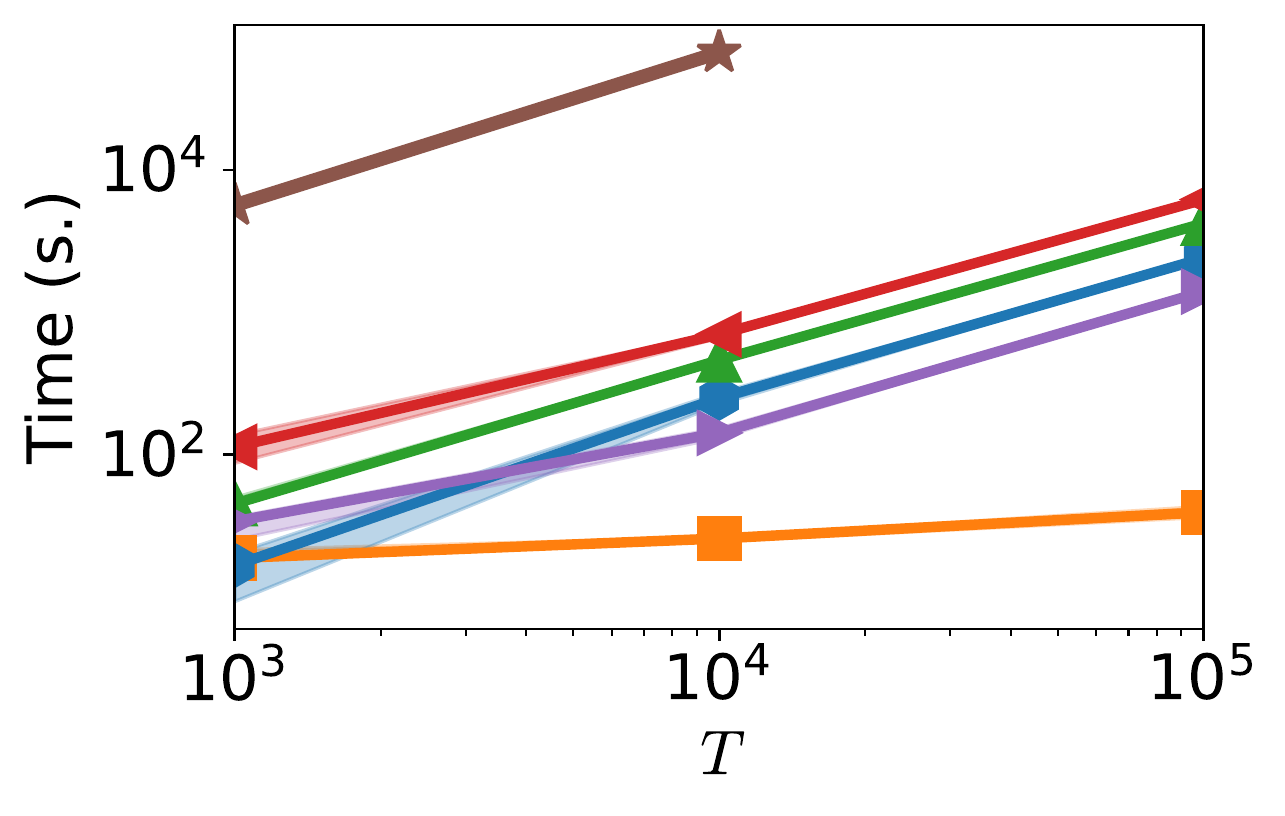}
    \end{tabular}
    \vspace{-0.6cm}
    \caption{Comparison of the statistical and computational efficiency of FaDIn with five benchmarked methods. The averaged (over ten runs) statistical error on the intensity function (left) and the computational time (right) are computed regarding the time $T$ (and thus the number of events).}
    \label{fig:benchmark}
\end{figure*}
The experiment is conducted as follows. We simulate a two-dimensional Hawkes process (repeated ten times) using the \texttt{tick} library with baseline $\boldsymbol{\mu}=[0.1, 0.2]$ and Raised Cosine kernels:
\begin{equation*}
   \phi_{i,j}(\cdot)=\alpha_{i,j} \left[{1 + \cos \pars{\frac{\cdot - u_{i,j}}{\sigma_{i,j}}\pi - \pi}} \right] , (i,j)\in \{1,2\}^2
\end{equation*}
on the support $[u_{i,j},  u_{i,j}+2\sigma_{i,j}]$ and zero outside with parameters $\boldsymbol{\alpha}=\begin{bmatrix}
1.5 & 0.1 \\
0.1 & 1.5
\end{bmatrix}$, $\mathbf{u}=\begin{bmatrix}
0.1 & 0.3 \\
0.3 & 0.3
\end{bmatrix}$  and ${\boldsymbol{\sigma}=\begin{bmatrix}
0.3 & 0.25\\
0.3 & 0.3
\end{bmatrix}}$. 
Further, we infer the intensity function of these underlying Hawkes processes using FaDIn and the four previously mentioned methods setting $\Delta=0.01$ for all these discrete approaches. The parameter $W$ of FaDIn is set to 1. This experiment is repeated for varying values of $T\in \{10^3, 10^4, 10^5 \}$. The averaged (over the ten runs) normalized $\ell_1$ error on the intensity (evaluated on the same discrete grid), as well as the associated computation time, are reported in \autoref{fig:benchmark}. Due to the high computational times of NeuralHawkes, this approach is performed once and is not applied for $T=10^5$.

From a statistical perspective, we can observe the advantages of FaDIn inference for varying $T$ over the benchmarked methods. It is worth noting that this result is expected by a parametric approach when the used kernel belongs to the same family as the one with which events have been simulated. Also, only one (long) sequence of data has been used, explaining the poor statistical results of the Neural Hawkes, which is efficient on many repetitions of short sequences due to the massive amount of parameters to infer. From a computational perspective, FaDIn is very efficient compared to benchmarked approaches. Indeed, it scales very well with an increasing time $T$ and then with a growing number of events. In contrast, other methods depend on the number of events and scale linearly with the time $T$. 




\section{Application to MEG data}

The response's latency related to a stimulus has been identified as a Biomarker of ageing \citep{price2017age} and many diseases such as epilepsy \citep{kannathal2005characterization}, Alzheimer's \citep{dauwels2010diagnosis}, Parkinson's \citep{tanaka2000event} or multiple sclerosis \citep{gil1993event}.
Therefore, obtaining information on such a feature after auditory or visual stimuli is critical to characterize and eventually detect the presence of a specific disease for a given subject.
FaDIn allows fitting a statistical model on this latency by inferring a model on the latency of these responses through Hawkes processes kernels (see \autoref{sec:app:addexpemeg}).
This approach characterizes the delays' distribution more finely compared to the latency estimates.

Electrophysiology signals recorded with M/EEG contain recurring prototypical waveforms that can be related to human behavior~\citep{shin-etal:2017}. Convolutional Dictionary Learning (CDL; \citealt{Jas2017}) is an unsupervised method to efficiently extract such patterns and study them in a quantitative way.
With CDL, multivariate neural signals are represented by a set of spatio-temporal patterns, called \textit{atoms}, with their respective onsets, called \textit{activations}.
Here, we make use of the   \href{https://alphacsc.github.io/}{\texttt{alphacsc}} software for CDL with rank-1 constraint~\citep{dupre2018multivariate}, as it includes physical priors for the patterns to recover, namely that the spatial propagation of the signal from the brain to sensors is linear and instantaneous.
The schema in the Appendix in \autoref{fig:app:cdl_resume} presents how CDL applies to MEG recordings.

Experiments on MEG data were run on two datasets from the \texttt{MNE} Python package~\citep{gramfort2013meg, gramfort2014mne}: the \emph{sample} dataset and the somatosensory (\emph{somato}) dataset\footnote{Both available at \url{https://mne.tools/stable/overview/datasets_index.html}}.

These datasets were selected as they elicit two distinct types of event-related neural activations: evoked responses which are time-locked to the onsets of the stimulation, and induced responses which exhibit larger random jitters. The \textit{sample} dataset contains M/EEG recordings of a human subject presented with audio and visual stimuli.
\begin{figure*}[!ht]
    \centering
    \includegraphics{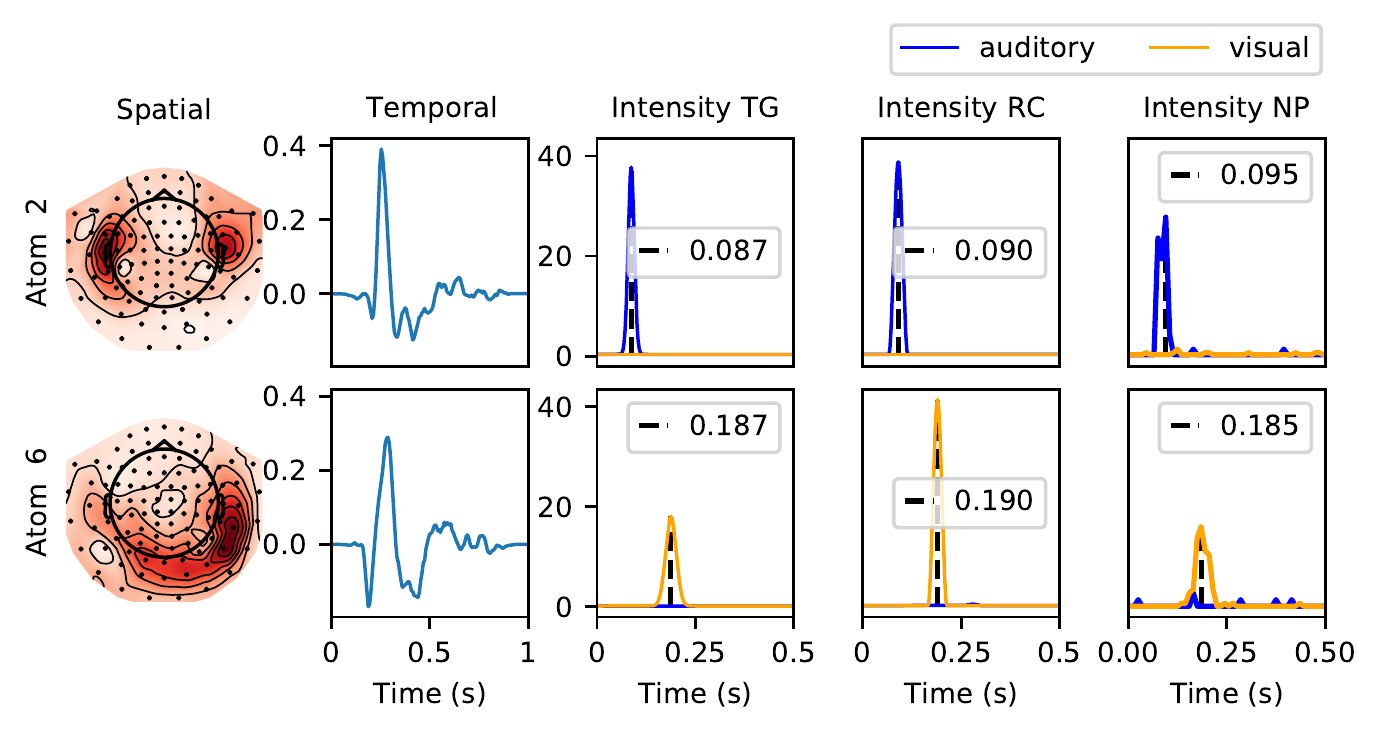}
     \vspace{-10pt}
    \caption{Spatial and temporal patterns of 2 atoms from \emph{sample} dataset, and their respective estimated intensity functions after a stimulus (cue at time = 0 s), for auditory and visual stimuli with non-parametric (NP), Truncated Gaussian (TG) and Raised Cosine (RC) kernels.
    }
    \label{fig:fig_sample}
\end{figure*}
This experiment presents checkerboard patterns to the subject in the left and right visual field, interspersed with tones to the left or right ear. The experiment lasts about \SI{4.6}{\minute}, and approximately 70 stimuli per type are presented to the subject.
%
For the \emph{somato} dataset, a human subject is scanned with MEG during \SI{15}{\minute}, while 111 stimulations of his left median nerve were made.
For both datasets, raw data are first preprocessed as done by \citet{allain2021dripp}, and CDL is then applied: 40 atoms of duration \SI{1}{\second} each are extracted on the \emph{sample} dataset, and 20 atoms of duration \SI{0.53}{\second} for the \emph{somato} dataset.
Finally, each dataset is represented by two sets of Temporal Point Processes: a set of stochastic ones representing atoms' activations, and a set of deterministic ones coding for external stimuli events.

The main goal of applying the TPP framework to such data is to characterize directly when and how each stimulus is responsible for the occurrence of neural responses, especially by estimating the distribution of latencies.
We are interested in the paradigm of Driven Point Process (DriPP; \citealt{allain2021dripp}) and for every extracted atom, its intensity function related to the corresponding stimuli is estimated using a non-parametric kernel (NP) and two kernel parametrizations: Truncated Gaussian (TG) and Raised Cosine (RC). 
Results on the \emph{sample} (resp. \emph{somato}) dataset are presented in~\autoref{fig:fig_sample} (resp. \autoref{fig:app:fig_somato} in the Appendix), where only the kernel related to each type of stimulus is plotted, for the sake of clarity. See Appendix \ref{sec:app:addexpemeg}.

Results show that all three kernels agree on a peak latency around \SI{90}{\milli\second} for the auditory condition and \SI{190}{\milli\second} for the visual condition.
Due to the limited number of events, one can observe that the non-parametric kernel estimated is less smooth, with spurious peaks later in the interval.
Overall, these results on real MEG data demonstrate that our approach with a RC kernel parametrization can recover correct latency estimates even with the discretization of stepsize 0.02.
Furthermore, the usage of RC allows us to have sharper peaks in intensity compared to TG, enforcing the link between the external stimulus and the atom's activation.
This difference mainly comes from the fact that RC does not need pre-determined support.
This advantage is even more pronounced in the case of induced responses, such as in the \emph{somato} dataset (see \autoref{fig:app:fig_somato}), where the range of possible latency values is more difficult to determine beforehand.

\section{Discussion}

This work proposed an efficient approach to infer general parametric kernels for Multivariate Hawkes processes.
Our method makes the use of parametric kernels computationally tractable, beyond exponential kernels.
The development of FaDIn is based on three key features: ($i$) finite-support kernels, ($ii$) timeline discretization, and ($iii$) precomputations reducing the computational cost of the gradients.
These allow for a computationally efficient gradient-based approach, improving state-of-the-art methods while providing flexible use of kernels well-fitted to the considered applications.
Moreover, this work shows that the bias induced by the discretization is negligible, both theoretically and numerically.
By allowing the use of a general parametric kernel in Hawkes processes, this contribution opens new possibilities for many applications.
This is the case with M/EEG data, where estimating information about the rate and latency of occurrences of brain signal patterns is at the core of neuroscience questions.
%
Therefore, FaDIn makes it possible to use a Raised Cosine kernel, allowing for efficient retrieval of these parameters.

\bibliography{main}
\bibliographystyle{icml2023}

\appendix
\onecolumn
\newpage

\def\thetable{\thesection.\arabic{table}}
\counterwithin{table}{section}
\counterwithin{figure}{section}

\section{Technical details}\label{sec:app:proofs}
This part presents proofs of the theoretical results proposed in the core paper.

\subsection{Proof of Proposition~\ref{prop:bias}}
\label{app:proof_bias}

Recall that by definition,
\begin{align}
    \lambda_i(s\Delta) &= \mu_i + \sum_{j=1}^p \sum_{t_m^j\in\mathscr F^j_{s\Delta}}\phi_{ij}(s\Delta- t_m^j) , \nonumber
\end{align}
and
\begin{align}
    \tilde{\lambda}_i[s] &= \mu_i + \sum_{j=1}^{p} \sum_{\tilde{t}_m^{j} \in \widetilde{\mathscr{F}}_{s\Delta}^j} \phi_{ij} (s\Delta - \tilde{t}_m^j)\nonumber \\
    &= \mu_i + \sum_{j=1}^{p} \sum_{{t}_m^{j} \in {\mathscr{F}}_{s\Delta}^j} \phi_{ij} (s\Delta - {t}_m^j - \delta_m^j)   ,\label{bias:l3}
\end{align}
where \autoref{bias:l3} is a consequence of hypothesis $\Delta < \min_{t_n^i, t_m^j \in \mathscr F_T} |t_n^i -t_m^j|$ which ensures that no event collapses on the same bin of the grid and that $\widetilde{\mathscr F}_{s\Delta}^j = \mathscr F_{s\Delta}^j$.
Note that this hypothesis also implies that the intensity function is smooth for all points on the grid $\mathcal G$. Applying the first-order Taylor expansion to the kernels $\phi_{ij}$ in $s\Delta - t_m^j$ and bounding the perturbation $\delta_n^i$ by $\Delta$ yields the first result of the proposition.

For the perturbation of the loss $\LG{}$, we have:
\begin{align*}
    \mathcal{L}_{\mathcal G}\pars{\theta, \widetilde{\mathscr{F}}_T}
        & =  \frac{1}{N_T}\sum_{i=1}^{p}  \pars{\Delta\sum_{s\in \intervalleEntier{0}{G}}\pars{\tilde{\lambda}_{i}[s]}^2
            - 2\sum_{\tilde{t}_n^i \in \widetilde{\mathscr{F}}_T^i}\tilde{\lambda}_{i}\bracks{\frac{\tilde{t}_n^{i}}{\Delta}}}   \\
        & =  \mathcal L(\theta) + \frac{1}{N_T}\sum_{i=1}^{p}\Bigg( \underbrace{\Delta\sum_{s=0}^G\tilde{\lambda}_{i}[s]^2 - \int_0^T \lambda_i(t)^2 \mathrm{d}t}_{(*)}
        - 2\underbrace{\sum_{t_n^{i} \in \mathscr{F}_T^i}\widetilde{\lambda}_{i}\Bigg[\frac{\widetilde t_n^i}{\Delta}\Bigg] - \lambda_{i}\pars{t_n^{i}}}_{(**)}\Bigg)   .
\end{align*}

The first term $(*)$ is the error of a Riemann approximation of the integral.
Theorem 1.2 in \citet{TASAKI2009477} shows that asymptotically with $\Delta\to 0$,
\begin{equation}\label{term1}
   \Delta\sum_{s=0}^G\tilde{\lambda}_{i}[s]^2 - \int_0^T \lambda_i(t)^2 \mathrm{d}t = \Delta . h_i(\theta) + O(\Delta^2) ,
\end{equation}
where $h_i(\theta) = \frac12\Big(\int_0^T |\lambda_i(t; \theta)\frac{\partial\lambda_i}{\partial t}(t; \theta)|^{1/2}\mathrm{d}t\Big)^2$ and we denote $h(\theta) = \frac{1}{N_T}\sum_{i=1}^{p} h_i(\theta)$.

For the second term $(**)$, we re-use the expression from\autoref{bias:l3} but use a Taylor expansion in $t_n^i - t_m^j$.
The perturbation becomes $\delta^j_m - \delta^i_n$,
\begin{equation}\label{term2}
    \sum_{t_n^{i} \in \mathscr{F}_T^i}\widetilde{\lambda}_{i}\Bigg[\frac{\widetilde t_n^i}{\Delta}\Bigg] - \lambda_{i}\pars{t_n^{i}}
    = \sum_{t_n^{i} \in \mathscr{F}_T^i}\pars{\delta_n^i - \delta_m^j}
    \frac{\partial\phi_{ij}}{\partial t}\pars{t_n^i - t_m^j; \theta} + O\pars{\Delta^2}   .
\end{equation}
Summing \autoref{term1} and \autoref{term2} concludes the proof.

\subsection{Proof of Proposition~\ref{prop2}}

    We consider the two estimators $\widehat \theta_\Delta = \argmin_\theta \LG{}(\theta)$ and $\widehat \theta_c = \argmin_\theta\mathcal L(\theta)$.
    With the loss approximation from \autoref{prop:bias}, we have a pointwise convergence of $\LG{}(\theta)$ towards $\mathcal L(\theta)$ for all $\theta \in \Theta$ as $\Delta$ goes to 0.
    By continuity of $\LG$, we have that the limit of $\widehat\theta_\Delta$ when $\Delta$ goes to 0 exists and is equal to $\widehat \theta_c$.
    This proves that the discretized estimator converges to the continuous one as $\Delta$ decreases.
    
    We now characterize its asymptotic speed of convergence.
    The KKT conditions impose that:
    \begin{equation}
        \label{eq:app:kkt}
        \nabla\LG{}\pars{\widehat\theta_\Delta} = 0 \qquad\text{and}\qquad
        \nabla\mathcal L\pars{\widehat\theta_c} = 0 .
    \end{equation}
    
    Using the approximation from \autoref{prop:bias}, one gets in the limit of small $\Delta$:
    \begin{align*}
        \nabla\LG{}\pars{\widehat\theta_\Delta} = \nabla \mathcal L\pars{\widehat\theta_\Delta}
        + \Delta . \frac{\partial h}{\partial \theta}\pars{\widehat\theta_\Delta} + O\pars{\Delta^2}+ \frac2{N_T} \sum_{i=1}^p \sum_{t_n^{i} \in \widetilde{\mathscr{F}}_{T}^i}
        \sum_{j=1}^p \sum_{t_m^{j} \in \widetilde{\mathscr{F}}_{s\Delta}^j}
        \pars{\delta_m^j - \delta_n^i}\frac{\partial^2 \phi_{ij}}{\partial t\partial \theta}\pars{t_n^i - t_m^j; \widehat \theta_\Delta}.
    \end{align*}
    Combining this with \autoref{eq:app:kkt}, we get:
    \begin{equation*}
        \nabla \mathcal L\pars{\widehat\theta_\Delta} = -\Delta . \frac{\partial h}{\partial \theta}\pars{\widehat\theta_\Delta} + \frac2{N_T} \sum_{i=1}^p \sum_{t_n^{i} \in \widetilde{\mathscr{F}}_{T}^i}
        \sum_{j=1}^p \sum_{t_m^{j} \in \widetilde{\mathscr{F}}_{s\Delta}^j}
        \pars{\delta_n^i - \delta_m^j}\frac{\partial^2 \phi_{ij}}{\partial t\partial \theta}\pars{t_n^i - t_m^j; \widehat \theta_\Delta}
        + O\pars{\Delta^2},
    \end{equation*}
    and

    \begin{align*}
        \norme{ \nabla \mathcal L(\widehat\theta_\Delta) - \nabla \mathcal L\pars{\widehat\theta_c} }_2
        &= \norme{-\Delta . \frac{\partial h}{\partial \theta}\pars{\widehat\theta_\Delta} + \frac2{N_T} \sum_{i,j=1}^p \sum_{t_n^{i} \in \widetilde{\mathscr{F}}_{s\Delta}^i}
         \sum_{t_m^{j} \in \widetilde{\mathscr{F}}_{s\Delta}^j}
        \pars{\delta_n^i - \delta_m^j}\frac{\partial^2 \phi_{ij}}{\partial t\partial \theta}\pars{t_n^i - t_m^j; \widehat \theta_\Delta} }_2 
        + O\pars{\Delta^2}\nonumber\\
        &\le \Delta \norme{\frac{\partial h}{\partial \theta}\pars{\widehat\theta_\Delta} + \frac2{N_T} \sum_{i,j=1}^p \sum_{t_n^{i} \in \widetilde{\mathscr{F}}_{s\Delta}^i}
         \sum_{t_m^{j} \in \widetilde{\mathscr{F}}_{s\Delta}^j}
        \frac{\partial^2 \phi_{ij}}{\partial t\partial \theta}\pars{t_n^i - t_m^j; \widehat \theta_\Delta} }_2
        + O\pars{\Delta^2}\nonumber \\
        &\le \Delta. g\pars{\widehat\theta_\Delta} ,
    \end{align*}
    where $g(\theta)$ is equal to $\|\frac{\partial h}{\partial \theta}(\widehat\theta_\Delta) + \frac2{N_T} \sum_{i,j=1}^p \sum_{t_n^{i} \in \widetilde{\mathscr{F}}_{s\Delta}^i}
         \sum_{t_m^{j} \in \widetilde{\mathscr{F}}_{s\Delta}^j}
        \frac{\partial^2 \phi_{ij}}{\partial t\partial \theta}(t_n^i - t_m^j; \widehat \theta_\Delta)\|_2 + O(\Delta)$.
    This function is a $O(1)$.
    Using the hypothesis that the hessian $\nabla^2\mathcal L(\widehat\theta_c)$ exists and is positive definite with smallest eigenvalue $\varepsilon$, we have:
    \begin{align*}
        \varepsilon \norme{\widehat\theta_\Delta - \widehat\theta_c}_2^2
            &\le\norme{ \nabla \mathcal L\pars{\widehat\theta_\Delta} - \nabla \mathcal L\pars{\widehat\theta_c}}_2^2 \\
        \text{\ie}\qquad
        \norme{\widehat\theta_\Delta - \widehat\theta_c}_2^2
            &\le \frac\Delta\varepsilon g\pars{\widehat\theta_\Delta}.
    \end{align*}
This concludes the proof.

\section{Additional Experiments}\label{sec:app:addexpe}

This section presents additional experimental results supporting the claims of the  paper. 
We compare the  $\ell_2$ loss involved in FaDIn with the popular negative Log-Likelihood in \autoref{sec:app:ll}. A sensitivity analysis of the kernel length $W$ is provided in \autoref{sec:app:w}. Additional comparisons with  popular non-parametric approaches are presented in  \autoref{subsec:nonparam}. The consistency of the discretization is supported in \autoref{subsec:app:em} and \autoref{subsec:app:fadin}.  Finally, we explain the methodology we employed on real MEG data and provide complementary results in \autoref{sec:app:addexpemeg}.

\subsection{Comparison of FaDIn with the negative log-likelihood loss}\label{sec:app:ll}

\begin{figure}[!ht]
    \centering
    \begin{tabular}{cc}
    \multicolumn{2}{c}{\includegraphics[scale=0.45]{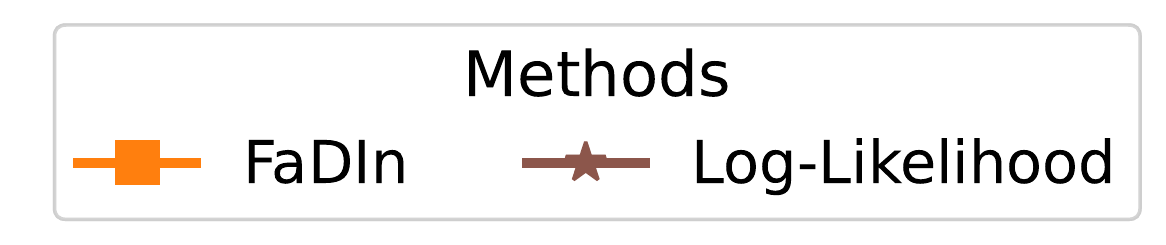}\hspace{0.2cm}\includegraphics[scale=0.45]{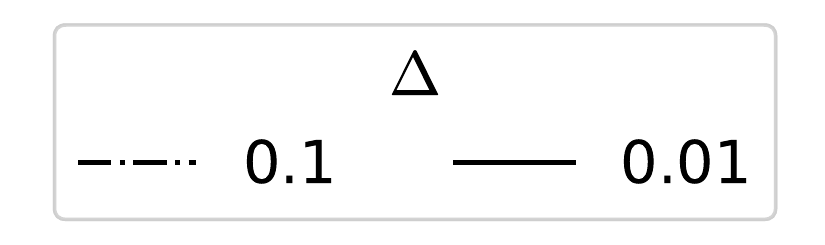}}\\
        \includegraphics[scale=0.5]{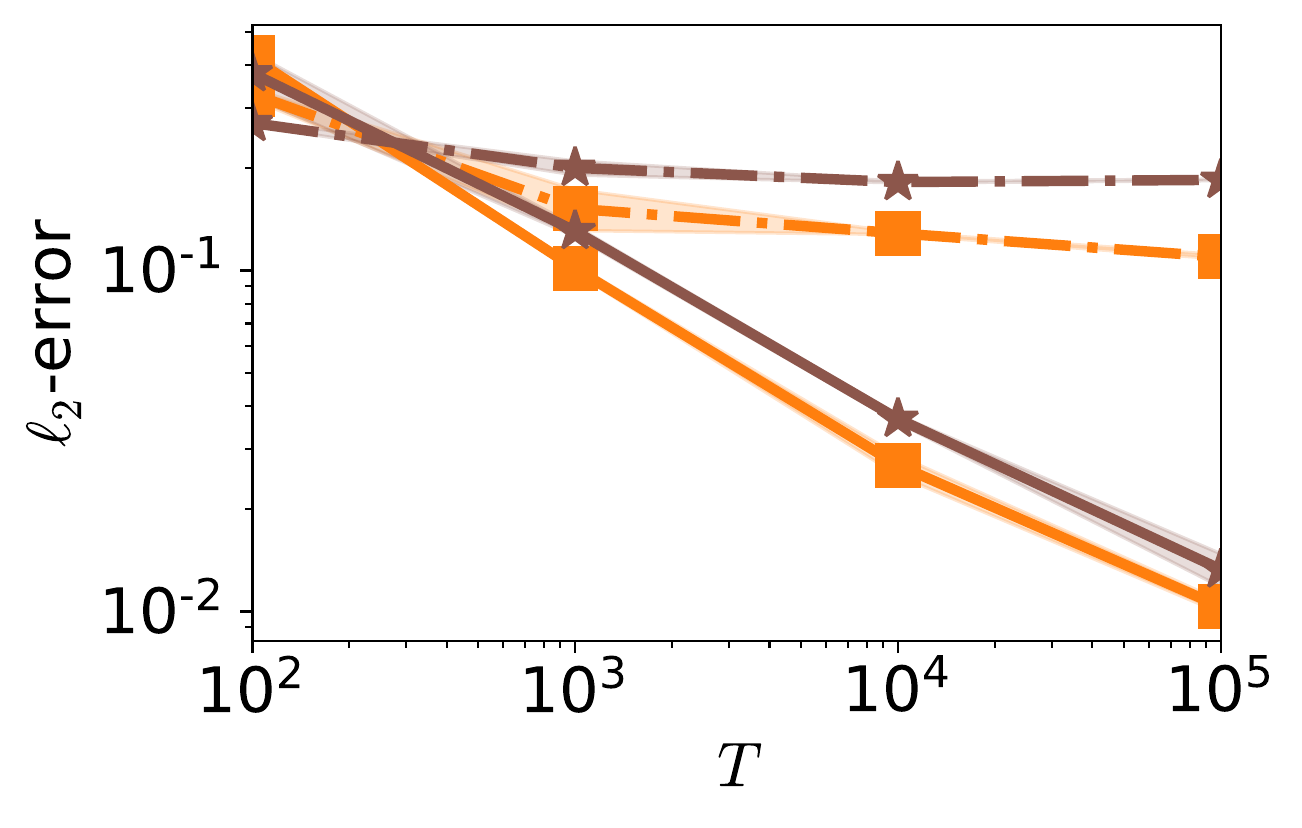} & \includegraphics[scale=0.5]{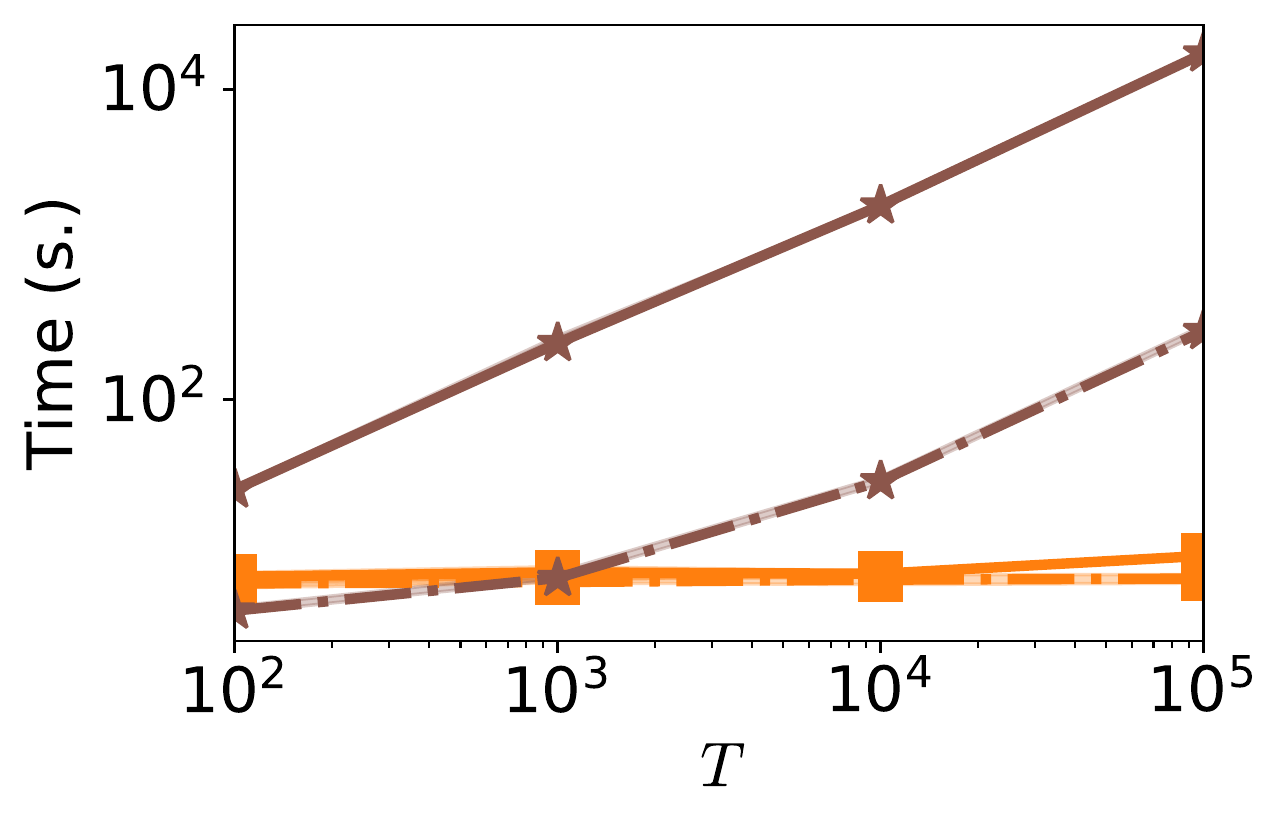} \\
                \includegraphics[scale=0.5]{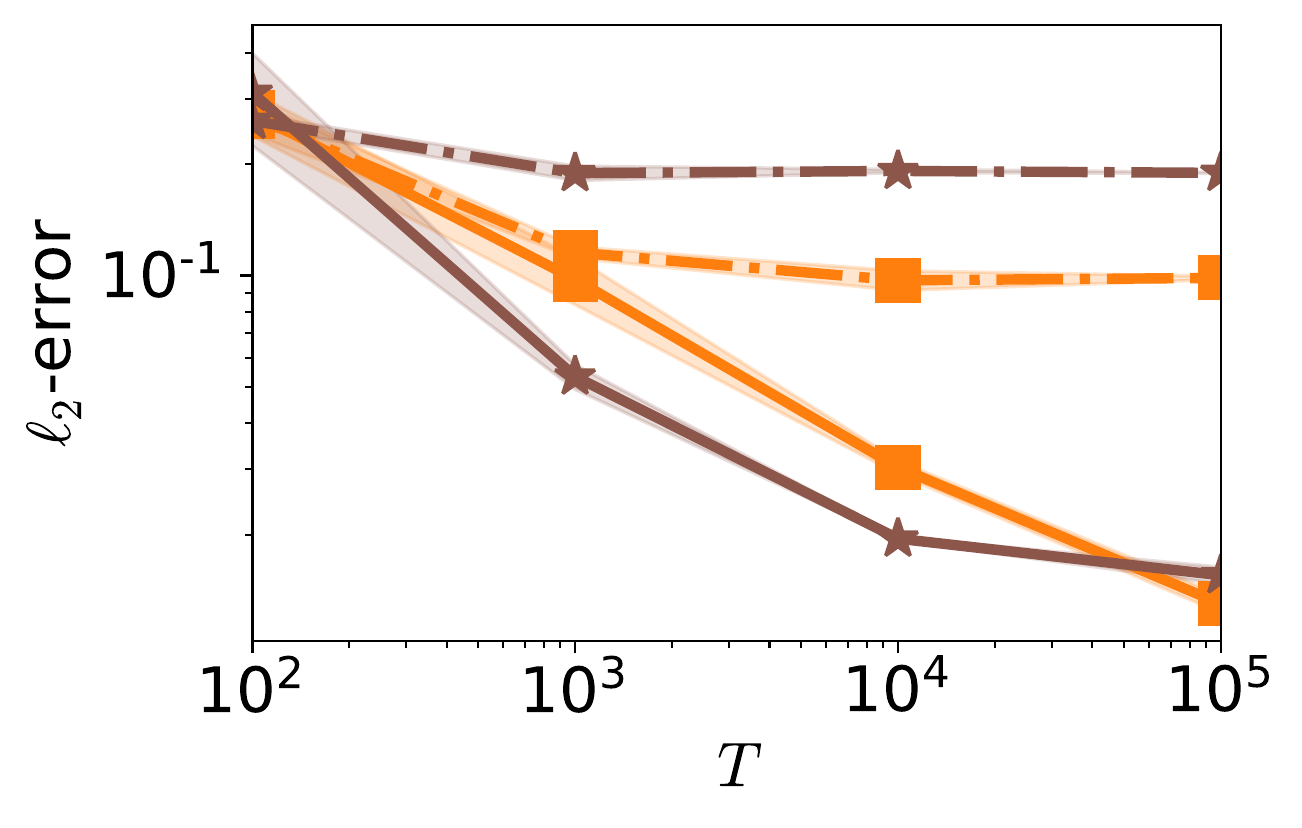} & \includegraphics[scale=0.5]{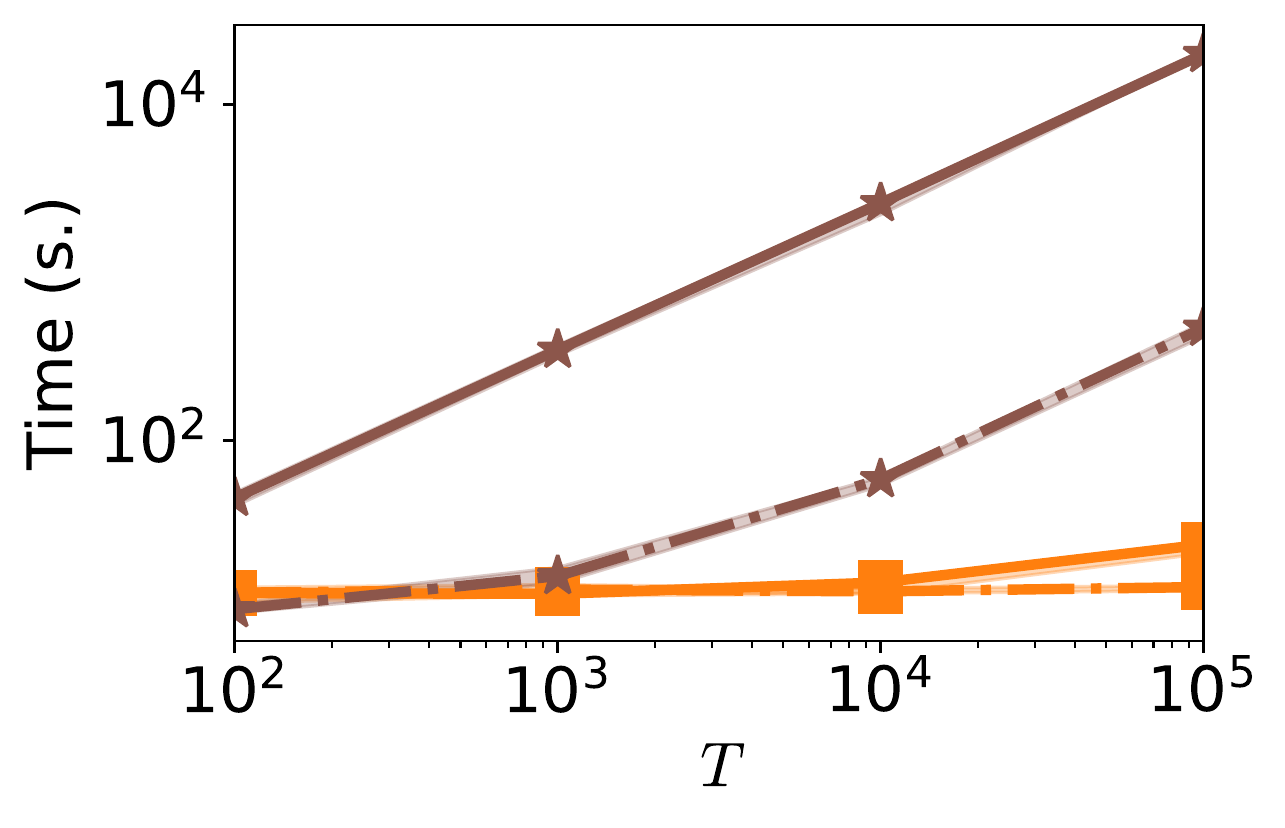}\\
                        \includegraphics[scale=0.5]{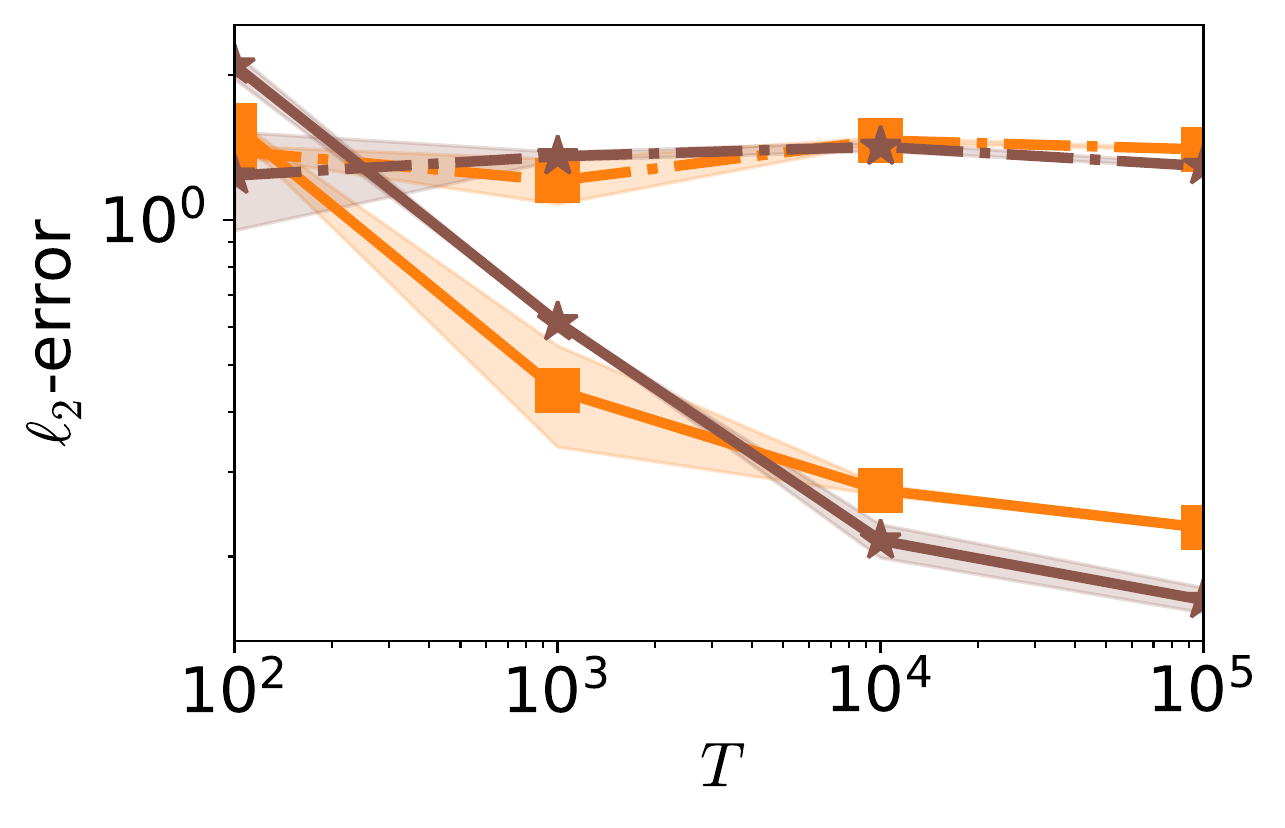} & \includegraphics[scale=0.5]{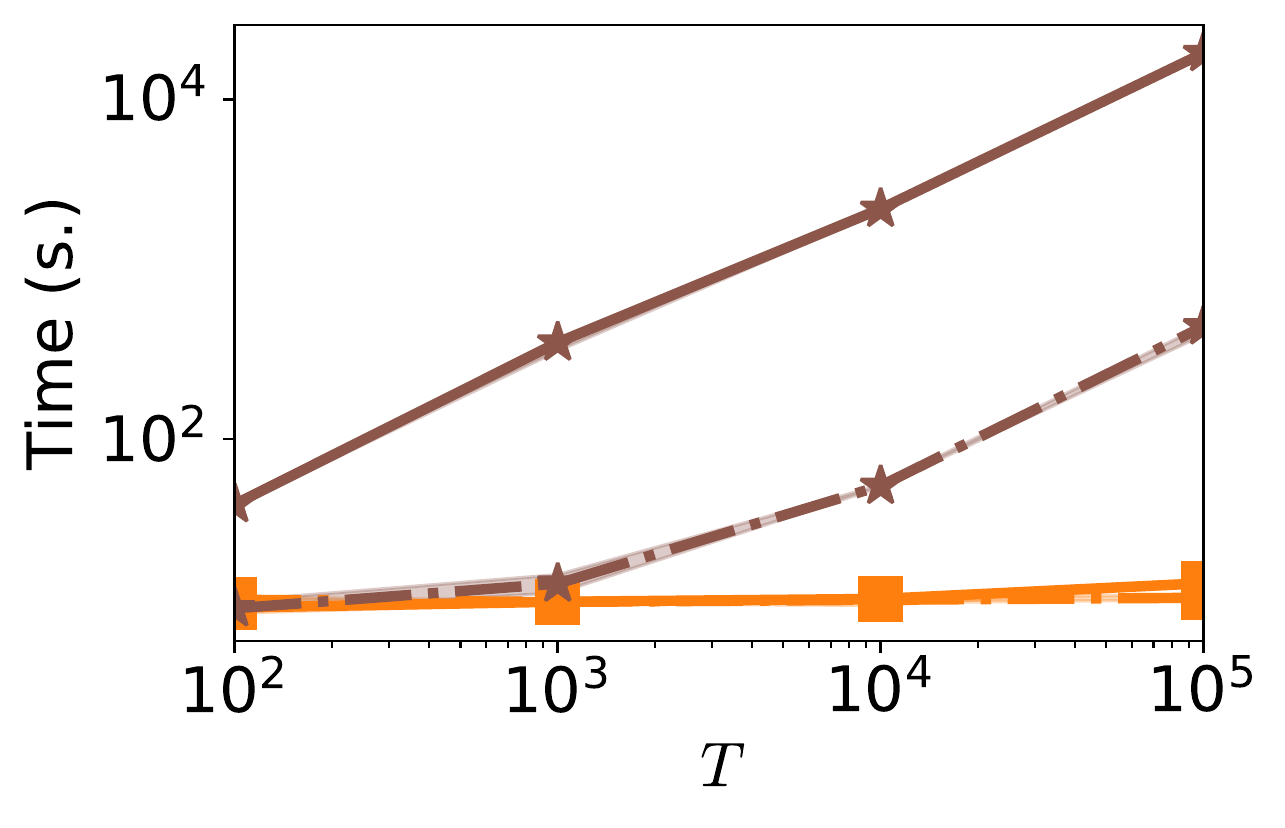}
    \end{tabular}
    \caption{Comparison of the statistical and computational efficiency of FaDIn with the Log-Likelihood loss. The averaged (over ten runs) statistical error on the intensity function (left) and the computational time (right) are computed regarding the time $T$ for the Raised Cosine (top), the Truncated Gaussian (middle) and the Truncated Exponential (bottom).}
    \label{fig:ll:rc}
\end{figure}

We compare both approaches' statistical and computational efficiency to highlight the benefit of using the  $\ell_2$ loss in FaDIn over the log-likelihood (LL).
Precisely, we compare the accuracy of the obtained parameter estimators from FaDIn and the minimization of the negative log-likelihood in the same setting as our approach (discretization and finite-support kernels).
We conduct the experiment as follows.
We place ourselves in the univariate setting for computational simplicity.
We sample a set of events in continuous time through the \texttt{tick} library.
Three sets are sampled from the kernel shapes: Raised Cosine, Truncated Gaussian, and Truncated Exponential.
The parameters are set as $\mu=0.3$, $\alpha=0.8$,  $(u,\sigma)=(0.2, 0.3)$ for the Raised Cosine, $(m,\sigma)=(0.5, 0.3)$ for the Truncated Gaussian and $\gamma=5$ for the Truncated Exponential.
We set the kernel length $W$ to 1 for each setting.
Further, we estimate the parameters of the intensity of sampled events using both FaDIn and LL approaches. The experiment is repeated ten times.
The median and $25$-$75\%$ quantiles of the statistical accuracy and the computation time are reported in \autoref{fig:ll:rc} for the three different kernels.
We can observe an equivalent accuracy of the parameter estimation for both methods along the different kernels, stepsize and number of events.
In contrast, the computational performance of FaDIn outperforms the LL approach.
Indeed, the computational time is divided by $\approx 5$ in a low data regime with $T=10^2$ and by  $\approx 1000$ when $\Delta=0.01$ and $T=10^5$.
This experiment clearly shows the advantages of using the  $\ell_2$ loss in FaDIn rather than the log-likelihood.

\subsection{Kernel length on FaDIn estimates}\label{sec:app:w}

\begin{figure}[!ht]
    \centering
    \begin{tabular}{cc}
        \multicolumn{2}{c}{\hspace{0.6cm}\includegraphics[scale=0.6]{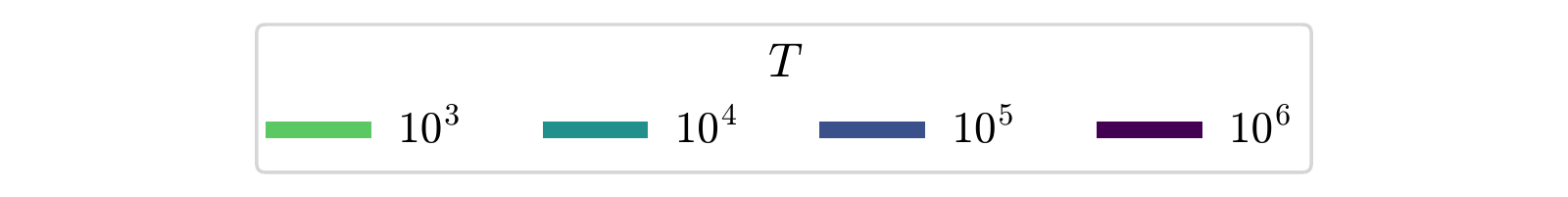}} \\
         \includegraphics[scale=0.5, trim=0cm 0 0 0]{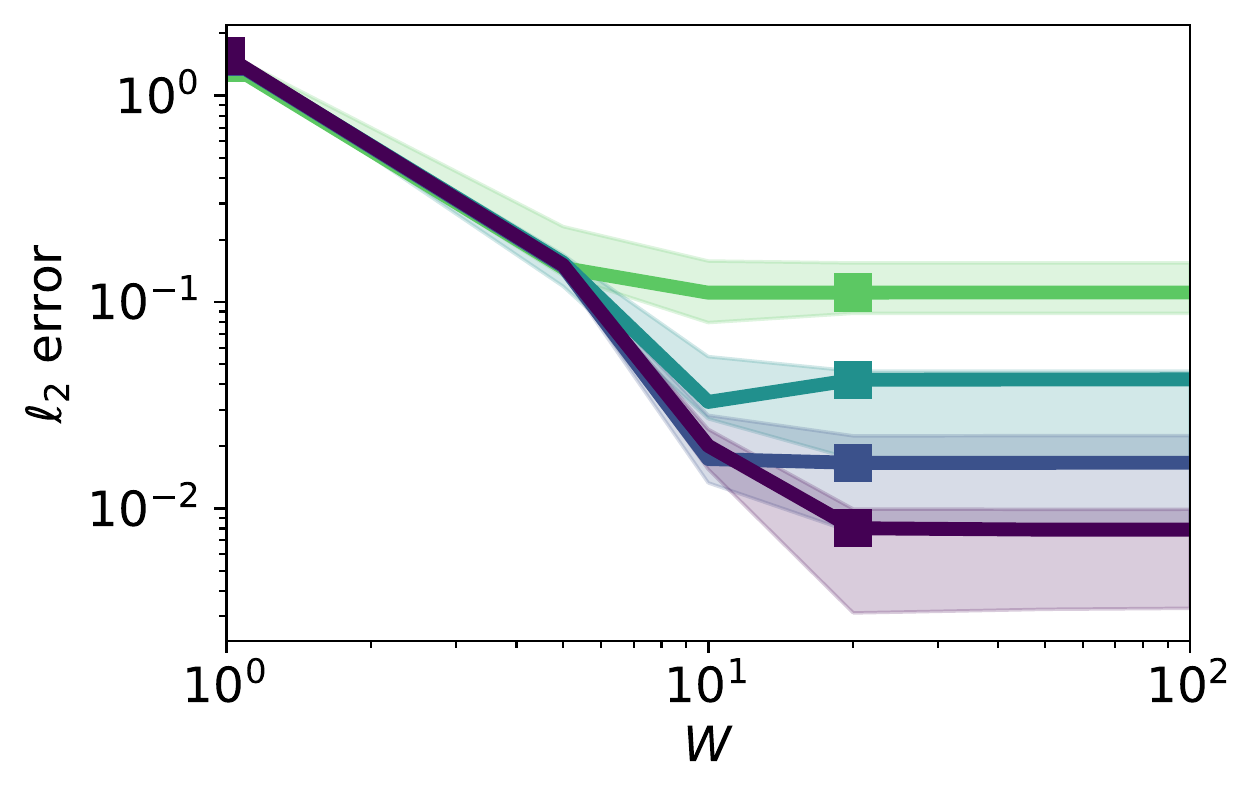} & \includegraphics[scale=0.5, trim=0cm 0 0 0]{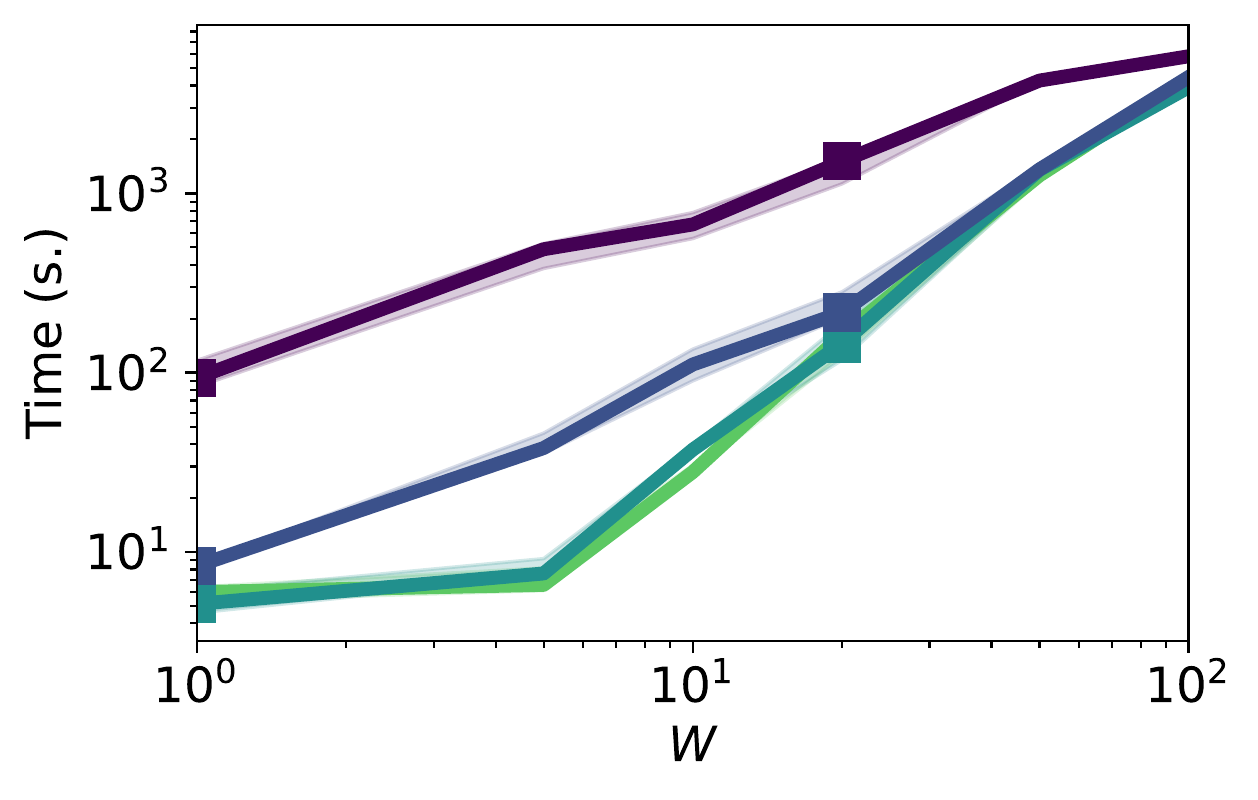}
    \end{tabular}
    \caption{Comparison of the influence of the kernel support size $W$ on the parameter estimation of FaDIn  for a Truncated Exponential kernel. The averaged (over 10 runs) statistical $\ell_2$-error (left) and computational time (right) are displayed w.r.t. the stepsize of the grid $\Delta$.}
    \label{fig:app:sensi_w}
\end{figure}

To study the estimation bias induced by the finite support kernels, 
we conduct an experiment using FaDIn with a (Truncated) Exponential kernel.
The general framework is a one-dimensional TPP with intensity parametrized as in\autoref{eq:intensity_mhp} with a Truncated Exponential kernel having a decay parameter $\gamma$, with fixed support $\intervalleFF{0}{W} \subset \bbR^+$, $W>0$. It corresponds to ${\phi(\cdot) = \alpha \kappa(\cdot)}, \alpha \geq 0$ with
\begin{equation*}
    \kappa(\cdot) \coloneqq \kappa\pars{\cdot ; \gamma, a, b} =  \frac{h\left(\cdot\right)}{H\left(b\right)-H\left(a\right)} \1[a\leq \cdot \leq b]
      ,
\end{equation*}
where here $h$ (resp. $H$) is the probability density function of parameter $\gamma$ (resp. cumulative distribution function) of the exponential distribution.
Therefore, when $W\rightarrow \infty$, this Truncated Exponential kernel converges to the standard exponential kernel, \ie $t \mapsto \alpha \gamma\exp(-\gamma t)$. The parameters to estimate are $\mu$ and $\eta= (\alpha, \gamma)$.
The experiment is conducted as follows. We simulate events (10 repetitions) from a Hawkes process with baseline $\mu=1.1$ and a standard Exponential kernel (non-truncated) with $\alpha=0.8, \; \gamma=0.5$ for varying $T\in \{10^3, 10^4, 10^5, 10^6\}$ using the \texttt{tick} Python library.
FaDIn is then computed on each of these sets of events using a Truncated Exponential kernel of length $W\in [1, 100]$ and a stepsize $\Delta=0.01$. The averaged (over ten runs) and the $25$-$75\%$ quantiles statistical $\ell_2$-error of parameters (left) and computational time (right) are displayed w.r.t. the stepsize of the grid $\Delta$ in \autoref{fig:app:sensi_w}.
On the one hand, one can observe that the $\ell_2$-error converges to a plateau once $W>10$, \ie the bias induced by the finite support length is reduced. On the other hand, the computational time increase when $W$ increases.
Interestingly, for each $T$, the computational time is close when $W$ is high enough (close to $100$).
Indeed, optimizing the loss becomes the bottleneck of FaDIn since the grid size ($G=TL+1$) only intervenes in the precomputation part. 

\subsection{Statistical and computational efficiency of FaDIn}\label{subsec:nonparam}

This part presents additional non-parametric comparisons.




\subsubsection{Qualitative Comparison with a non-parametric approach}\label{subsubsec:nonparam}

We compare FaDIn with the use of a non-parametric kernel by assessing the statistical and computational efficiency of both approaches. To learn the non-parametric kernel, we select the EM algorithm, provided in \citet{zhou2013learning} and implemented in the \texttt{tick} library \citep{ticklibrary}. The kernel is set with one basis function. In addition, we display the running time when computing gradients using PyTorch and automatic differentiation applied to the $\mathcal{L}_G$ discretized loss (\ref{eq:l2discret}).

The experiment is conducted as follows. We fix $p=1$ for simplicity, set $\mu=1.1$ and choose a Raised Cosine kernel defined by:
\begin{equation*}
   \phi(\cdot)=\alpha \left[{1 + \cos \pars{\frac{\cdot - u}{\sigma}\pi - \pi}} \right]\mathbb{I}\left\{\cdot\in[u; \; u+2\sigma]\right\}   , 
\end{equation*}
setting parameters $\alpha=0.8$,  $u=0.2$ and $\sigma=0.3$. We simulate events in a continuous time using the \texttt{tick} library \citep{ticklibrary}. FaDIn and the non-parametric kernel are optimized over $800$ iterations (with an early stopping for the EM algorithm). The RMSprop algorithm is used in FaDIn. The discretization size of the non-parametric kernel is settled as in FaDIn. This experiment is done varying $T\in \{10^3, 10^5, 10^6\}$.

\begin{figure}[!ht]
    \centering
\begin{tabular}{cc}
   \hspace{0.5cm} $T=10^3$ & \hspace{0.5cm} $T=10^6$ \\
       \includegraphics[width=0.45\linewidth]{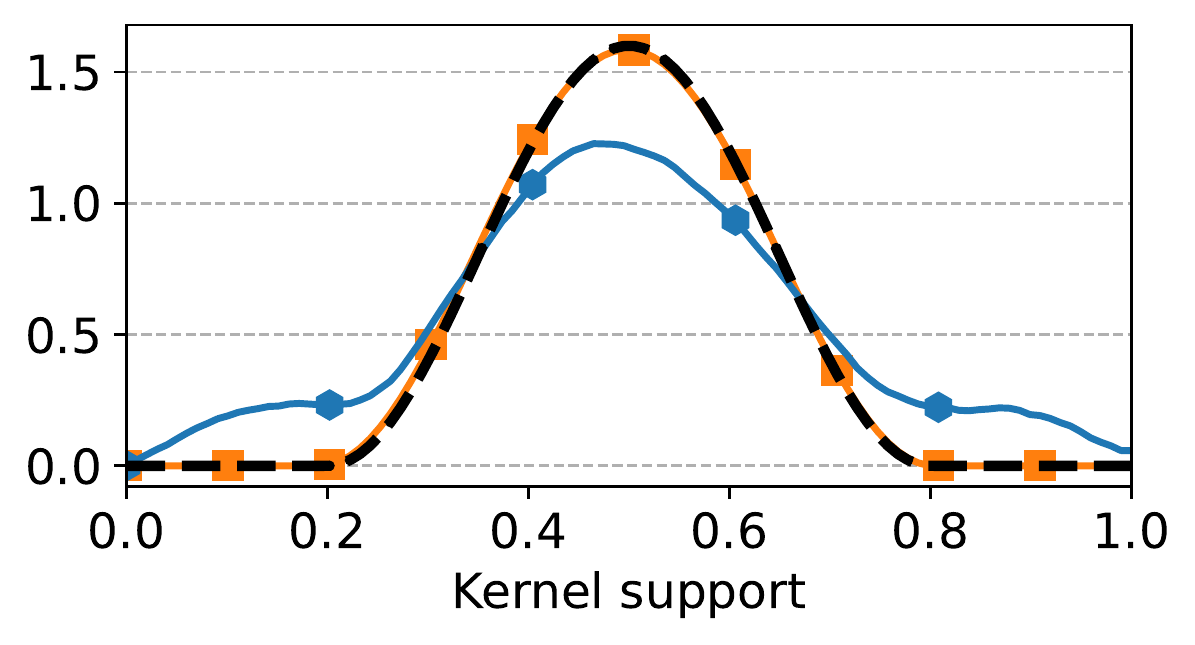}  & 
    \includegraphics[width=0.45\linewidth]{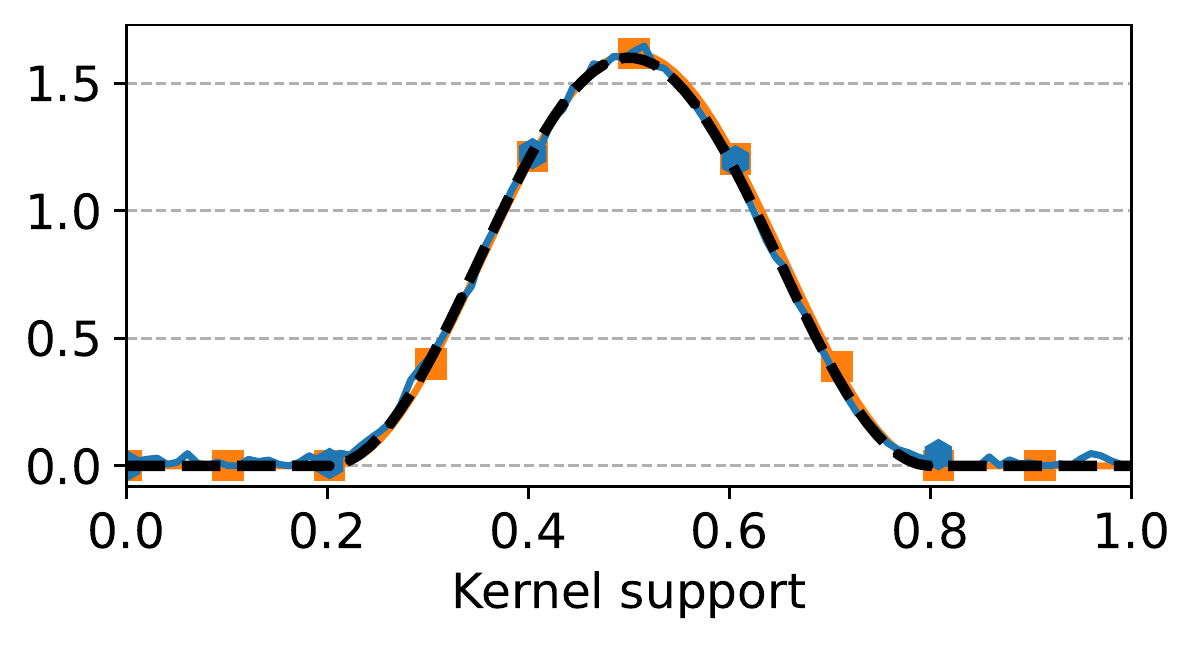}\\
    \multicolumn{2}{c}{    \includegraphics[width=0.98\linewidth]{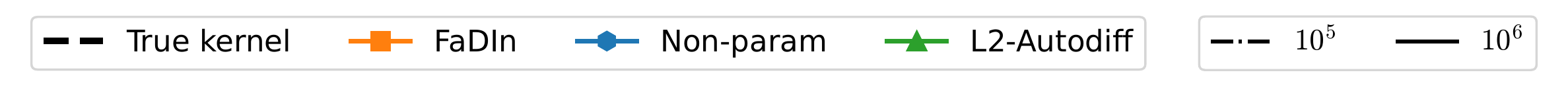}} \\
     \includegraphics[width=0.45\linewidth]{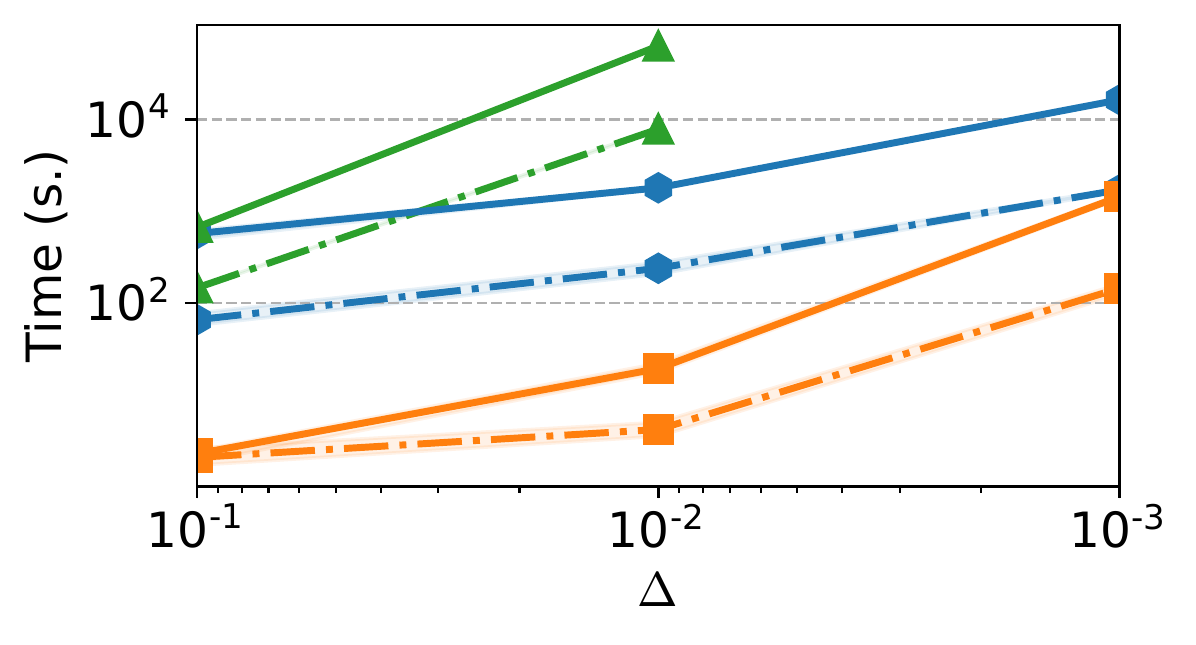} &\includegraphics[width=0.45\linewidth]{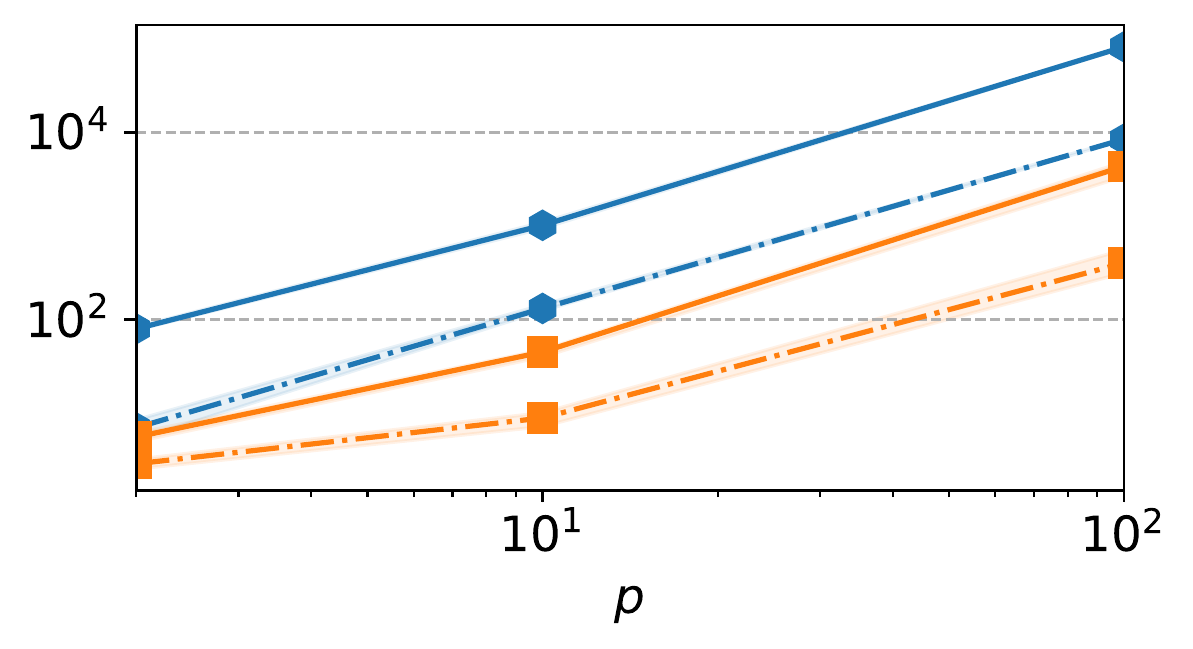}
\end{tabular}
    \vspace{-10pt}
    \caption{Comparison between our approach FaDIn and non-parametric approach.
    Estimated kernels with $\Delta=0.01$ in a relatively small data setting with $T=10^3$ (top left), in a large data setting with $T=10^6$ (top right), and computation time in a large data setting with $T\in \{10^5,10^6\}$ w.r.t. the stepsize $\Delta$ (bottom left) and the dimension $p$ (bottom right).
    In contrast to non-parametric kernels, FaDIn estimates well the true kernel in a small regime while it is computationally faster than non-parametric kernels in a large regime.}
    \label{fig:nonparam}
\end{figure}

\begin{figure}[!ht]
    \centering
\begin{tabular}{cc}
   \hspace{0.5cm} $T=10^3$ & \hspace{0.5cm}$T=10^4$ \\
       \includegraphics[width=0.45\linewidth]{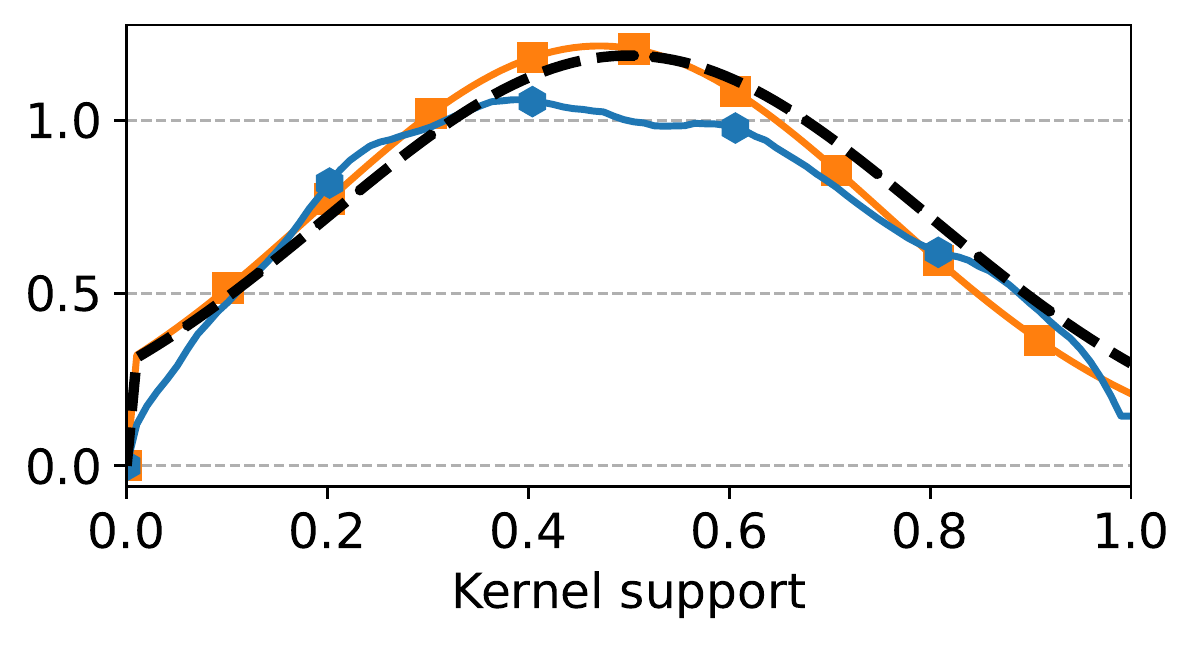}  & 
    \includegraphics[width=0.45\linewidth]{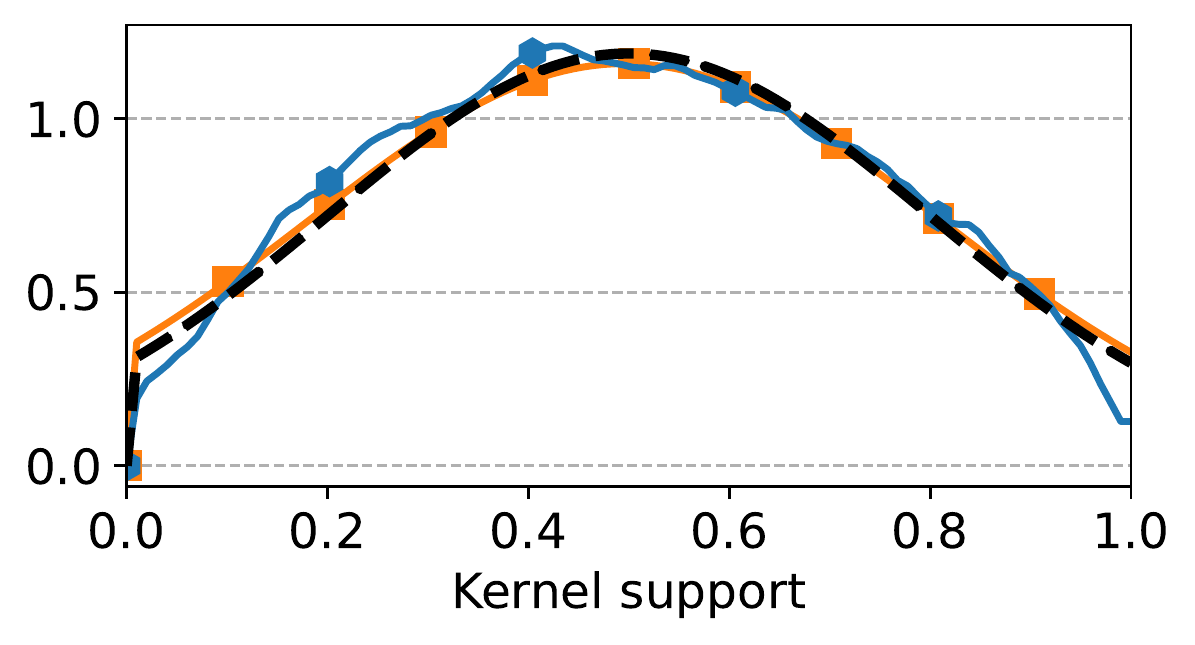}\\
    \hspace{0.5cm} $T=10^5$ & \hspace{0.5cm}$T=10^6$ \\
       \includegraphics[width=0.45\linewidth]{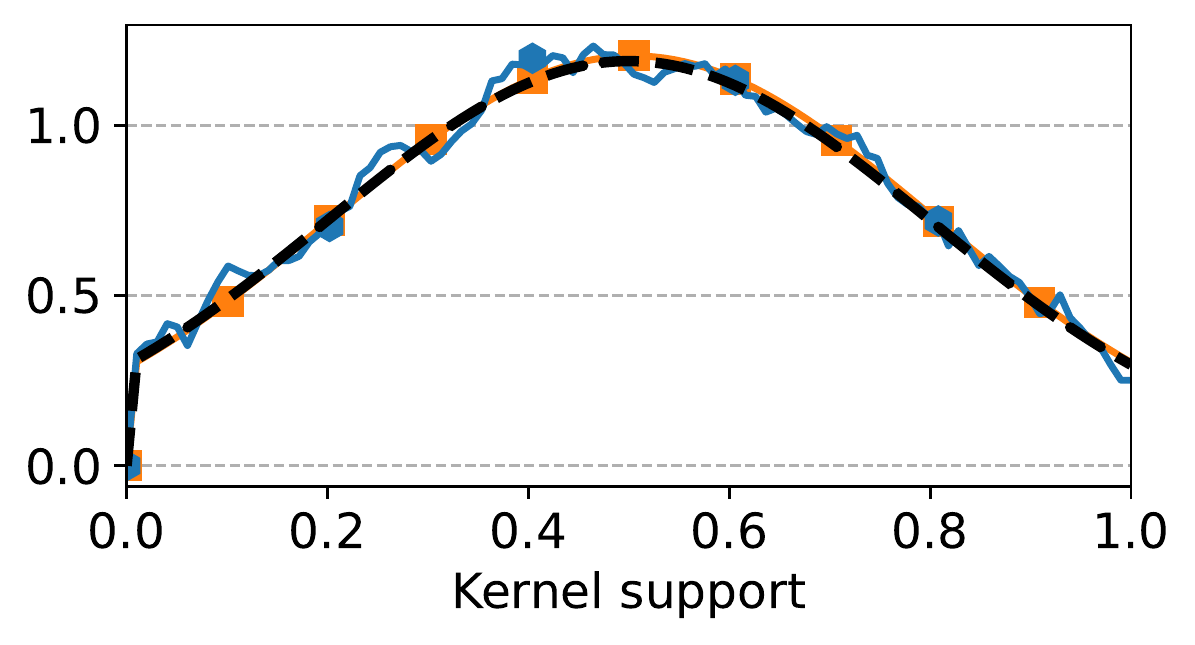}  & 
    \includegraphics[width=0.45\linewidth]{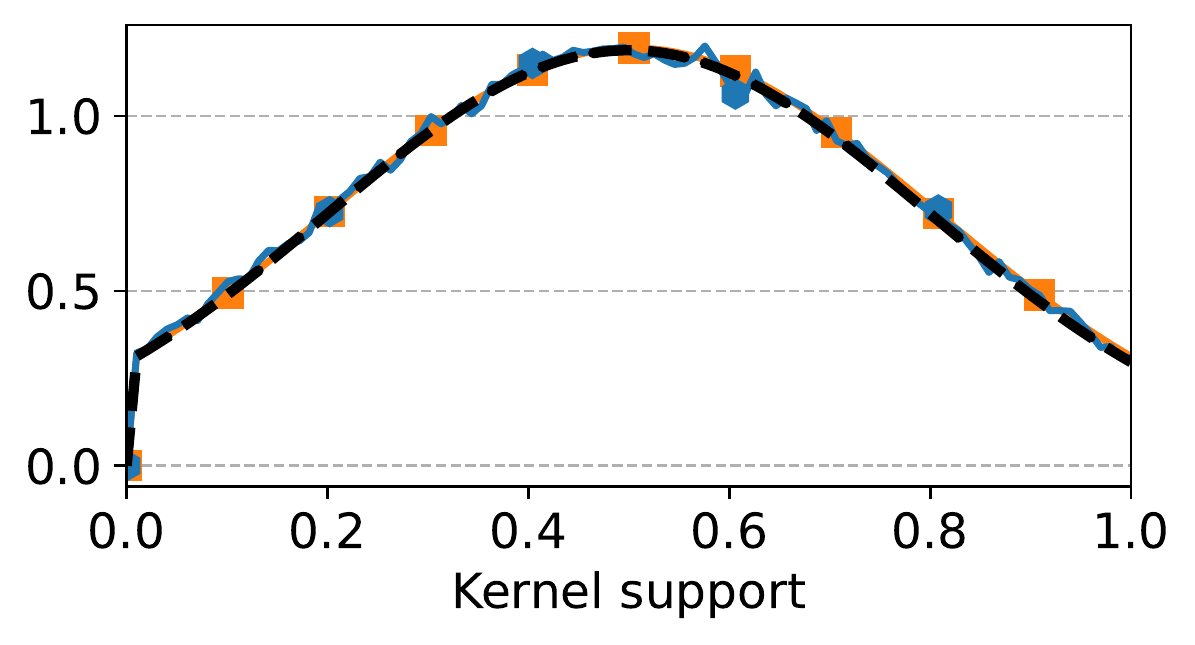}\\
\end{tabular}
    \vspace{-10pt}
    \caption{Comparison between our approach FaDIn and non-parametric approach for a Truncated Gaussian kernel.  Estimated kernels with $\Delta=0.01$ and $T\in \{10^3, 10^4, 10^5, 10^6 \}$. The true kernel, FaDIn and the non-parametric approach are depicted in black, orange and blue, respectively.}
    \label{fig:nonparam:tg}
\end{figure}

\begin{figure}[!ht]
    \centering
\begin{tabular}{cc}
   \hspace{0.5cm} $T=10^3$ & \hspace{0.5cm}$T=10^4$ \\
       \includegraphics[width=0.45\linewidth]{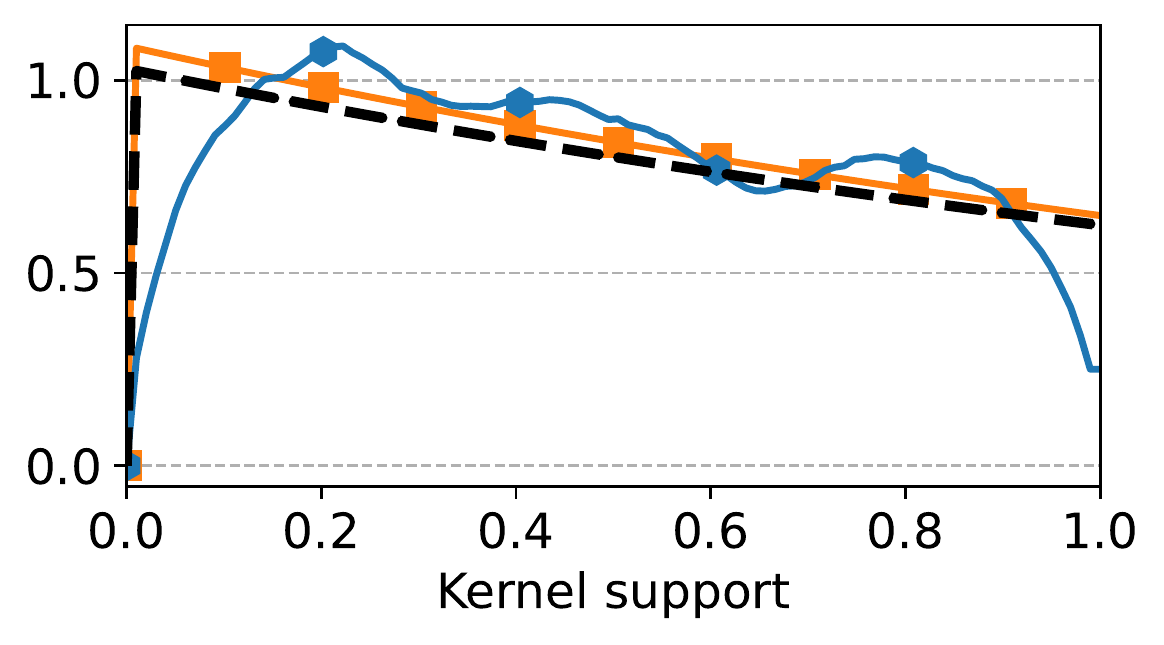}  & 
    \includegraphics[width=0.45\linewidth]{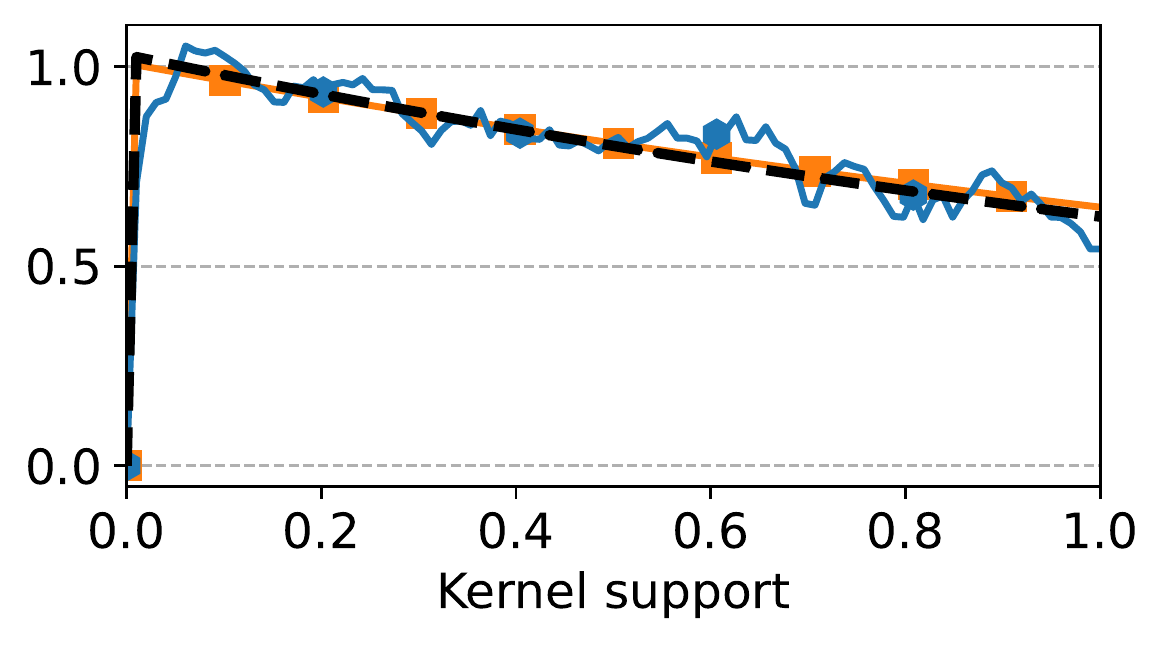}\\
    \hspace{0.5cm} $T=10^5$ & \hspace{0.5cm}$T=10^6$ \\
       \includegraphics[width=0.45\linewidth]{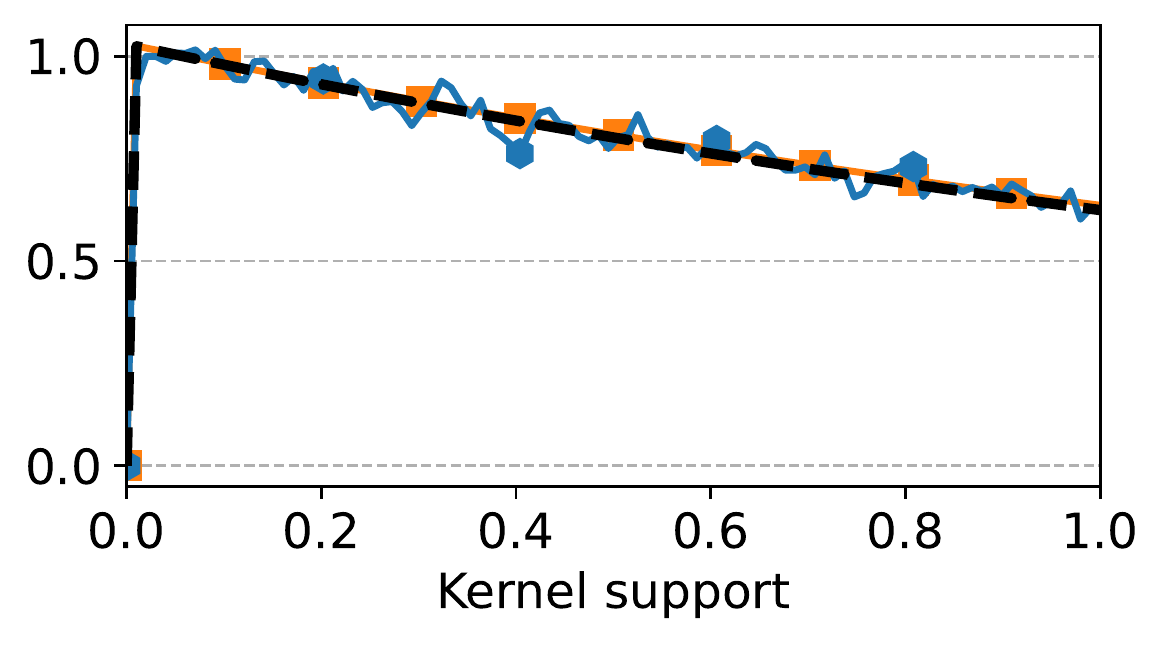}  & 
    \includegraphics[width=0.45\linewidth]{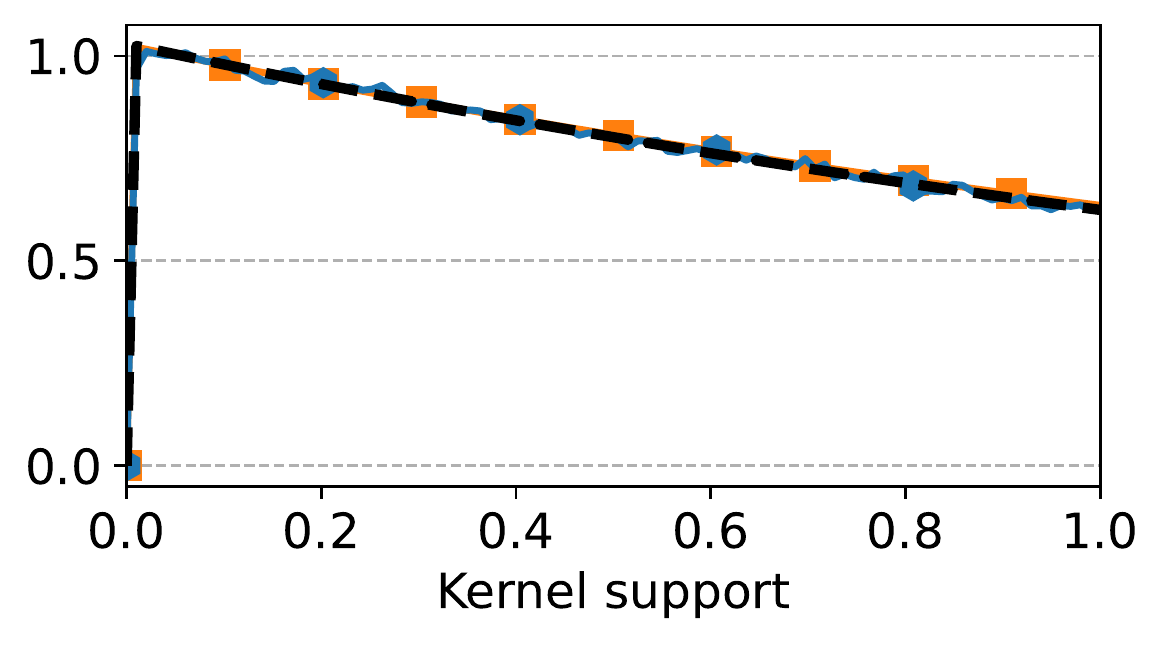}\\
\end{tabular}
    \vspace{-10pt}
    \caption{Comparison between our approach FaDIn and non-parametric approach for a Truncated Exponential kernel.  Estimated kernels with $\Delta=0.01$ and $T\in \{10^3, 10^4, 10^5, 10^6 \}$. The true kernel, FaDIn and the non-parametric approach are depicted in black, orange and blue, respectively.}
    \label{fig:nonparam:exp}
\end{figure}

On the one hand, in a relatively small data regime where $T=10^3$, we evaluate the statistical accuracy of the estimated kernel of both methods with the discretization parameter $\Delta=0.01$.
As we can see in \autoref{fig:nonparam} (top left), the non-parametric approach fails to recover the structure of the kernel.
The non-parametric approach results in noisy kernel estimates, with probability mass where the kernel is zero.
In contrast, FaDIn can recover the kernel parameters used to simulate data even with a small number of events.
On the other hand, we evaluate the computational times varying the discretization steps in a large data regime where $T=10^5$ and $T=10^6$ with the same simulation parameters.
\autoref{fig:nonparam} (bottom left) reports the average computational times (over 10 runs) regarding the discretization stepsize $\Delta$ and the dimension $p$.
Although both approaches can recover the kernel under which we simulate data (see  \autoref{fig:nonparam}, top right), FaDIn is a great deal more computationally efficient than the non-parametric and the automatic differentiation implementations, improving the computational speed by $\approx 100$ when $\Delta\in [0.1, 0.01]$ and by $\approx 10$ when $\Delta \approx 0.001$.
The computation speed regarding the dimension of the MHP is improved by $\approx 10$.
It is worth noting that the  $\ell_2$-Autodiff explodes in memory when $\Delta>0.01$ or when the dimension grows.
Additional shapes of kernels are displayed in \autoref{fig:nonparam:tg} for the Truncated Gaussian and in \autoref{fig:nonparam:exp} for the Truncated Exponential kernels.

\color{black}

\subsection{Discretization on EM estimates (DriPP)}\label{subsec:app:em}
\begin{figure}[!ht]
    \centering
    \includegraphics[scale=0.8]{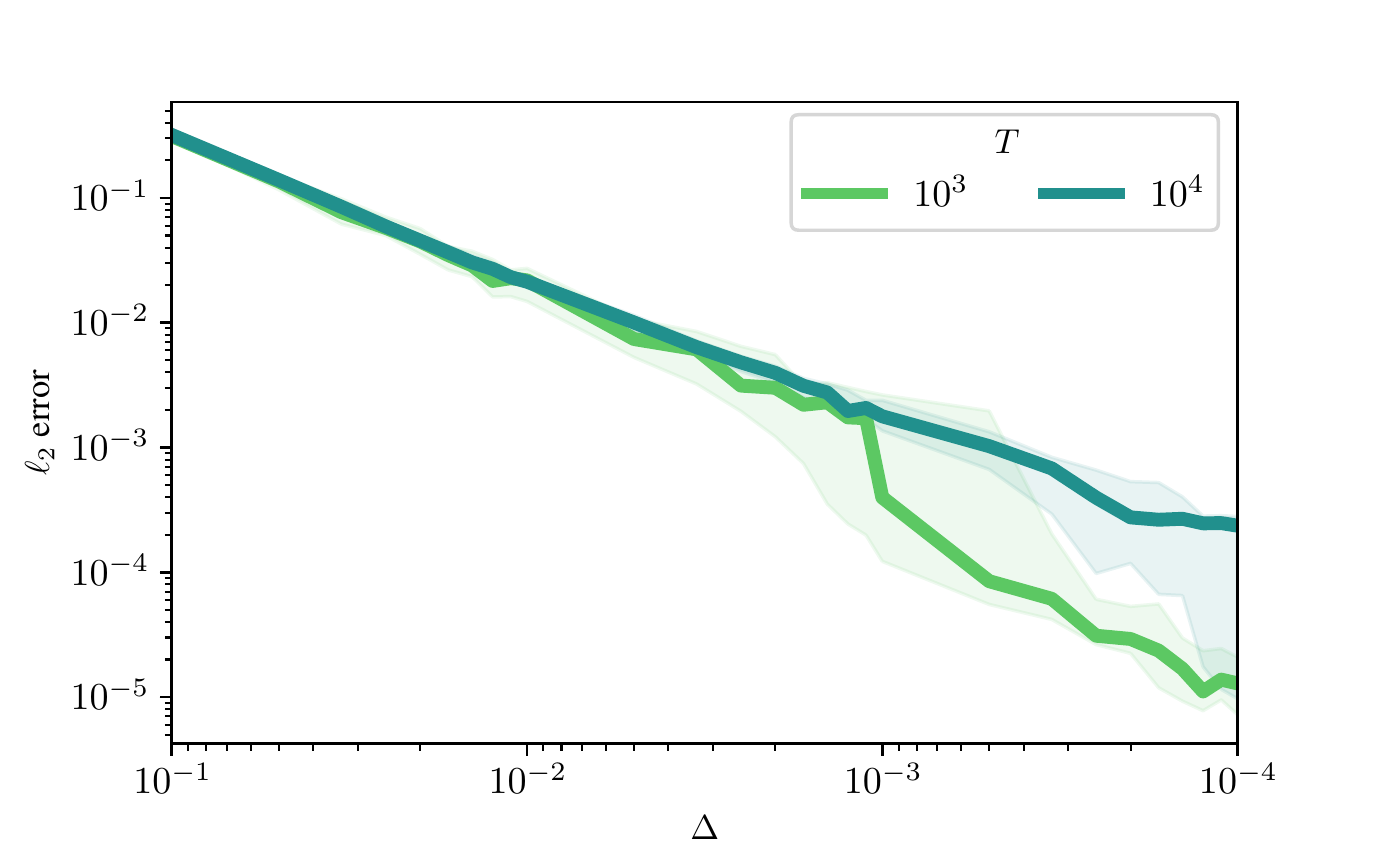}
    \caption{Median and interquartile error bar of the $\ell_2$ norm between the parameters estimated computed with EM algorithm, continuously and discretely, w.r.t. the stepsize $\Delta$.
    This figure confirms the results from \autoref{prop:bias}; that is, that the convergence of $\widehat\theta_\Delta$ towards $\widehat\theta_c$ is linear with respect to $\Delta$.}
    \label{fig:app:emp_bias}
\end{figure}

\autoref{fig:app:emp_bias} displays the convergence of the estimator $\widehat\theta_\Delta$ towards $\widehat\theta_c$ as $\Delta$ goes to 0 in the same experimental setup as the right part of \autoref{fig:consistency}.

\begin{figure}[!ht]
    \centering
    \includegraphics{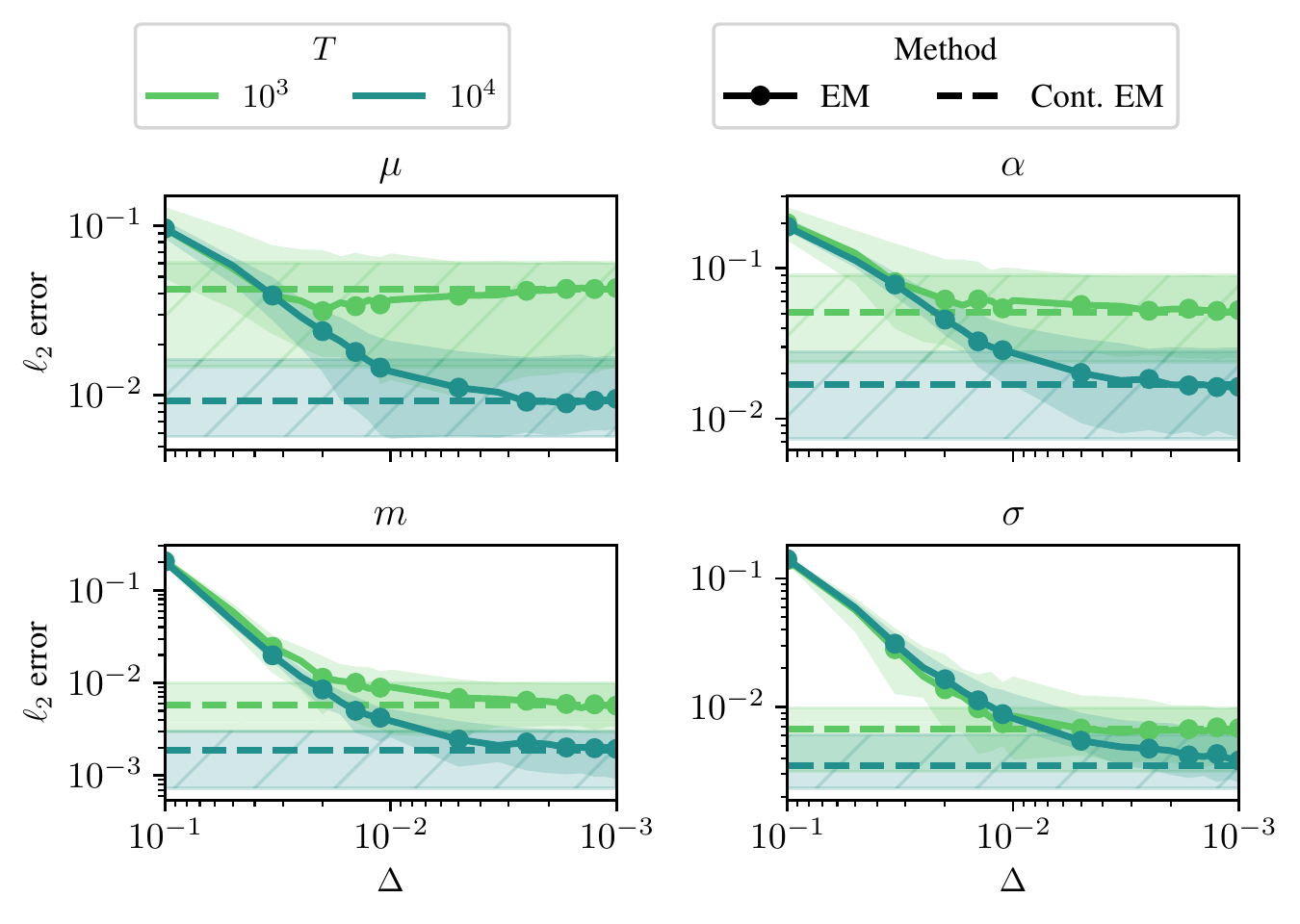}
    \caption{Median and interquartile error bar of the $\ell_2$ norm between true parameters and parameter estimates computed with EM algorithm, continuously and discretely, w.r.t. the stepsize $\Delta$.}
    \label{fig:app:fig1_poisson_0_5_baseline_3}
\end{figure}

\autoref{fig:app:fig1_poisson_0_5_baseline_3} presents the detailed results, \ie  parameter-wise, of the experiment shown in \autoref{fig:consistency} (right).
In this experiment, we are interested in the context of Driven PP \citep{allain2021dripp} with an exogenous homogeneous PP.
The simulation parameter of the latter is set to 0.5, meaning that on average, 1 event occurs every 2 seconds on the driver.

\begin{figure}[!ht]
    \centering
    \includegraphics{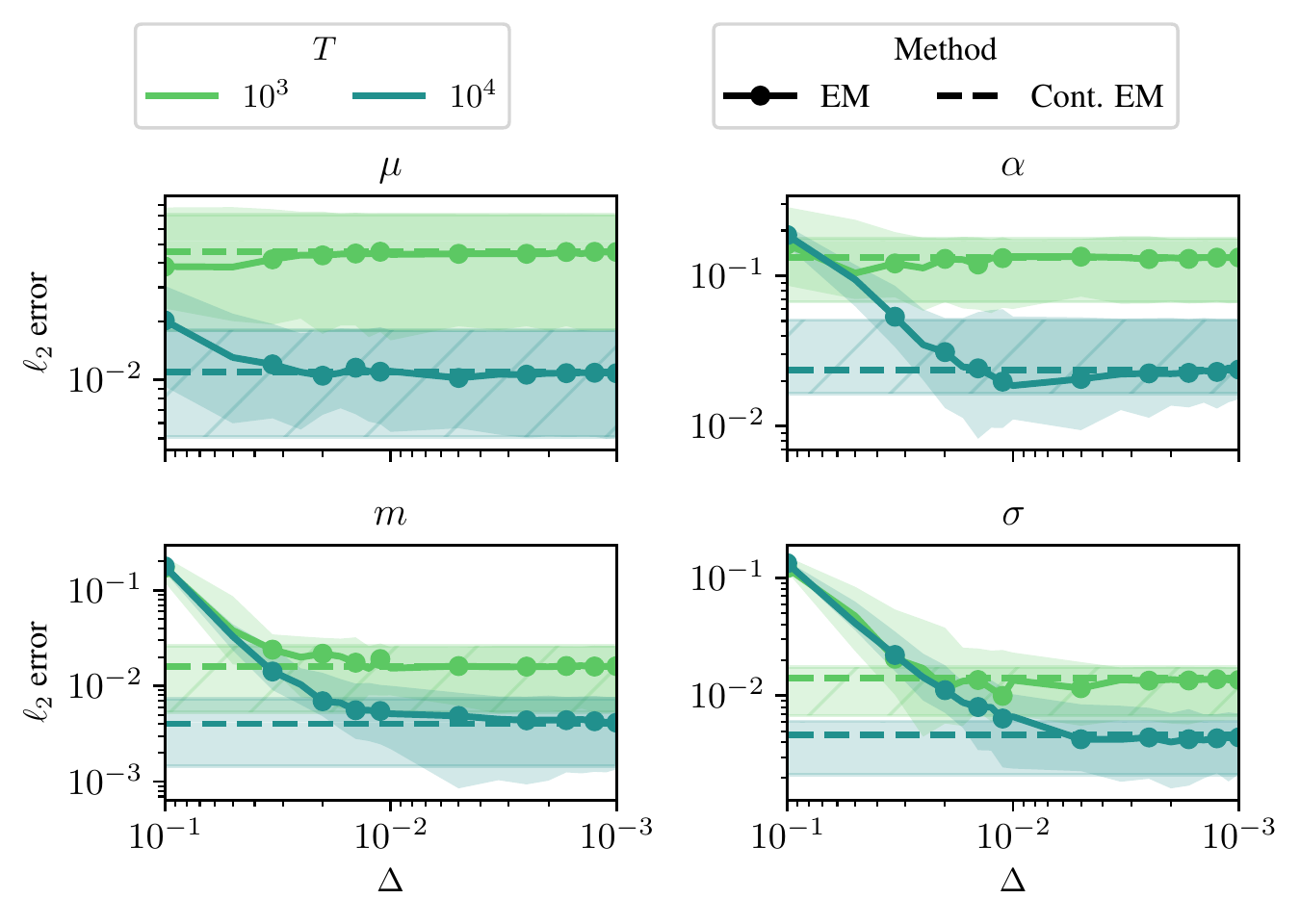}
    \caption{Median and interquartile error bar of the $\ell_2$ norm between true parameters and parameter estimates computed with EM algorithm, continuously and discretely, w.r.t. the stepsize $\Delta$.}
    \label{fig:app:fig1_poisson_0_1_baseline_3}
\end{figure}

\autoref{fig:app:fig1_poisson_0_1_baseline_3} presents the results of the same experiment with Poisson parameter set to 0.1 which represents roughly five times less events.

\subsection{Discretization effect on FaDIn estimates}\label{subsec:app:fadin}

This section presents additional results related to the \autoref{subsec:consistency}. We reproduce the experiments of this section with FaDIn and two other kernels: Raised Cosine and Truncated Exponential.

The Raised Cosine kernel is defined by:
\begin{equation*}
   \phi(\cdot)=\alpha \left[{1 + \cos \pars{\frac{\cdot - u}{\sigma}\pi - \pi}} \right]\mathbb{I}\left\{\cdot\in[u; \; u+2\sigma]\right\}   .
\end{equation*}
The parameters to estimate are $\mu, \alpha, u$ and $\sigma$. The Truncated Exponential kernel of decay parameter $\gamma \in \bbR_+$, with fixed support $\intervalleFF{a}{b} \subset \bbR^+$, $b>a$ is defined as $\phi(\cdot) = \alpha \kappa(\cdot), \alpha \geq 0$ with
\begin{equation*}
    \kappa(\cdot) \coloneqq \kappa\pars{\cdot ; \gamma, a, b} =  \frac{h\left(\cdot\right)}{H\left(b\right)-H\left(a\right)} \1[a\leq \cdot \leq b]
      ,
\end{equation*}
where here $h$ (resp. $H$) is the probability density function of parameter $\gamma$ (resp. cumulative distribution function) of the exponential distribution.
The parameters to estimate are $\mu, \alpha$ and $\gamma$.

\begin{figure}[!ht]
    \centering
    \begin{tabular}{cc}
     
        \multicolumn{2}{c}{\hspace{0.6cm}\includegraphics[scale=0.6]{ICLR/figures/approx/legend_TG_baseline.pdf}} \\
        \hspace{0.65cm}Raised Cosine & \hspace{0.65cm}Truncated Exponential \\
         \includegraphics[scale=0.5, trim=0cm 0 0 0]{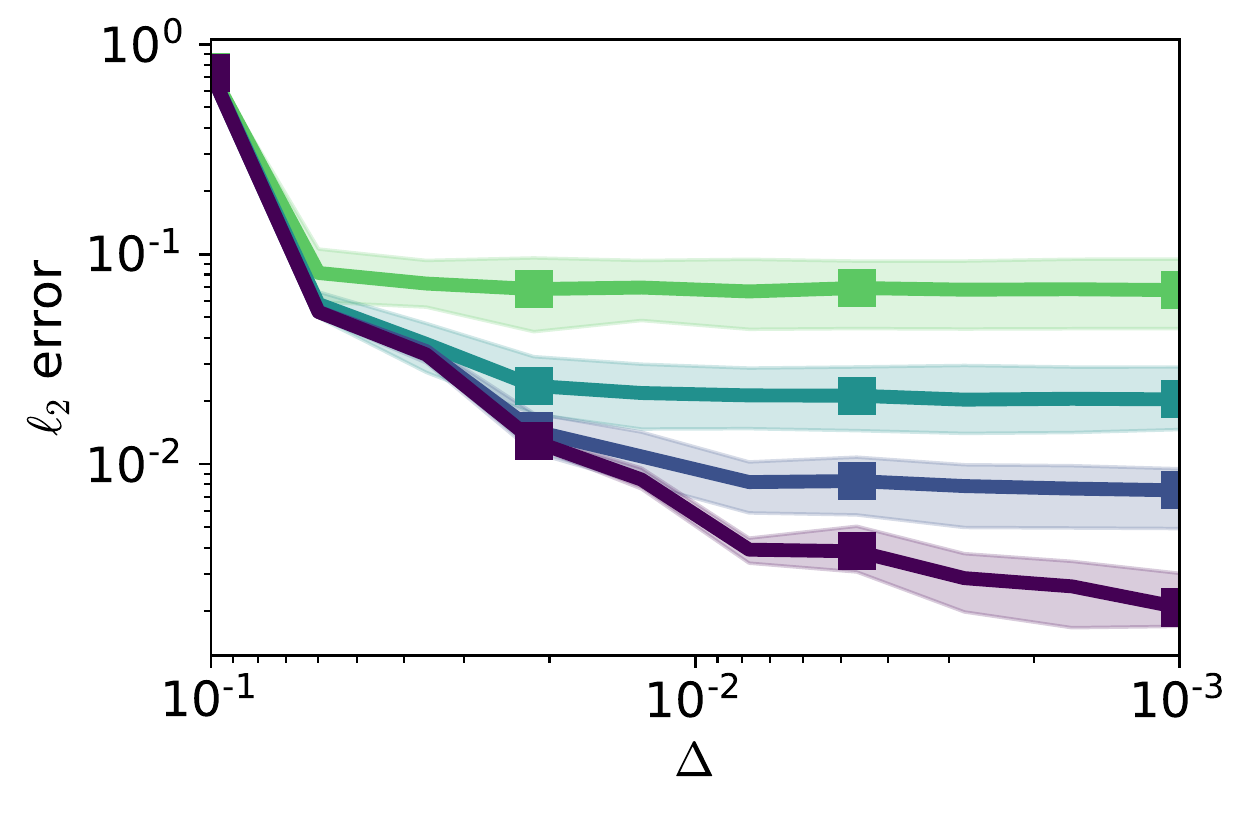} & \includegraphics[scale=0.5, trim=0cm 0 0 0]{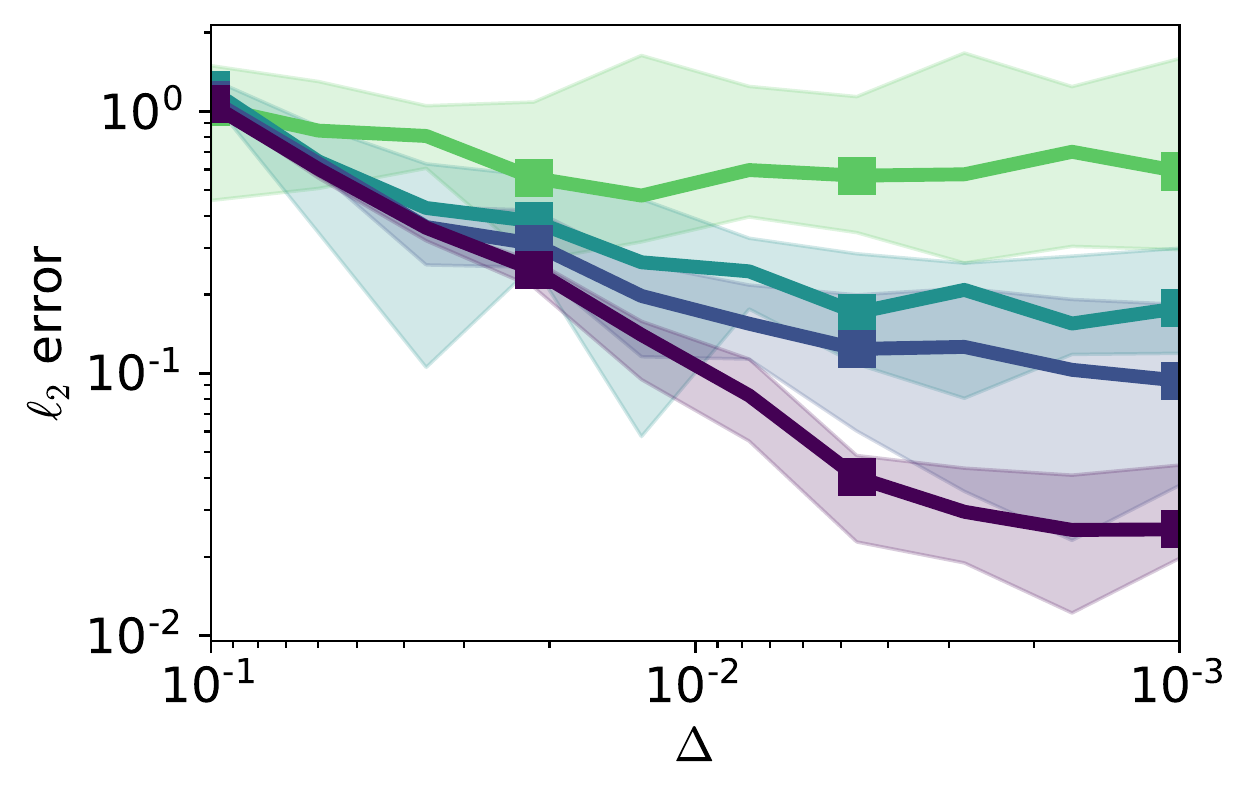}
    \end{tabular}
   
    \caption{Comparison of the influence of the discretization on the parameter estimation of FaDIn   for a Raised Cosine kernel (left) and an Exponential kernel (right) w.r.t. the stepsize of the grid $\Delta$.}
    \label{fig:app:approx_l2}
\end{figure}

\begin{figure}[!ht]
    \centering
    \begin{tabular}{cc}
    \multicolumn{2}{c}{\hspace{0.6cm} \includegraphics[scale=0.6]{ICLR/figures/approx/legend_TG_baseline.pdf}} \\
    \hspace{0.8cm}$ (\hat{\mu}-\mu)^2$ & \hspace{0.8cm}$ (\hat{\alpha}-\alpha)^2$\\
         \includegraphics[scale=0.5, trim=0cm 0 0 0]{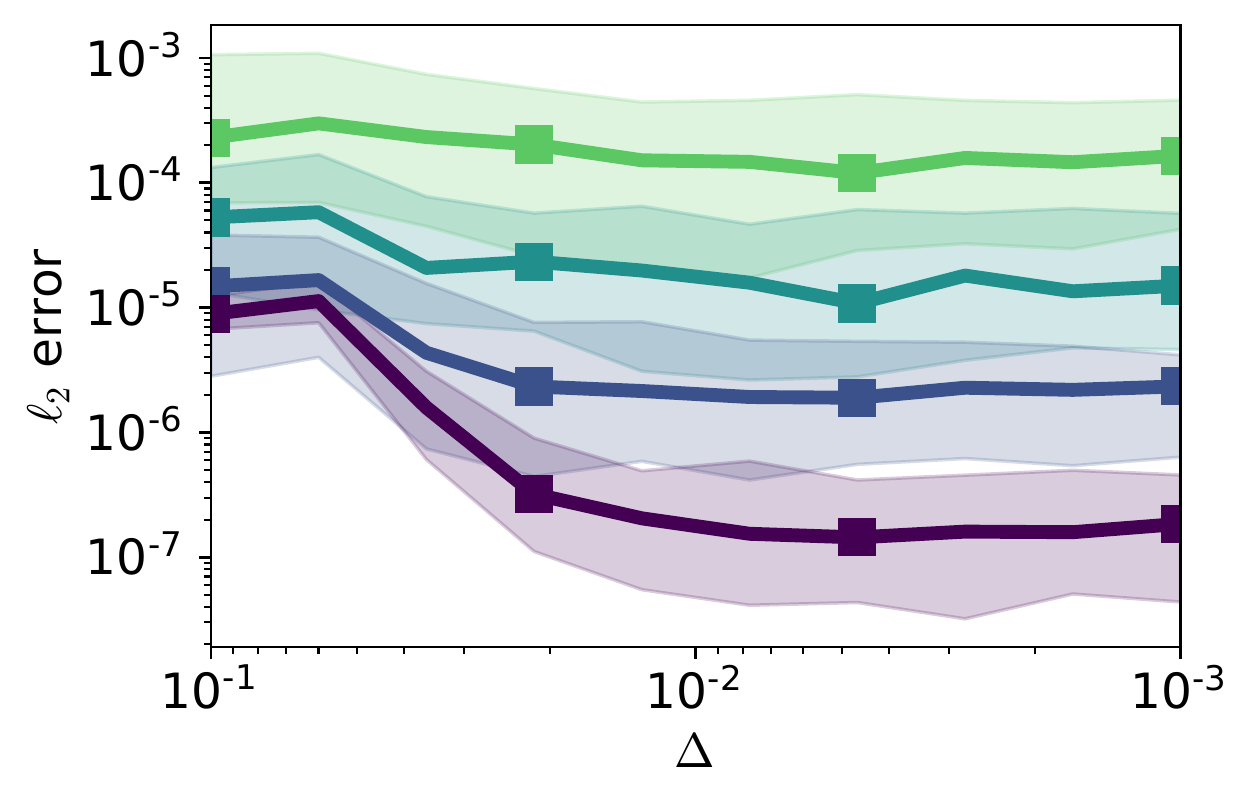} & \includegraphics[scale=0.5, trim=0cm 0 0 0]{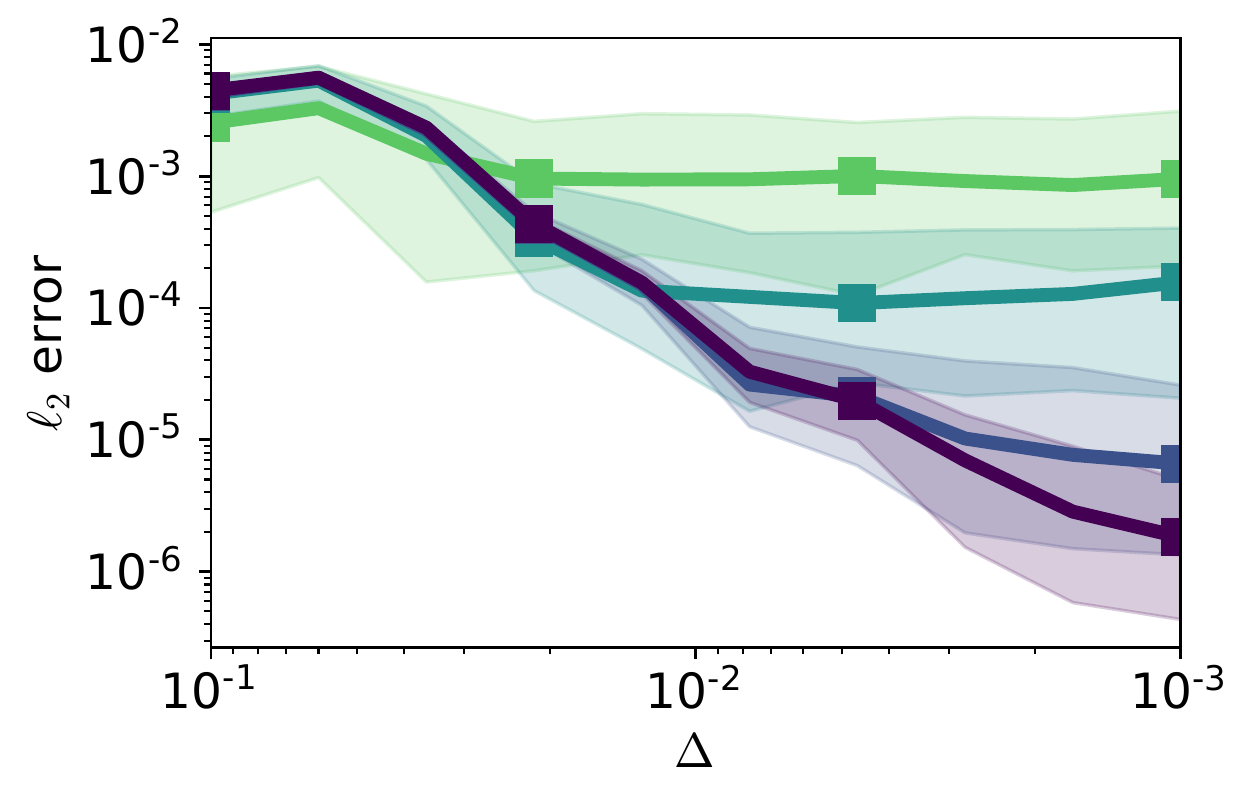} \\
         \hspace{0.8cm}$ (\hat{m}-m)^2$ & \hspace{0.8cm}$ (\hat{\sigma}-\sigma)^2$\\
         \includegraphics[scale=0.5, trim=0cm 0 0 0]{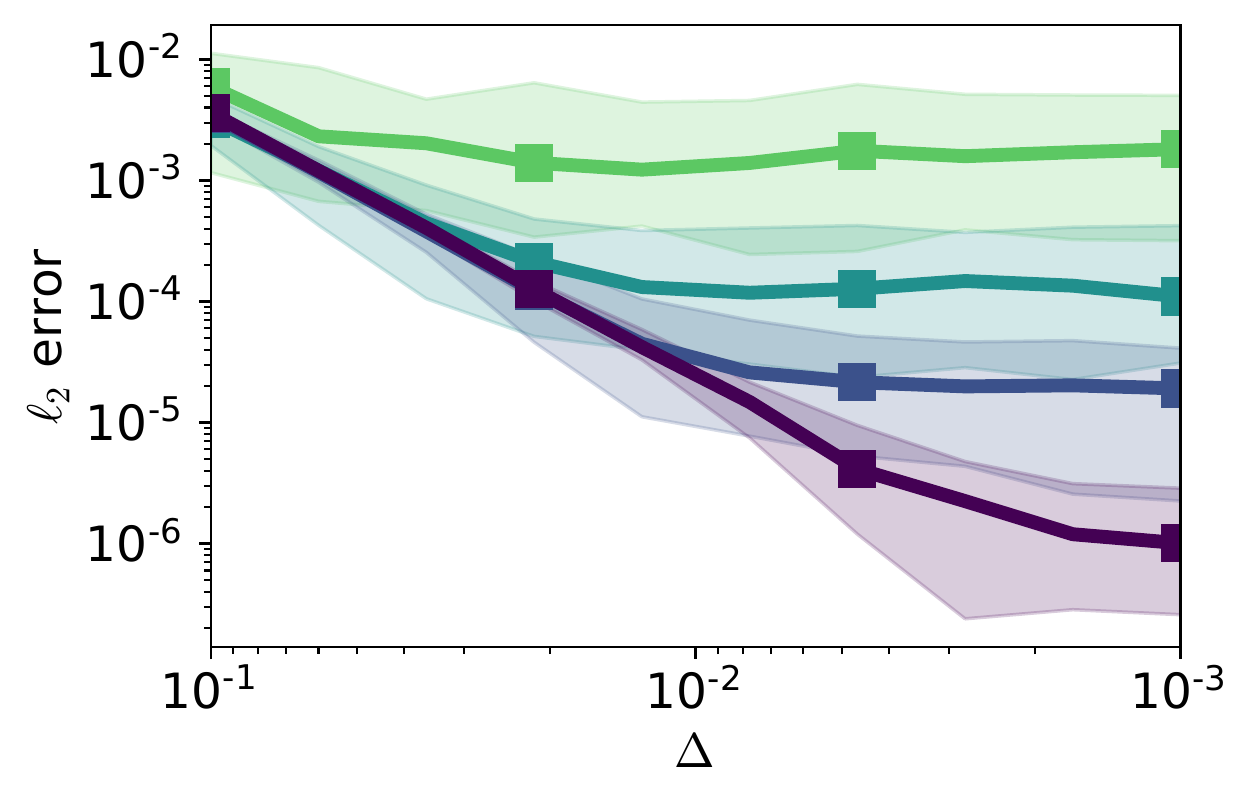} & \includegraphics[scale=0.5, trim=0cm 0 0 0]{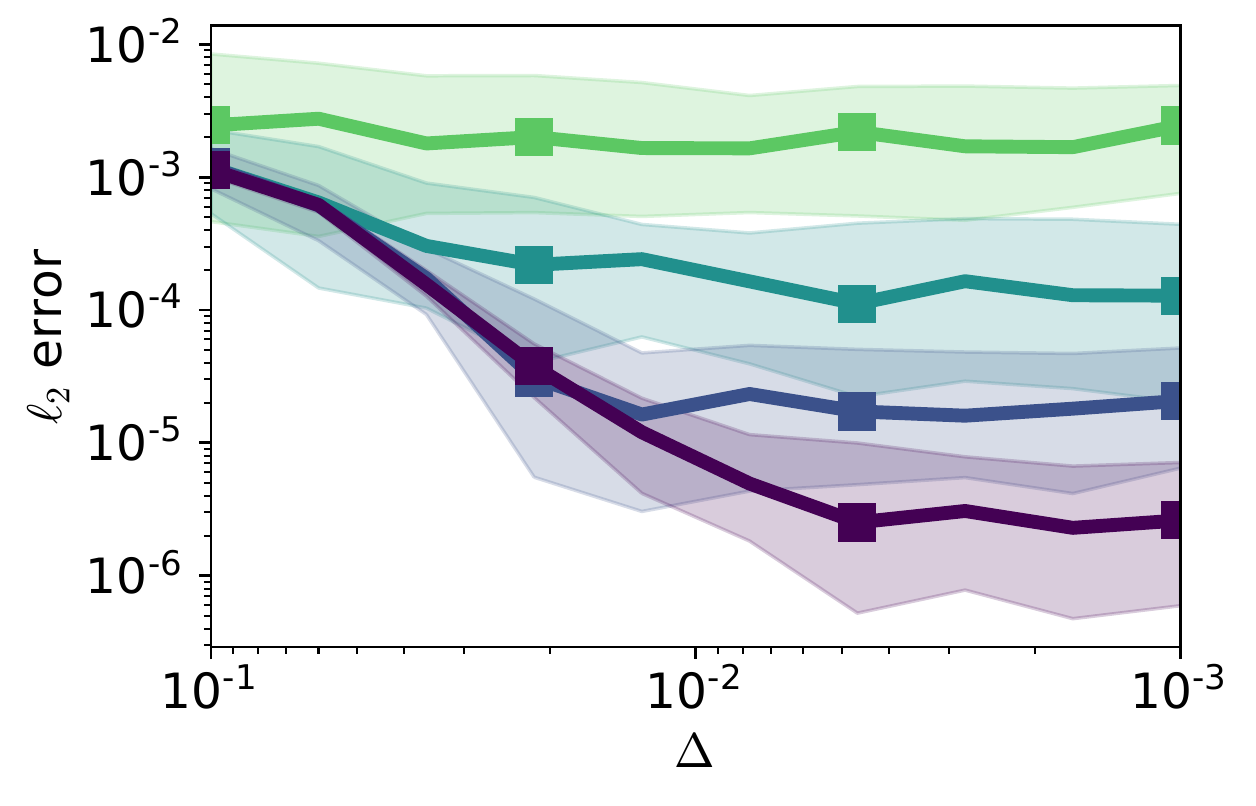}  
           \end{tabular}
    \caption{Error on parameters for the Truncated Gaussian kernel as a function of $T$ and $\Delta$.}
    \label{fig:app:approx_TG}
\end{figure}

\begin{figure}[!ht]
    \centering
    \begin{tabular}{cc}
    \multicolumn{2}{c}{\hspace{0.6cm}\includegraphics[scale=0.6]{ICLR/figures/approx/legend_TG_baseline.pdf}}\\
    \hspace{0.8cm}$ (\hat{\mu}-\mu)^2$ & \hspace{0.8cm}$ (\hat{\alpha}-\alpha)^2$\\
         \includegraphics[scale=0.5, trim=0cm 0 0 0]{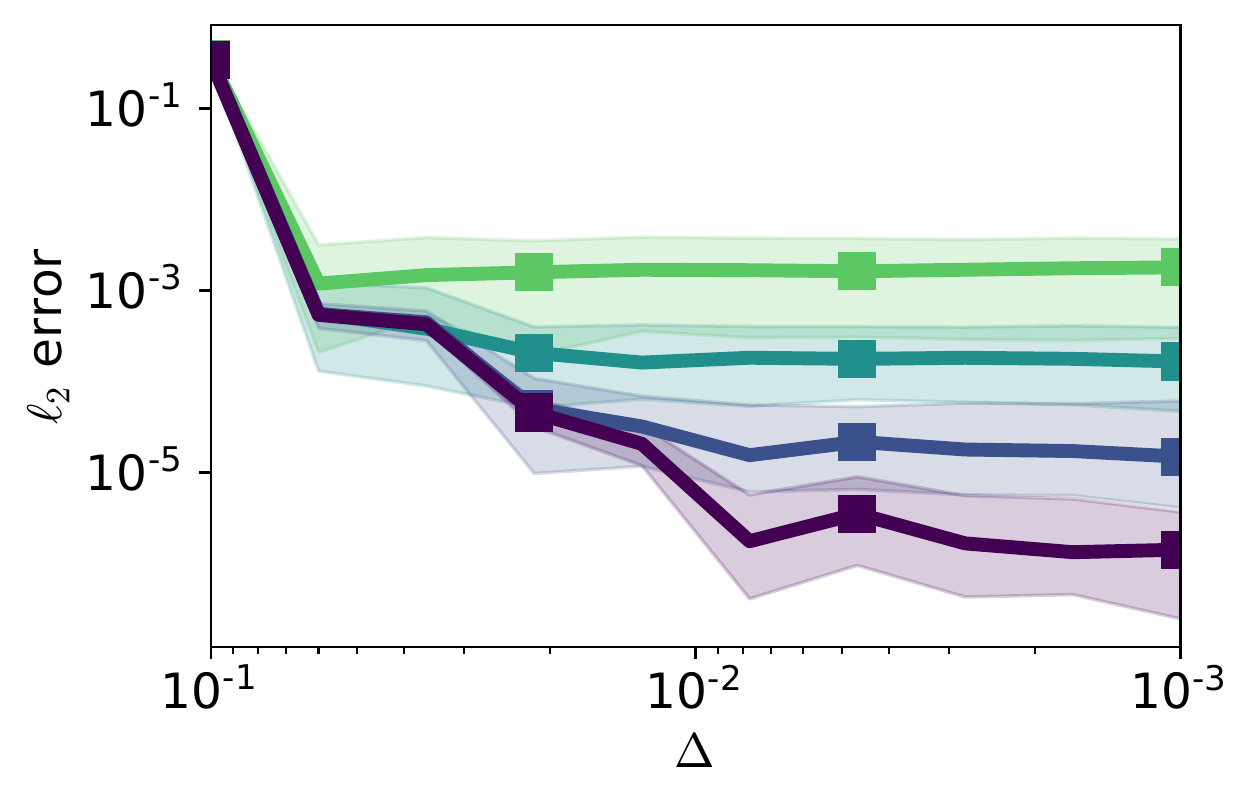} & \includegraphics[scale=0.5, trim=0cm 0 0 0]{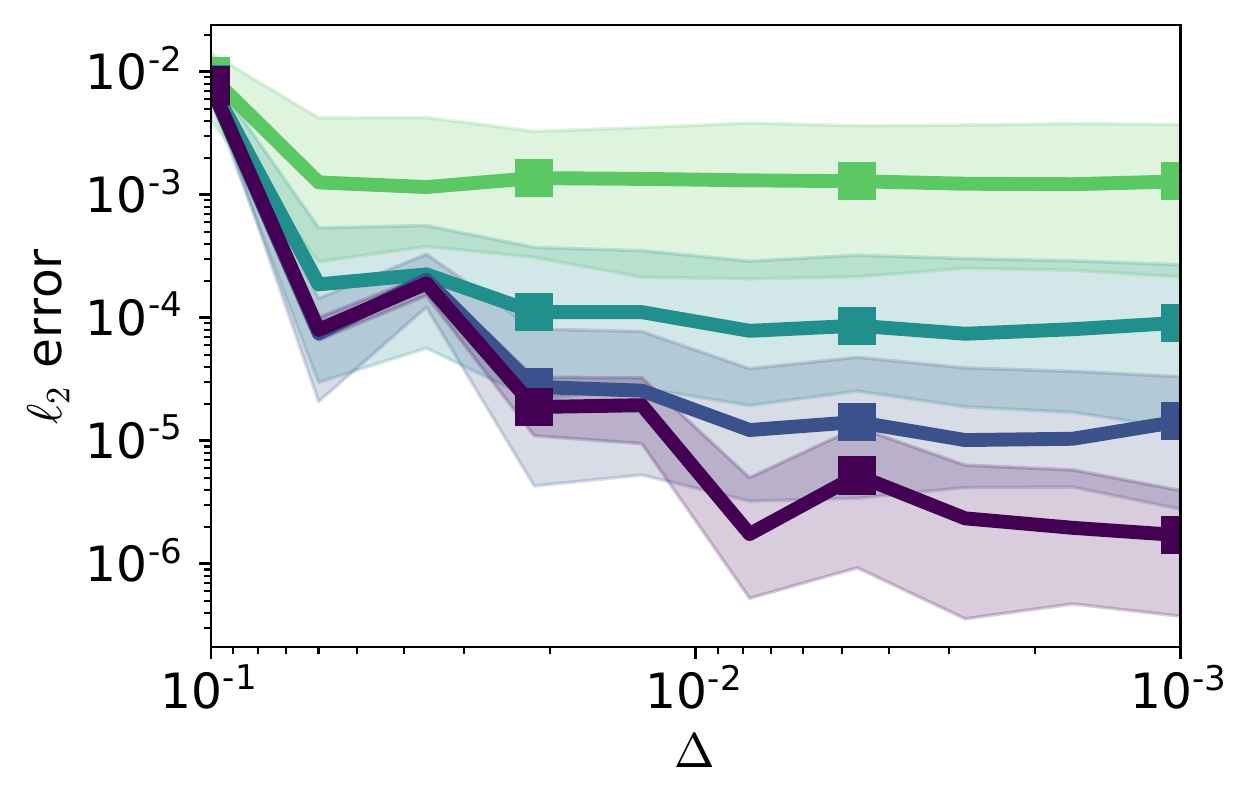} \\
         \hspace{0.8cm}$ (\hat{u}-u)^2$ & \hspace{0.8cm}$ (\hat{\sigma}-\sigma)^2$\\
         \includegraphics[scale=0.5, trim=0cm 0 0 0]{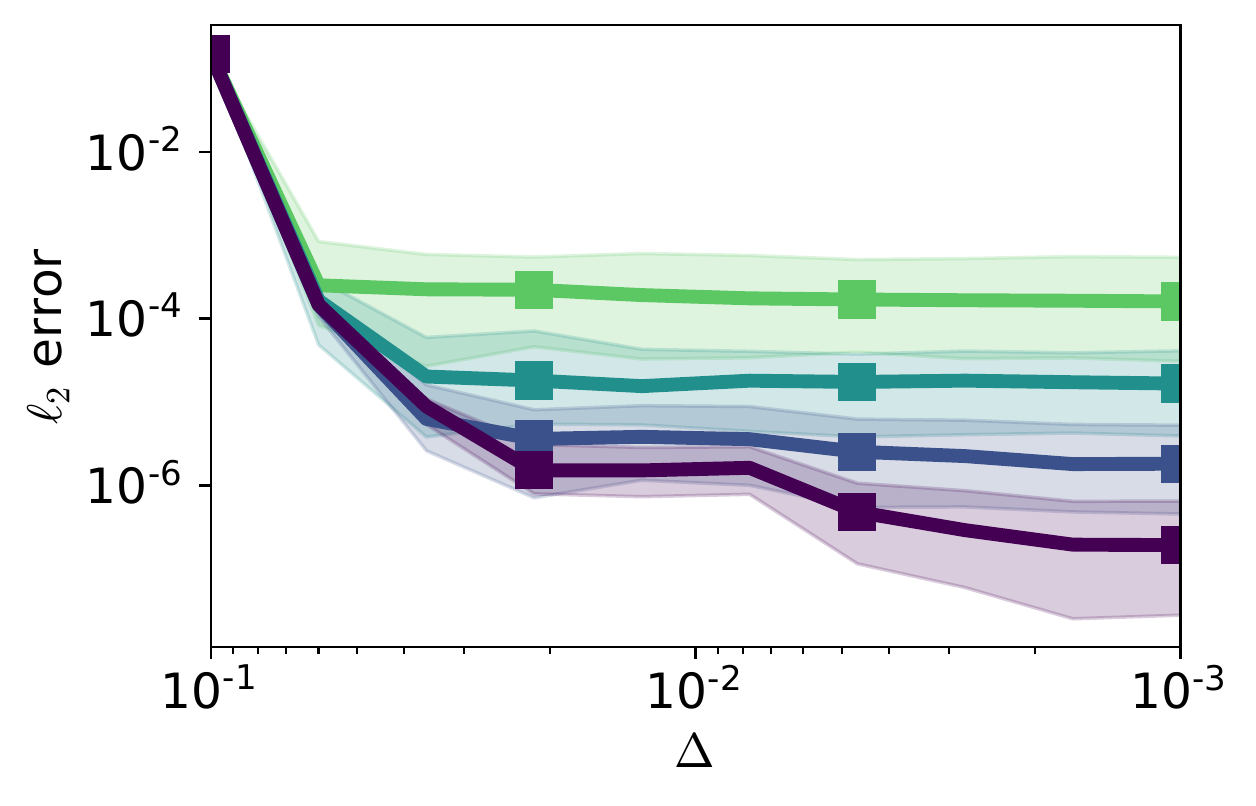} & \includegraphics[scale=0.5, trim=0cm 0 0 0]{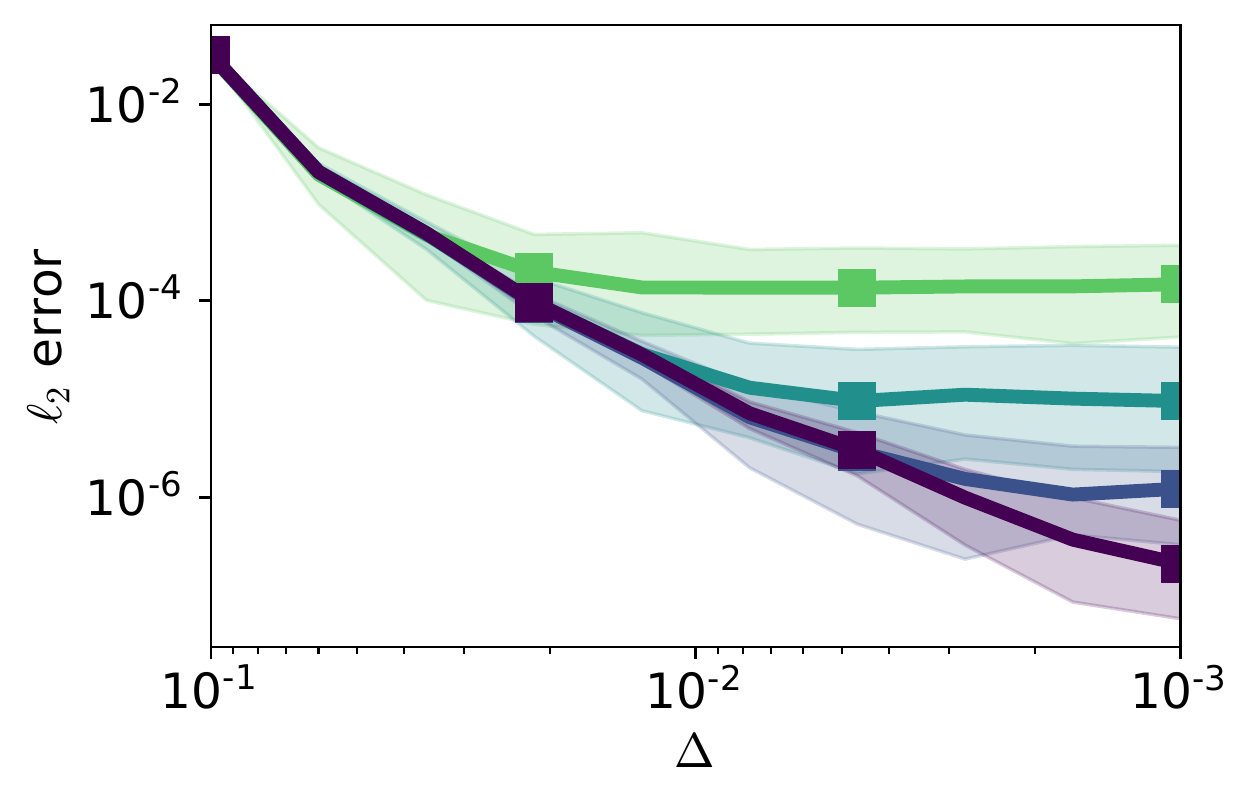}  
           \end{tabular}
    \caption{Error on parameters for the Raised Cosine kernel as a function of $T$ and $\Delta$.}
    \label{fig:app:approx_RC}
\end{figure}

\begin{figure}[!ht]
    \centering
    \begin{tabular}{cc}
    \multicolumn{2}{c}{\hspace{0.6cm} \includegraphics[scale=0.6]{ICLR/figures/approx/legend_TG_baseline.pdf}}\\
    \hspace{0.8cm}$ (\hat{\mu}-\mu)^2$ & \hspace{0.8cm}$ (\hat{\alpha}-\alpha)^2$\\
         \includegraphics[scale=0.5, trim=0cm 0 0 0]{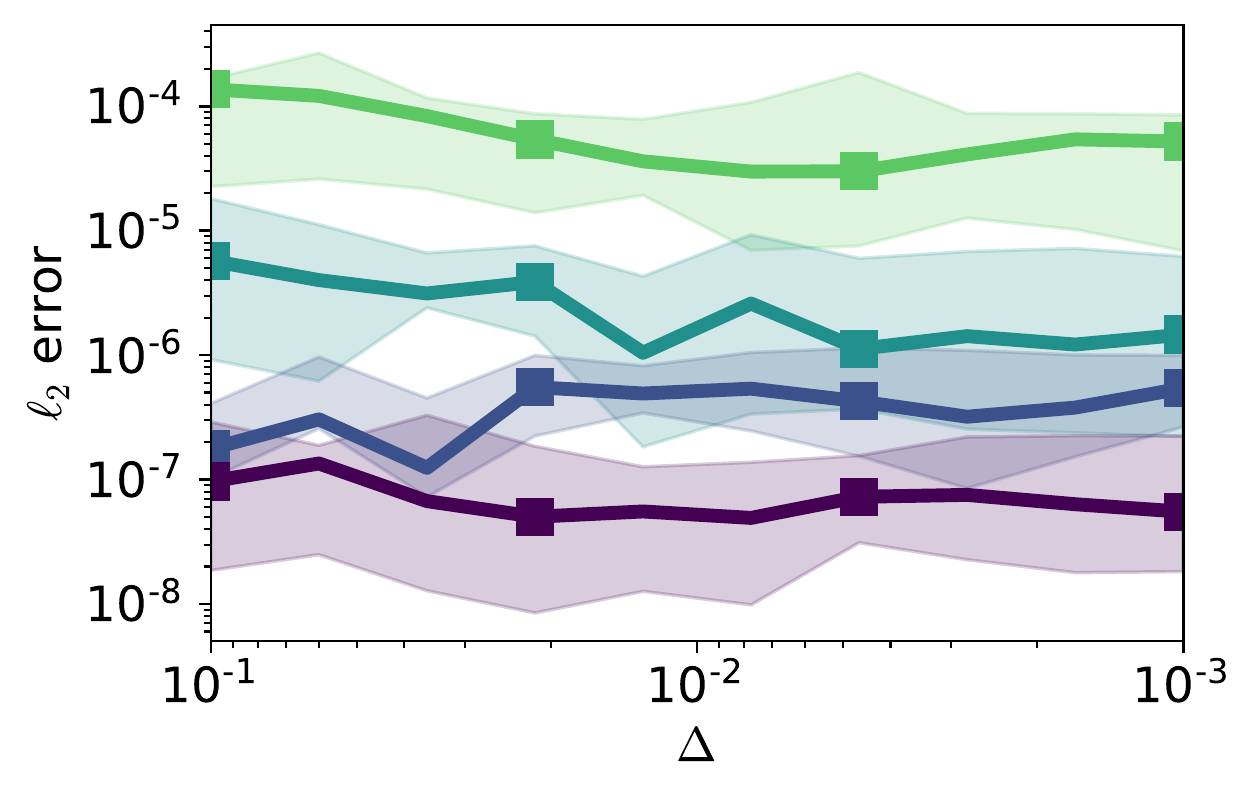} & \includegraphics[scale=0.5, trim=0cm 0 0 0]{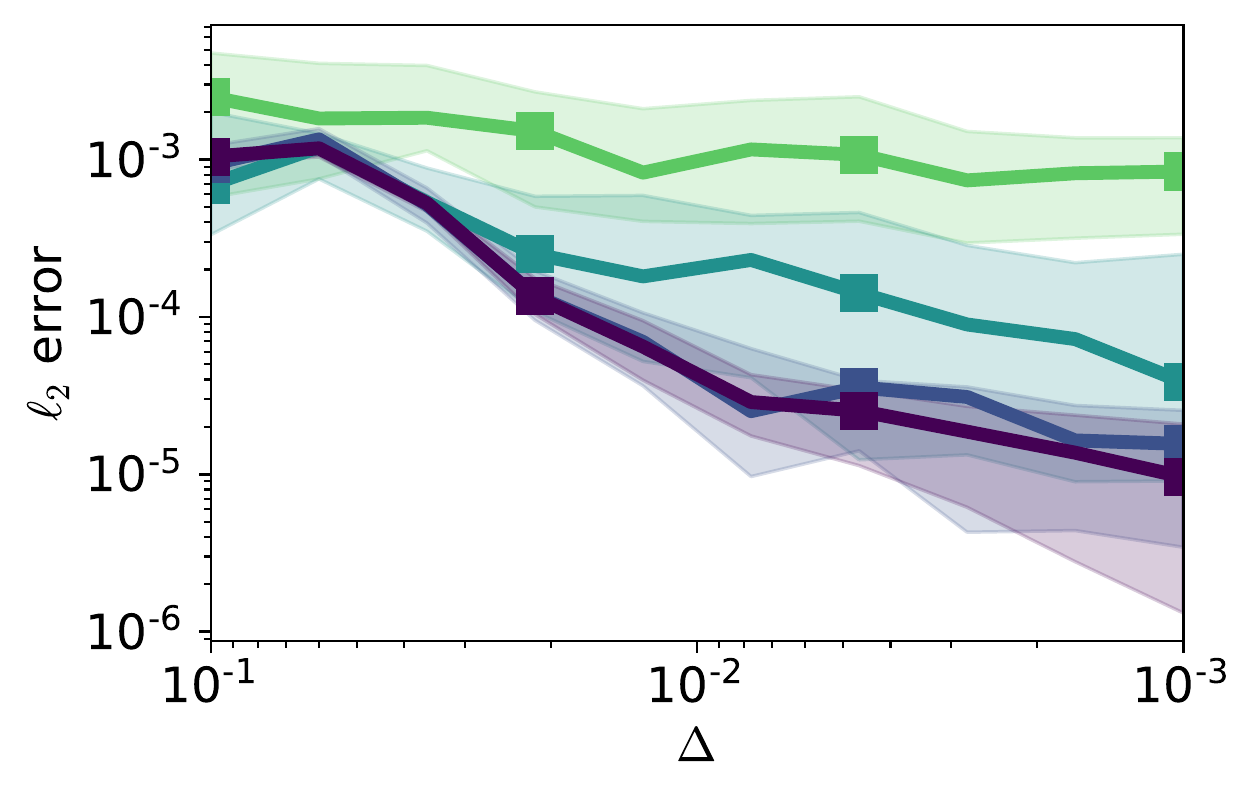} \\
    \multicolumn{2}{c}{\hspace{0.8cm} $(\hat{\gamma} - \gamma)^2$}\\
    \multicolumn{2}{c}{\includegraphics[scale=0.5]{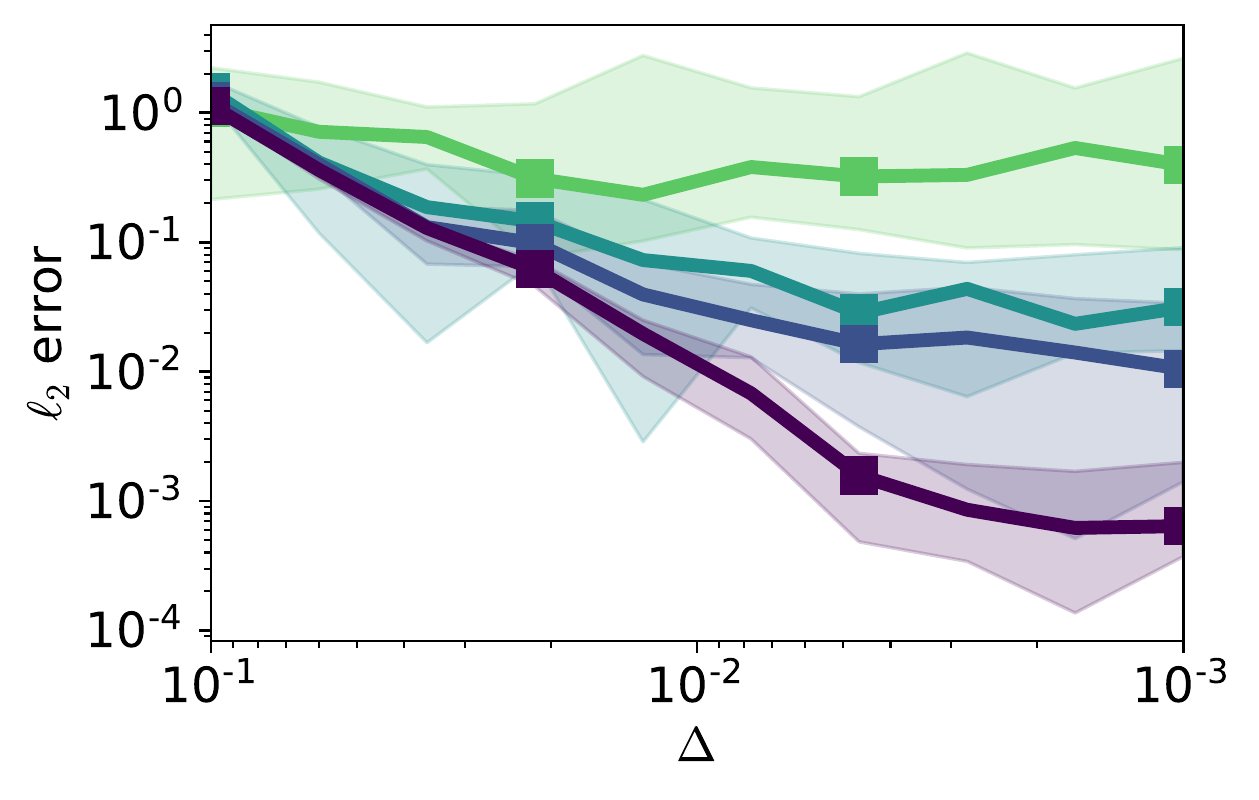}}
           \end{tabular}
    \caption{Error on parameters for the Truncated Exponential kernel as a function of $T$ and $\Delta$.}
    \label{fig:app:approx_EXP}
\end{figure}

Estimation results (median and $20$-$80\%$ quantiles) are displayed in \autoref{fig:app:approx_l2} and confirm the conclusion presented in  \autoref{subsec:consistency} about the consistency of the discretization for FaDIn. In addition, we display the quadratic error for each parameter separately in \autoref{fig:app:approx_TG} for the Truncated Gaussian, \autoref{fig:app:approx_RC} for the Raised Cosine and  \autoref{fig:app:approx_EXP} for the Truncated Exponential kernels.


\clearpage

\subsection{Other experiments on real data}\label{sec:app:addexpemeg}

After detailing the procedure we use in MEG experiments and adding extra information about used datasets, we provide a second experiment on the somatosensory dataset. 

\paragraph{Background on CDL}
The objective of CDL is to decompose a signal as the convolution between a translationally invariant pattern called atom and its sparse activation vector~\citep{Grosse-etal:2007}.
To do this, we minimize the following objective function:
\begin{equation*}
    \min _{D_{k}, z_{k}^{n}} \sum_{n=1}^{N} \frac{1}{2}\left\|X^{n}-\sum_{k=1}^{K} z_{k}^{n} * D_{k}\right\|_{F}^{2}+\lambda \sum_{k=1}^{K}\left\|z_{k}^{n}\right\|_{1},
    \text { s.t. }\left\|D_{k}\right\|_{F}^{2} \leq 1 \text { and } z_{k}^{n} \geq 0
\end{equation*}
where $\left\{ X^n \right\}_{n=1}^N \subset \mathbb{R}^{P\times T}$ is the observed signals, $\left\{ D_k \right\}_{k=1}^K \subset \mathbb{R}^{P\times L}$ is the dictionaries of atoms, $\left\{ z_k^n \right\}_{k=1}^K \subset \mathbb{R}^{\widetilde{T}}$ the sparse activations associated with $X^n$, $\widetilde{T} = T-L+1$, and $\lambda>0$ the regularization parameter.
Here $\norme{\cdot}_F$ stands for the Frobenius norm.

In the application to M/EEG signals, we add a rank-1 constraint on the dictionary to account for the physics of the signals~\citep{dupre2018multivariate}: $D_k = u_k {v_k}^\top \in \mathbb{R}^{P\times L}$, where $u_k\in\R^P$ is the pattern over the channels (sensors) and $v_k\in\R^L$ the pattern over time.
The new minimization problem is as follows:
\begin{equation*}
    \min _{u_{k}, v_{k}, z_{k}^{n}} \sum_{n=1}^{N} \frac{1}{2}\left\|X^{n}-\sum_{k=1}^{K} z_{k}^{n} * \left(u_k v_k^\top\right)\right\|_{F}^{2}+\lambda \sum_{k=1}^{K}\left\|z_{k}^{n}\right\|_{1},
    \\
    \text { s.t. } \left\|u_{k}\right\|_{2}^{2} \leq 1, \left\|v_{k}\right\|_{2}^{2} \leq 1 \text { and } z_{k}^{n} \geq 0
\end{equation*}

\begin{figure}[!ht]
    \centering
    \includegraphics[width=\textwidth]{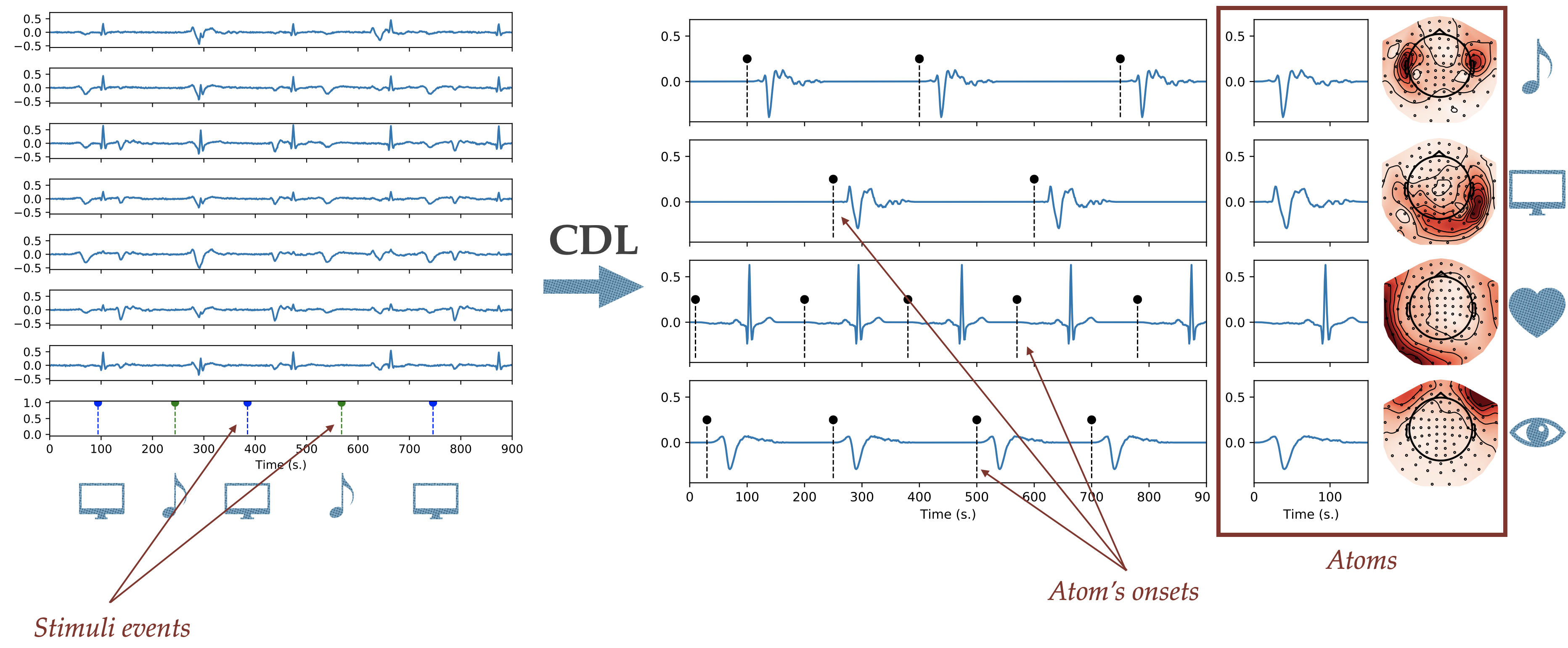}
    \caption{Schematic operation of the CDL on MEG signals.
     Raw MEG signals alongside timestamps of external stimuli of type visual and auditory (left). CDL output composed of a set of spatio-temporal atoms alongside their respective onsets (right). One may claim to associate each atom to a physical phenomenom, \ie heartbeat or eye blink artifact, auditory or visual neural response.}
    \label{fig:app:cdl_resume}
\end{figure}

The optimization is done by block coordinate descent, alternating the optimization over atoms and activations. \autoref{fig:app:cdl_resume} presents in a schematic way the functioning of CDL on MEG data.

\paragraph{Extra information about used datasets}
\autoref{tab:datasets_summary} presents the main information related to real MEG datasets that we used, both available with the \texttt{MNE} Python package~\citep{gramfort2013meg, gramfort2014mne}.
Regarding the \textit{sample} dataset, as mentioned before, four external stimuli are presented to the subject during the MEG recording session: auditory left and right and visual left and right.
Each type of stimulus leads to a so-called ``deterministic'' point process, where each event denotes the exact time the stimulus was presented to the subject.
Once the CDL was applied, and 40 atoms of duration 1\,s were extracted from the signal, a quick visual inspection of the atoms revealed that, among atoms that could be linked to audio-visual stimuli, there were mostly bimodal atoms.
Thus, this observation led us to consider both auditory (resp. visual) stimuli as one, and the two corresponding point processes were merged.
For the somatosensory dataset, as previously mentioned, 20 atoms of duration 0.53 seconds are extracted from the MEG signal corresponding to 15 minutes of recording during which the single subject has received 111 stimulations of his left median nerve at the hand level.
For both datasets, intensities functions for the EM with a Truncated Gaussian kernel were obtained similarly as \citet{allain2021dripp}, always between one atom's point process and the considered stimuli.
For the intensities with a Raised Cosine kernel, however, they were obtained using the method presented in this paper, with a grid discretization $\Delta$ equal to data re-sampling rate of 150\,Hz (\ie $\Delta = 1/150$).
Indeed, setting a $\Delta$ smaller than the discretization imposed by the data would not lead to better estimation.
Finally, the intensities estimated with the non-parametric (NP) method were obtained using \texttt{Tick} Python package \citep{ticklibrary}, with the same grid discretization parameters to have accurate comparisons.

\begin{table}[!ht]
\centering
\begin{tabular}{lllllll}
\hline
Dataset    & \# Atoms & \begin{tabular}[c]{@{}l@{}}Duration of\\ Atoms (s.)\end{tabular} & \begin{tabular}[c]{@{}l@{}}\# Atom's\\ events\end{tabular} & \# Drivers & \begin{tabular}[c]{@{}l@{}}\# Driver's\\ events\end{tabular} & \begin{tabular}[c]{@{}l@{}}Sequence\\ length (min.)\end{tabular} \\ \hline
\textit{sample} & 40          & 1                                                                & $\approx 401.025$                                          & 4             & $\approx 72.25$                                              & 4.6                                                              \\
\textit{somatosensory} & 20          & 0.53                                                             & 10408                                                      & 1             & 111                                                          & 15                                                               \\ \hline
\end{tabular}
\caption{Statistics of each dataset. $\approx N$ denotes that $N$ is the average number.}
\label{tab:datasets_summary}
\end{table}

\paragraph{Experiment on somatosensory dataset}

\begin{figure}[!ht]
    \centering
    \includegraphics[width=\textwidth]{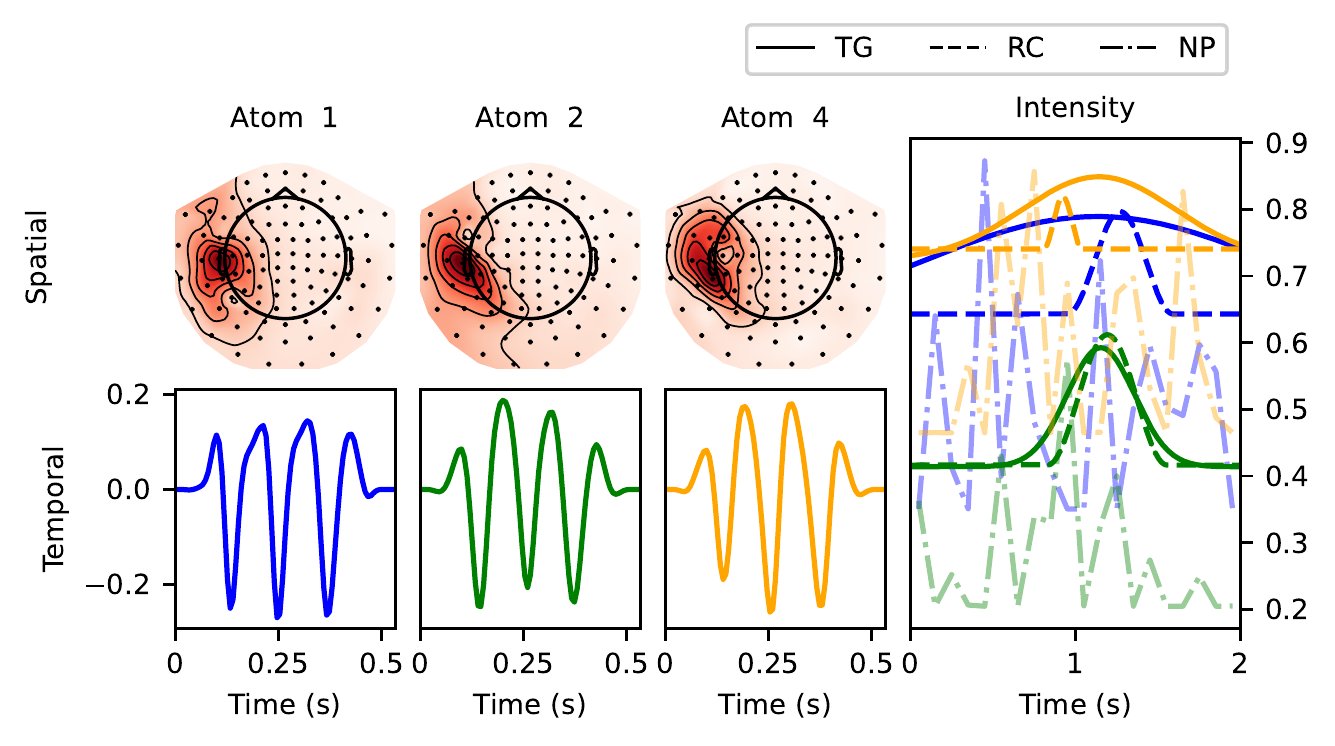}
    \caption{Spatial and temporal patterns of three $\mu$-wave atoms from \emph{somato} dataset, and their respective estimated intensity functions following a stimulus (cue at time = 0 s), for somatosensory stimuli with non-parametric kernel (NP) and two parametrized kernels: Truncated Gaussian (TG) and Raised Cosine (RC).}
    \label{fig:app:fig_somato}
\end{figure}

\autoref{fig:app:fig_somato} presents results on three atoms estimated from the \emph{somato} dataset. All three atoms elicit classical induced responses and have waveforms with a prototypical $\mu$-shape (sharp trough) \citep{hari2006action}.
Remark that in the \emph{somato} paradigm, the subject receives only one type of external stimulus.
Similarly, as in~\autoref{fig:fig_sample}, for each atom, the intensity related to the stimulus is learned with a non-parametric kernel (NP) and two kernel parametrizations: Truncated Gaussian (TG) and Raised Cosine (RC).
The non-parametric kernel cannot characterize the link between stimulus and neural response.

\clearpage


%

%

\end{document}